\documentclass{article} 
    \PassOptionsToPackage{numbers, compress}{natbib}

\usepackage[preprint]{neurips_data_2024}

\usepackage{microtype}
\usepackage{hyperref}
\hypersetup{
    colorlinks=true,
    linkcolor=red,
    citecolor=blue,
    filecolor=magenta,      
    urlcolor=blue,
linktocpage}
\usepackage{url}
\usepackage{booktabs}
\usepackage{amsmath}

\usepackage{multicol}
\usepackage{url}
\usepackage{booktabs}
\usepackage{multirow}
\usepackage{makecell}
\usepackage{arydshln}
\usepackage{xcolor}
\usepackage{pdfpages}
\usepackage{xspace}
\usepackage{CJKutf8}
\usepackage[utf8]{inputenc}
\usepackage{siunitx}
\usepackage{newunicodechar}
\newunicodechar{▪}{\ensuremath{\bullet}}
\newunicodechar{○}{\ensuremath{\bigcirc}}
\newunicodechar{●}{\ensuremath{\bullet}}
\newunicodechar{■}{\ensuremath{\blacksquare}}
\newunicodechar{□}{\ensuremath{\square}}

\usepackage{longtable}
\usepackage{graphicx}  
\usepackage{subcaption}  
\usepackage{color}
\usepackage{pifont}
\newcommand{\neo}{MAP-Neo\xspace}

\usepackage{CJKutf8}
\newcommand{\CHECK}{\textcolor{green}{\ding{51}}} 
\newcommand{\CROSS}{\textcolor{red}{\ding{55}}} 

\title{MAP-Neo: Highly Capable and Transparent \\Bilingual Large Language Model Series}

\author{M-A-P, University of Waterloo, Wuhan AI Research, 01.AI}

\usepackage{amsmath,amsfonts,bm}

\def\eqref#1{equation~\ref{#1}}

\def\1{\bm{1}}

\DeclareMathAlphabet{\mathsfit}{\encodingdefault}{\sfdefault}{m}{sl}
\SetMathAlphabet{\mathsfit}{bold}{\encodingdefault}{\sfdefault}{bx}{n}

\newcommand{\E}{\mathbb{E}}

\usepackage{amsthm}
\def\pisft{{\pi_{\text{sft}}}}

\newcommand{\lightblue}[1]{\definecolor{lightblue}{rgb}{0.68, 0.85, 0.9}\colorbox{lightblue}{#1}}
\usepackage{float}
\usepackage{listings}
\begin{document}

\maketitle
\vspace{-0.7cm}
\begin{center}
    \url{https://map-neo.github.io/}
\end{center}
\vspace{5pt}

\begin{abstract}
Large Language Models (LLMs) have made great strides in recent years to achieve unprecedented performance across different tasks. However, due to commercial interest, the most competitive models like GPT, Gemini, and Claude have been gated behind proprietary interfaces without disclosing the training details.
Recently, many institutions have open-sourced several strong LLMs like LLaMA-3, comparable to existing closed-source LLMs.
However, only the model's weights are provided with most details undisclosed (e.g., intermediate checkpoints, pre-training corpus, and training code, etc).
To improve the transparency of LLMs,
the research community has formed to open-source truly open LLMs (e.g., Pythia, Amber, OLMo),
where more details (e.g., pre-training corpus and training code) are being provided. These models have greatly advanced the scientific study of these large models including their strengths, weaknesses, biases and risks. 
However,
we observe that the existing truly open LLMs are still inferior to existing state-of-the-art LLMs with similar model sizes on reasoning, knowledge, and coding tasks.
To this end,
we open-source \textbf{MAP-Neo}, a highly capable and transparent bilingual language model with 7B parameters trained from scratch on {\textbf{4.5T} high-quality tokens}.
Our MAP-Neo is the \textbf{first fully open-sourced bilingual LLM} with comparable performance compared to existing state-of-the-art LLMs.
Moreover,
we open-source all details to reproduce our MAP-Neo,
where the cleaned pre-training corpus, data cleaning pipeline,  checkpoints, and well-optimized training/evaluation framework\footnote{\url{https://github.com/multimodal-art-projection/MAP-NEO}} are provided.
Finally,
we hope our MAP-Neo will enhance and
strengthen the open research community and inspire more innovations and creativities to facilitate the further improvements of LLMs.
\end{abstract}

\vspace{0cm}
\begin{figure}[!h]
    \centering
    \includegraphics[width=0.99\linewidth]{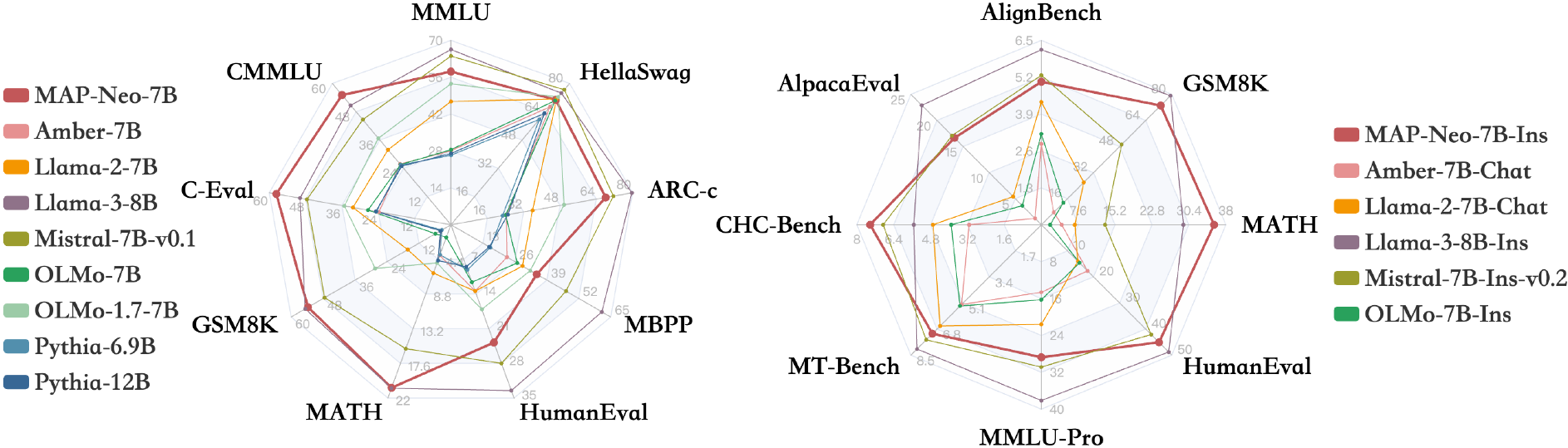}
    \caption{MAP-Neo shows impressive performance on base (\textbf{Left}) and chat (\textbf{Right}) models compared to both popular open-weight and recent transparent large language models with similar sizes.}
    \label{fig:front}
\end{figure}

\newpage
\tableofcontents
\newpage

\section{Introduction}

The advent of generalist large language models (LLMs) such as GPT-4~\citep{achiam2023gpt}, Claude~\citep{anthropic2024claude}, and Gemini~\citep{reid2024gemini} has significantly expanded the boundaries of Natural Language Processing (NLP) and is paving the way towards Artificial General Intelligence (AGI). 
These models exhibit universal capabilities, including complex reasoning~\citep{yue2023mammoth, shao2024deepseekmath}, role-playing~\citep{wang2023rolellm}, creative writing~\citep{wang2024weaver}, psychological assessment~\citep{psychogat}, scientific education~\citep{llm-science-tutor}, and music generation~\citep{yuan2024chatmusician,qu2024mupt,deng2024composerx}, among others. However, the most advanced ones remain closed-source due to commercial interests~\citep{achiam2023gpt, anthropic2024claude,reid2024gemini}. 
In this paper, we argue that open-source and transparent LLMs are essential for both the democratization of LLMs and further academic research, especially considering the substantial resources these models consume.

Previous works have released numerous open-source or even transparent LLMs. For example, the LLaMA series~\citep{touvron2023llama, touvron2023llama2, llama3modelcard} released the weights, thereby significantly boosting the development of the open-source LLM community. However, they are not transparent because they do not disclose the details of their training data. BLOOM~\citep{bloom} trained a multilingual language model with 176 billion parameters and open-sourced its model weights, intermediate checkpoints, and training corpus. Models like LLM360~\citep{liu2023llm360} and Pythia~\citep{biderman2023pythia} further provided their training codes, optimizer state checkpoints, analysis codes, and data pipelines. 

These models make significant contributions to building transparent ecosystems, yet generally lag behind industry-level LLMs such as LLaMA~\citep{llama3modelcard}, Mistral~\citep{jiang2023mistral} and Yi~\citep{young2024yi}, etc. OLMo~\citep{groeneveld2024olmo} has made a great stride in narrowing this gap by improving pre-training data and data processing pipelines, and introducing more open-source components, including training logs and ablations. 
Nonetheless, it remains less proficient, especially in areas like coding (HumanEval~\citep{chen2021evaluating}), reasoning (MATH~\citep{hendrycks2021measuring}, GSM8K~\citep{cobbe2021training}), knowledge (MMLU~\citep{hendrycks2020measuring}), and multilingualism (CMMLU~\cite{li2023cmmlu}).

To remedy these issues, we introduce MAP-Neo, a fully open-source and transparent bilingual LLM suite that achieves superior performance to close the gap with closed-source models. 
Specifically, the entire workflow of building an LLM includes:

\begin{enumerate}
    \item \textbf{Data Curation Pipeline}: We provide the code for the curation and cleaning of training data (both English and Chinese), including a stable OCR system, the data recalling mechanism in DeepSeek-Math~\citep{shao2024deepseekmath}, the integration of previous open-source data processing pipelines, and support for distributed data processing based on Spark\footnote{\url{https://spark.apache.org/}}, among others. 
    \item \textbf{Data}: We release our pre-training corpus, namely Matrix Data Pile, along with the training data for supervised fine-tuning and alignment training.
    \item \textbf{Model Architecture}: We provide the codes and details of our modeling architecture.
    \item \textbf{Model Training}: We offer the training codes for our tokenizer, base models, instruction-tuned models, and aligned models. Additionally, we address some issues of the Megatron-LM framework\footnote{\url{https://github.com/NVIDIA/Megatron-LM}}, enhancing its support for more robust and efficient distributed training. Moreover, we introduce the NEO Scaling Law designed to optimize scaling up LLMs using a pre-training dataset sourced from diverse corpora. 
    \item \textbf{Model Checkpoints}: We not only release the final models on HuggingFace but also make the intermediate checkpoints available for reproducibility. 
    \item \textbf{Infrastructure}: This report details the infrastructure for stable training.
    \item \textbf{Evaluation}: We also provide detailed evaluation codes and thorough evaluation settings for benchmarking the performance of LLMs.
    \item \textbf{Analysis and Lessons}: This report elaborates on numerous techniques and recipes, such as optimization tricks at different phases of pre-training, and offers insights into building LLMs through rigorous analysis and ablations.
\end{enumerate}

Our work is a milestone towards fully transparent LLMs with advanced abilities, even competitive with the top closed-source LLMs. Notably, our contribution is not just a novel foundational model but also a comprehensive handbook for building LLMs from scratch, covering the entire workflow. We believe that our model provides a critical reference for the community, particularly for non-English regions of the world engaged in LLM research.

\section{Related Works}
\begin{table}[ht]
\centering
\caption{Compare with other open-source large language models (LLMs). All metrics are obtained using the same evaluation manner, and the details are shown in Table~\ref{tab:base_model_compare}. Non-transparent models are listed above the dashed line, while the transparent LLMs are shown below.}
\label{tab:compare-llm}
\resizebox{1\linewidth}{!}{
\begin{tabular}{lcccc:cccc}
\toprule
\multirow{2}{*}{\textbf{Model}} & Intermediate & Pre-training  & Reproduction &  Data Cleaning & \multirow{2}{*}{C-EVAL} & \multirow{2}{*}{MMLU} & \multirow{2}{*}{GSM8K} & \multirow{2}{*}{HumanEval} \\ 
& Checkpoints & Corpus & Code & Process &  &  &  &  \\
\midrule
\textcolor{gray}{Mistral-7B~\citep{jiang2023mistral}} & \CROSS & \CROSS & \CROSS & \CROSS & \textcolor{gray}{47.54} & \textcolor{gray}{64.04} & \textcolor{gray}{47.46} & \textcolor{gray}{28.0} \\
\textcolor{gray}{LLaMA2-7B~\citep{touvron2023llama2}}  & \CROSS & \CROSS & \CROSS & \CROSS & \textcolor{gray}{32.37} & \textcolor{gray}{46.80} & \textcolor{gray}{16.22} & \textcolor{gray}{13.4} \\
\textcolor{gray}{LLaMA3-8B~\citep{llama3modelcard}}  & \CROSS & \CROSS & \CROSS & \CROSS & \textcolor{gray}{\textbf{49.83}} & \textcolor{gray}{\textbf{66.52}} & \textcolor{gray}{\textbf{54.74}} & \textcolor{gray}{\textbf{33.5}} \\
\hdashline
Pythia-6.9B~\citep{biderman2023pythia} & \CHECK & \CHECK & \CHECK & \CHECK & 24.64 & 26.39 & 3.41 & 9.1 \\
Amber-7B~\citep{liu2023llm360} & \CHECK & \CHECK & \CHECK & \CROSS & 23.82 & 28.07 & 3.64 & 13.4 \\
OLMo-7B~\citep{groeneveld2024olmo} & \CHECK & \CHECK & \CHECK & \CHECK & 35.21 & 53.52 & 28.43 & 11.6 \\
\textbf{MAP-Neo-7B} & \CHECK & \CHECK & \CHECK & \CHECK & \textbf{57.68} & \textbf{58.14} & \textbf{53.68} & \textbf{23.8} \\
\bottomrule
\end{tabular}
}
\end{table}

The development of open-source large language models (LLMs) is pivotal for advancing artificial intelligence research and applications. Recent efforts in this domain have been focused on not only enhancing model performance~\citep{jiang2023mistral,llama3modelcard} but also ensuring transparency and reproducibility~\citep{biderman2023pythia, liu2023llm360, groeneveld2024olmo, zhuang2024chuxin}. Our model, MAP-Neo-7B, emerges as the new lead in this evolving landscape, as shown in Table \ref{tab:compare-llm}, which balances performance and transparency. 

The MAP-Neo model series represents a step forward in emphasizing full transparency, aligning it alongside other contemporary models such as Mistral~\citep{jiang2023mistral}, LLaMA3~\citep{llama3modelcard}, Pythia~\citep{biderman2023pythia}, Amber~\citep{liu2023llm360}, and OLMo~\citep{groeneveld2024olmo}. Unlike these models, which often lack either intermediate checkpoints, comprehensive data cleaning processes, or accessible pre-training corpus and reproduction code, MAP-Neo excels by integrating all these elements. 
This commitment to the openness of MAP-Neo facilitates in-depth analysis and independent validation by the research community.

Performance-wise, MAP-Neo-7B demonstrates superior capabilities across a broad scope of benchmarks including Chinese and English understanding on C-EVAL~\citep{huang2024c} and MMLU~\citep{chung2024scaling}, mathematical ability on GSM8K~\citep{cobbe2021training} and MATH~\citep{hendrycks2021measuring}, and code ability on HumanEval~\citep{chen2021evaluating}. Notably, MAP-Neo-7B is the only model in our comparative analysis to achieve all checks in transparency, as well as the highest scores across all tests compared with other transparent LLMs, underscoring the effectiveness of the training and the quality of the data. 

The most similar work to MAP-Neo is OLMo~\citep{groeneveld2024olmo}, which is the pioneering work to fully open-source LLMs.
However, their performance is compromised in several aspects like knowledge, coding, and mathematical reasoning. Moreover, OLMo cannot handle languages beyond English. \neo sets a new standard for transparency and performance in the field of open-source LLMs. By fostering a fully transparent development process, MAP-Neo not only enhances its utility and trustworthiness but also provides a valuable framework for future research, promoting further advancements and collaborative efforts in the community. 

\section{Tokenizer}
\label{sec:tokenizer}
We train our tokenizer using the byte-pair encoding (BPE) algorithm \citep{sennrich2015neural} via the implementation of SentencePiece \citep{kudo2018sentencepiece}. The training data consists of 50B samples from the pre-training corpus, and the maximum length is cut to 64K. We assign higher sampling weights to code, math, and high-quality academic data. To balance the computational efficiency and model performance, we propose to set the vocabulary size to 64000 and constrain the max sentence-piece length to 16 to improve the Chinese performance. 

Notably, we slice all numbers into individual digits and fall back unknown UTF-8 characters to byte granularity. We do not use any normalization strategy on the training samples and do not add dummy prefixes. The character coverage rate is set to 0.9999. Particularly, the remove extra whitespaces parameter is set to False, which is turned on by default in the SentencePieceTrainer. This setting can severely impact code performance during pre-training, as normal code indentation is treated as a single space. We encountered a specific issue during the initial phase of our model's pre-training. Initially, we did not disable the `remove extra whitespaces' parameter, which is enabled by default in the SentencePieceTrainer. 
In the training process, we observe steady improvements in the QA reasoning and mathematics benchmarks,
but the code metrics exhibit fluctuations and do not show expected improvements. To address this issue, we fixed this bug in the second phase of our training (\S \ref{sec:decay_phase}), which stabilizes and significantly improves the code metrics. Furthermore,
we observe that 
this issue is well addressed in the decay phase training stages under the new tokenizer settings,
where rapid improvements are achieved.

Moreover,
we also investigate the compression rates across various categories of data, categorized by both language (Chinese and English) and data source quality (high-quality and web-sourced) as shown in Table~\ref{tab:compression_rates}.
Specifically,
first,
we observe that the
high-quality data (HQ) including complex reasoning, mathematical, and general knowledge texts, showing different compression rates between Chinese (HQ\_cn) and English (HQ\_en). The HQ\_cn category has a compression rate of 1.577, while the HQ\_en category exhibited a higher rate of 3.311 characters per token. Second, data sourced from the web (Web) also comprise more characters than Chinese ones. This suggests a significant variation in tokenization efficiency or character usage between languages, possibly due to the linguistic structure and the tokenization methods. 
Third,
it should be mentioned that
even with similar compression rates, the settings of the tokenizer can cause significant fluctuations in the pre-training process. Therefore, it remains necessary to further investigate tokenization strategies for subsequent usage scenarios.

\begin{table}[h]
\centering

\caption{Average Compression Rates by Category. These subsets are not uniformly proportioned in the training set. A detailed distribution is shown in {Appendix \ref{tab:detailed_compression}}.}
\label{tab:compression_rates}
\begin{tabular}{@{}lccccc@{}}
\toprule
\textbf{Code} & \textbf{HQ\_cn} & \textbf{HQ\_en}  & \textbf{Web\_cn} & \textbf{Web\_en} & \textbf{Others} \\ 
\midrule
2.951 & 1.577 & 3.311 &  1.418 & 3.699 & 2.558 \\ 
\bottomrule
\end{tabular}
\end{table}

\section{Matrix Data Pile}

\begin{figure}[htbp]
    \centering
\includegraphics[width=0.6\textwidth]{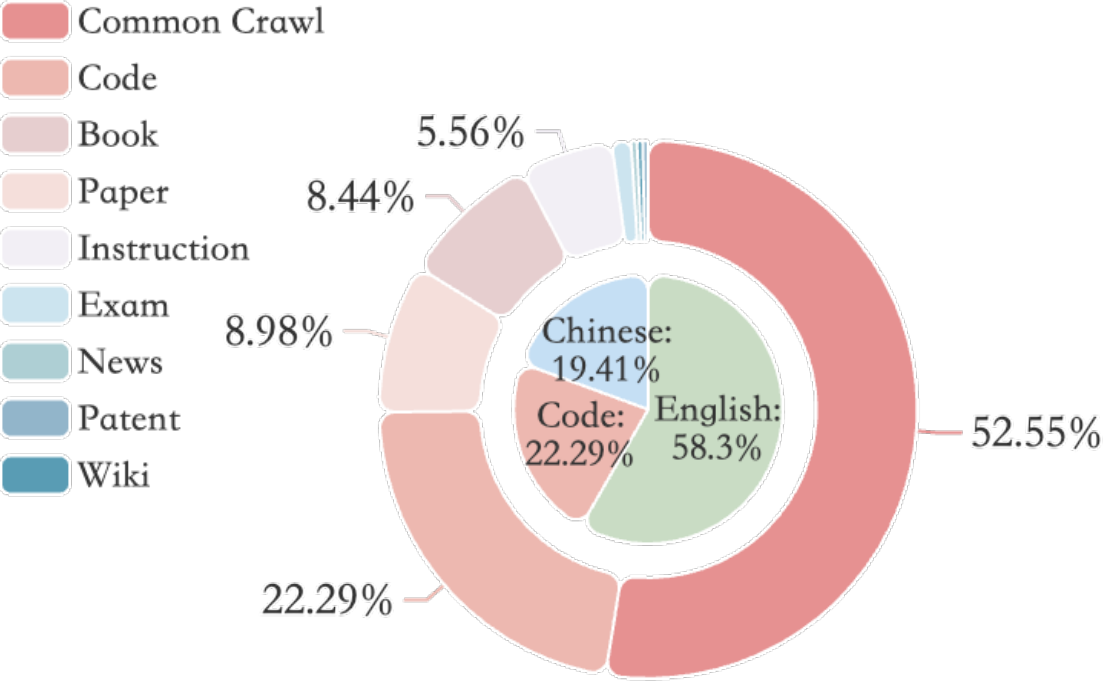}
  \caption{Statistics of the Matrix Pile Data Distribution: The inner pie chart represents the language distribution, while the outer loop indicates the proportion of meta-categories in the corpus.}
  \label{fig:matrix_stats}
\end{figure}

It is widely recognized that a well-constructed training corpus is essential for training LLMs. The training corpus serves as the fuel driving advancements in language modeling, as demonstrated by the emergent capabilities of models like ChatGPT, Claude, Gemini, and Llama. However, due to intellectual property restrictions, the pre-training data and processing toolkits of these (partially) proprietary LLMs are not disclosed upon release. Although the open-source research community has made substantial efforts to increase transparency in the collection and processing pipeline of language model pre-training data~\citep{biderman2023pythia, bloom, dolma}, the development of fully open-sourced LLMs still lags behind proprietary LLMs to some extent, primarily due to gaps in the quantity and quality of the datasets.

To address the pressing need for more diverse and transparent datasets in language modeling, we introduce Matrix, a bilingual pre-training corpus of 4.5T tokens. 
Upon its release, Matrix could be {the largest transparent LLM pre-training corpus to our best knowledge}.
Specifically,
Matrix provides the details of the data collection and processing along with a high-performance toolkit. 
Additionally,
we design Matrix based on the idea of retrieving, filtering, and cleaning high-quality data under various practical circumstances, which are discussed as follows:
\begin{itemize}
    \item Given a set of existing (English) pre-training datasets, how do we re-process and \textit{improve the quality}?  \S\ref{sec:en_web_pipeline}
    \item How do we construct a large-scale, topic-comprehensive corpus \textit{from scratch}, on the less explored Chinese content?\S\ref{sec:cn_web_pipeline} 
    \item If we have enormous \textit{printed documents}, how do we build an efficient and effective system to extract viable textual contents? \S\ref{sec:doc_pipeline}
    \item When specifying a \textit{domain of interest},  how do we find relevant high-quality data from the wild of web content? \S\ref{sec:crawl_pipeline} 
\end{itemize}

The final composition of the corpus is as follows: 52.55\% from Common Crawl, 22.29\% from programming code, and the rest from academic papers, books, and other printed materials, as illustrated in Figure~\ref{fig:matrix_stats}. 
The detailed methodologies for processing these sources are described in the subsequent sections, and a comprehensive illustration of the sources is provided in Table~\ref{tab:open-source-dataset-list}.

\begin{table}[ht]
\caption{The composition sources of re-processed English web subset. The proportion denotes dividing the size of the current dataset by the total size of the whole dataset.}
\label{tab:en_data_distribution}
\centering
\small
\begin{tabular}{
  l
  l
  l
  S[table-format=1.2]
  S[table-format=1.2]
}
\toprule
\textbf{Dataset} & \textbf{Parts} & {\textbf{UTF-8 bytes (TB)}} & {\textbf{Availability}} & {\textbf{Proportion (\%)}} \\
\midrule
\textbf{RedPajama-Data-V2~\citep{together2023redpajama}} & Head and Middle & 200 & Public & 92.38 \\
\textbf{Dolma~\citep{dolma}} & CC & 6.4 & Public & 2.96 \\
\textbf{Cultrax~\citep{nguyen2023culturax}} & EN & 1.2 & Public & 0.55 \\
\textbf{Amber~\citep{liu2023llm360}} & Refined-Web & 4.23 & Public & 1.95 \\
\textbf{SlimPajama~\citep{cerebras2023slimpajama}} & Whole & 2.43 & Public & 1.12 \\
\textbf{Falcon~\citep{refinedweb}} & Whole & 1.01 & Public & 0.47 \\
\textbf{CultraY~\citep{nguyen2024culturay}} & EN & 1.24 & Public & 0.57 \\
\bottomrule
\end{tabular}
\end{table}

\subsection{Re-processing Pipeline for Open Datasets}\label{sec:en_web_pipeline}
Although several processed pre-trainig corpus (mostly in English) have been released by previous works~\citep{dolma,refinedweb}, we argue that there is still room for a more meticulously designed pipeline to improve the existing data.
Besides, it should be mentioned that
existing LLMs can be easily improved by continuous pre-training with high-quality data.
Therefore,
we further re-process the selected web content-based corpora to produce the English subset of Matrix data mixture. The source comes from the Head and Middle parts of RedPajama-Data-V2~\citep{together2023redpajama}, CC part of Dolma~\citep{dolma}, the EN part of Cultrax~\citep{nguyen2023culturax}, the Refined-Web part of Amber~\citep{liu2023llm360}, SlimPajama~\citep{cerebras2023slimpajama} and falcon~\citep{refinedweb}. 
The precise distribution of our English dataset is listed in Table~\ref{tab:en_data_distribution}. The procedure involves filtering and multi-step deduplication. The diagram in Figure \ref{fig:english_web_pipeline} shows the processing orders and the retention rates.

\subsubsection{Filtering}
To further filter out the relatively low-quality corpus from open-source datasets, we propose to use heuristic rules for text filtering. These rules are designed to identify and remove poor-quality data, thereby preventing potential model performance degradation caused by a flawed pre-training corpus. 
Since our composite dataset is made up of corpora from multiple sources, we adapt well-designed cleaning methods~\citep{refinedweb, chen2023chinesewebtext, rae2022scaling, 2019arXiv191010683R} and tailor our rules for each one to ensure quality consistency.

For the RedPajama-Data-v2 dataset ~\citep{together2023redpajama}, which provides quality annotations for each text, we integrate our heuristic rules with these annotations to refine data quality evaluation and further perform random sampling on the dataset to confirm the thresholds for every rule. For datasets lacking quality annotations, we apply the established rules and thresholds derived from RedPajama-V2, while customizing them to align with the unique characteristics of each dataset. For example, the Dolma dataset \cite{dolma} comprises six subsets, namely Wikipedia, PeS2o, Stack Code, Gutenberg, C4, and CC, each with different data characteristics. Given the unique characteristics of each subset, we conduct individual sampling and evaluation to ensure that the modifications in rules and thresholds are aligned with our filtering requirements. Specifically, for the CC subset, we adjust the unique word and text length thresholds. For the Gutenberg subset, which predominantly contains book texts, we apply only a few rules to avoid the time-consuming process of executing extensive heuristic checks on long texts.

The filtering process involves: 1) Document-level and sentence-level filtering to ensure text length adequacy, character meaningfulness, and consistency; 2) Duplicate text removal, including n-grams and sentences; 3) Sensitive word check to eliminate texts containing any terms from a blacklist.

\subsubsection{Deduplication}\label{sec:en_dedup}
It has been reported that repetitive text can lead to a decline in model performance~\citep{lee2021deduplicating, kaddour2023minipile, hernandez2022scaling}, which makes deduplication a crucial step in corpus processing. 
By eliminating duplicates, we can significantly reduce the rate of emitted memorization and make model training more efficient~\cite{lee2021deduplicating}. 
Repetitions can be categorized into exact duplicates and near duplicates. For exact duplicates, we employ exact document deduplication to remove them. For near duplicates, we utilize Minhash LSH deduplication to remove them as much as possible. In addition, there are instances where parts of the text are completely duplicated, and in these cases, the Minhash method struggles to remove them. To address this, we have adopted two methods for partially removing such content: paragraph deduplication and exact substring deduplication.

\paragraph{Exact Document Deduplication}
\label{en_exact_document_dedupliaction}
Exact document deduplication is a method used to evaluate an entire text to determine if it is identical to another. If it is found to be exactly the same, the duplicate will be removed. For processing data in English, Spark is employed to handle the dataset. Due to the vast volume of data, there may be issues with insufficient memory. The solution to this problem involves batching the text data into separate buckets for storage. Each bucket's data is then processed in turn to remove duplicates.

\paragraph{Minhash LSH Deduplication}
\label{en_minhash}
Minhash~\citep{broder1997resemblance} is an excellent method for removing near duplicates, especially for web page data, and is widely used for similarity search and duplicate detection in large datasets~\citep{versley2012not,gabriel2018identifying,gyawali2020deduplication}. 
It can handle very common scenarios where the text content is essentially the same, but the scattered template blocks of the web pages are different.
The principle of MinHash is to represent a set with smaller hash values, which can then be used to estimate the Jaccard similarity~\citep{jaccard1912distribution} between two sets: $\text{Jaccard}(A, B) = (A \cap B)/(A \cup B)$.

MinHash involves using multiple distinct hash functions that map each element of a set to a larger numerical domain. For each set, these multiple hash functions are applied to all elements within the set, and the smallest hash value produced by each hash function is chosen as its minimum hash value. Thus, each set can be represented by a vector of these minimum hash values, forming the set's MinHash signature. For text data, an n-gram approach can be used to construct a set. 

After obtaining the signature of the text, Locality-Sensitive Hashing (LSH) ~\citep{gionis1999similarity} is employed to rapidly identify candidate set pairs that exceed a certain threshold in Jaccard similarity. This accelerates the search process for similar items. The specific approach divides the signature into several bands, each containing several hash values. Another hash function is then used to map each band to a hash bucket. All sets with the same band hash are mapped to the same hash bucket. All set pairs in the same hash bucket are considered candidate similar pairs without further specificity regarding their similarity. Here, we utilize 128 unique hash functions to form signatures, divided into 9 bands, with each band containing 13 hash values. Consequently, the Jaccard threshold is set at 0.8.

Upon identifying similar pairs, connected components are constructed. Within each component of the connected components, one text is retained while the others are eliminated. For processing vast amounts of data efficiently, a distributed implementation~\citep{kiveris2014connected} based on map-reduce is adopted.

\paragraph{Paragraph Deduplication}
Paragraph deduplication involves removing all duplicate paragraphs within a text. A paragraph is defined as a section of text separated by the newline UTF-8 character "\textbackslash n". Paragraph deduplication is an effective method for removing website navigation headers, advertisements, and similar elements. Since paragraph deduplication involves deleting parts of the text, it may cause some interference with content analysis.

Its concrete implementation first involves splitting the text into multiple paragraphs using newline utf-8 character "\textbackslash n", with each paragraph being tagged with its corresponding document id and offset in the text. Then, each paragraph is hashed using SHA256. Next, the hash values are deduplicated. After deduplication, the deduplicated text is restored according to the document ID and offset.

\paragraph{Exact Substring Deduplication}
This method follows \cite{lee2021deduplicating}. Given the diversity of languages, when the length of repeated text is sufficiently long, it is highly likely that they are either derived from one another or sourced from the same reference. Therefore, when two texts, $t_i$ and $t_j$ share sufficiently a long substring, that is $t_i^{a..a+k} = t_j^{b..b+k}$, one of them is removed. For the selection of the length threshold, we adhere to the setting in \cite{lee2021deduplicating}, choosing k=50. Due to our distributed environment, the memory of a single node is insufficient to hold all the data. Therefore, we did not adopt the implementation in \cite{lee2021deduplicating}. In our work, we segment each text into sliding windows of 50 characters with a step size of 1. We then compute the SHA256 hash value for each window along with its corresponding document ID and offset. Subsequently, for windows with identical hash values, we mark them as duplicates except the first one. Finally, using the text ID and offset, we restore the original strings and decide whether to delete a segment based on the duplicate marker.

\begin{figure}[h]
\centering
 \begin{subfigure}[b]{0.4\textwidth}
     \centering
     \includegraphics[width=\textwidth]{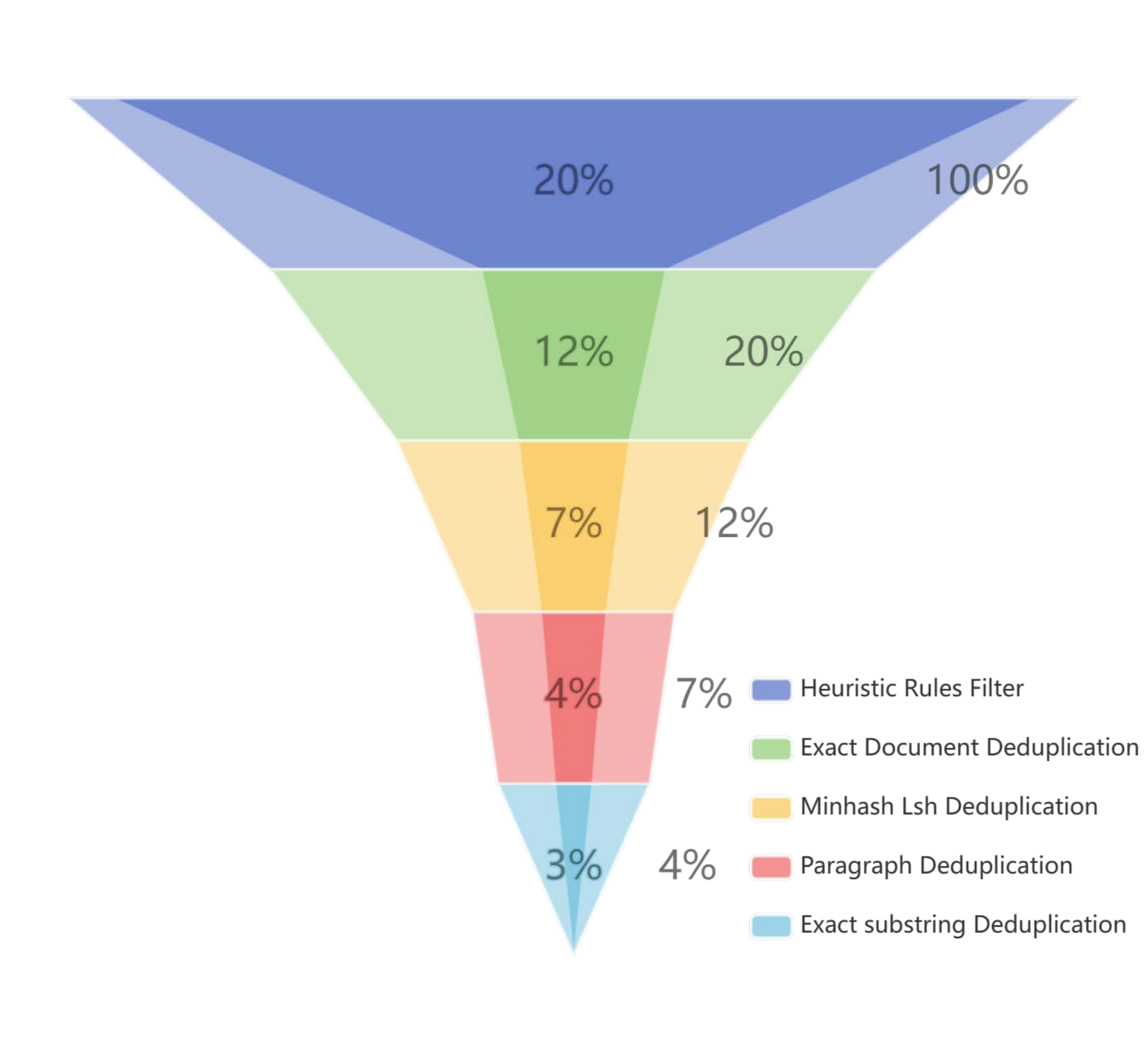}
     \caption{Re-processing retention rates for the corpora in \S\ref{sec:en_web_pipeline}.}
     \label{fig:english_web_pipeline}
 \end{subfigure}
 \hspace{0.1\textwidth}
 \begin{subfigure}[b]{0.405\textwidth}
     \centering
     \includegraphics[width=\textwidth]{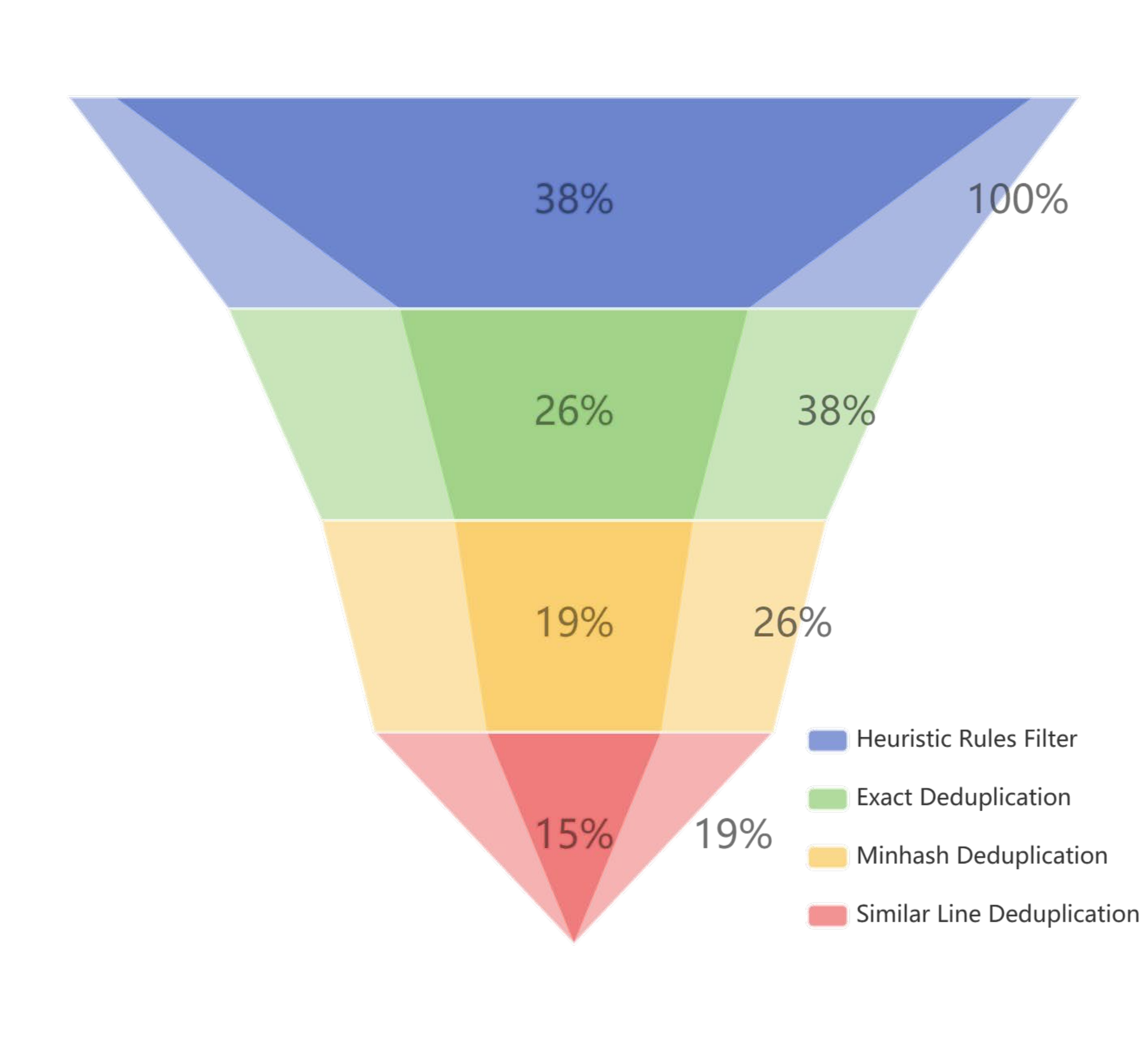}
     \caption{Processing retention rates for the corpora crawled from scratch in \S\ref{sec:cn_web_pipeline}.}
     \label{fig:chinese_web_pipeline}
 \end{subfigure}
\caption{
Funnel Diagram for the two main data pipelines. 
The darker part of each row represents the retention proportion for each processing step and the lighter one for the filtered corpora.
}
\label{fig:matrix_funnel}
\end{figure}

\subsection{Corpora Crawl from Scratch Pipeline}\label{sec:cn_web_pipeline}
We further provide a pipeline to crawl and process the web content from scratch and showcase it with the Chinese language data, which could be a step-by-step guide for follow-up research to build a new up-to-date corpus.
We take the corpus produced in such a pipeline as the Chinese subset of Matrix, where $80.6\%$ is derived from the Chinese web pages we crawled and others from several open datasets, as listed in Table~\ref{tab:zh_data_distribution}.
The pipeline overview and details are illustrated in Figure~\ref{fig:chinese_web_pipeline}.

\begin{table}[ht]
\centering
\small{
\caption{The composition sources of the Chinese web subset.}
\label{tab:zh_data_distribution}
\begin{tabular}{
  l
  c
  c
  c
  S[table-format=1.2]
}
\toprule
\textbf{Dataset} & \textbf{Parts} & {\textbf{UTF-8 bytes (TB)}} & {\textbf{Availability}} & {\textbf{Proportion (\%)}} \\
\midrule
\textbf{Crawled Web Data} & Whole & 14.3 & Self-constructed & 80.60 \\
\href{https://huggingface.co/datasets/BAAI/CCI-Data}{\textbf{CCI}} & Whole & 0.10 & Public & 0.59 \\
\textbf{Chinesewebtext~\citep{chen2023chinesewebtext}} & Whole & 1.40 & Public & 7.89 \\
\textbf{Wanjuan~\citep{he2023wanjuan}} & Text & 0.57 & Public & 3.19 \\
\textbf{Yayi2~\citep{luo2023yayi}} & Whole & 0.49 & Public & 2.76 \\
\textbf{Cultrax~\citep{nguyen2023culturax}} & ZH & 0.28 & Public & 1.56 \\
\textbf{Skypile~\citep{wei2023skywork}} & Whole & 0.60 & Public & 3.41 \\
\bottomrule
\end{tabular}
}

\end{table}

\subsubsection{Filtering}
The filtering rules for Chinese datasets are specifically tailored to address their unique challenges, differing from those applied to relatively well-processed English datasets in \S\ref{sec:en_web_pipeline}. 
Considering the large proportion of HTML-converted data in Chinese datasets, we focus intensively on eliminating HTML-related artifacts and rectifying textual inconsistencies. Furthermore, given the significant linguistic differences between Chinese and English, we conduct targeted sampling of documents within Chinese datasets, which aims to reassess and adjust the thresholds and details of our filtering rules, ensuring their suitability for the unique language characteristics of Chinese text. For example, we refine the rules to distinguish between `characters' and `words' in Chinese texts, adapting the tokenization method accordingly.

Our Chinese filtering steps are similar to the rules adapted to filter Massive Appropriate Pre-train Chinese Corpus (MAP-CC) \cite{du2024chinese}: 1) Data format unification to boost processing efficiency. 2) URL removal. This step is conducted in two stages: first, removing texts with URLs listed in Blacklist T1; followed by a comprehensive sweep to eliminate residual URLs. 3) Sentence-level and document filtering to discard text that is excessively brief, substandard, or logically incoherent. 4). Duplicates removal, including n-grams and sentences.
\subsubsection{Deduplication}
The deduplication of Chinese data includes Exact Document Deduplication, MinHash Deduplication, and Similar Line Deduplication. Due to difficulties in deploying Spark in the environment for processing Chinese, we have re-implemented the first two methods. For Exact Document Deduplication, there are slight differences from the implementation for English, mainly to save memory, where we have adopted a Bloom Filter approach and set the false positive rate of the Bloom Filter to 0.001. Discussions on Exact Document and MinHash LSH Deduplication can be found in \S\ref{sec:en_dedup}. 

We did not use Exact substring deduplication because when crawling web pages, it is common to repeatedly crawl the same content multiple times in a signal document. Additionally, when extracting the main text from HTML, there is often a loss of one or two words. The combination of these two situations violates the assumption in \cite{lee2021deduplicating} that ``it is rare for the same idea to be expressed identically in multiple documents unless one expression is derived from the other, or both are quoting from a shared source." Therefore, after Exact substring deduplication, there will be cases where extra words are retained, greatly reducing the readability of the text. Hence, we propose a Similar Line deduplication method to address this issue.

\subsubsection{Similar Line Deduplication}
To address the scenario where identical content appears multiple times within a text,
a direct method is to divide the text into lines using specific delimiters and then compare the similarity between each line. If they are similar, the subsequent line is removed. The division of lines includes the use of the following delimiters: 
``[", ``.", ``!", ``?", ``\textbackslash", ``……", ``]". We use edit distance to judge whether two lines $L_1$ and $L_2$ are similar as follows:
\[
\text{isSimilar}(L_1, L_2) = 
\begin{cases} 
True & \min(|L_1|, |L_2|) \geq 15 \land \text{editDist}(L_1, L_2) < 0.1 \times \min(|L_1|, |L_2|) \\
True & \min(|L_1|, |L_2|) < 15  \land L_1 = L_2 \\
False & \text{otherwise},
\end{cases}
\]
where $|L|$ is the length of line $L$ and ``editDist" is short for edit distance.

Due to the computational complexity of calculating edit distance being $O(len(L_1) \times len(L_2))$, to accelerate this process, we additionally propose two  methods to judge dissimilarity:
\begin{enumerate}
    \item Is the length difference between the two lines greater than one-tenth of the length of the shorter line?
    \item Is the ratio of the intersection of the sets of characters and the union of the sets of characters in  $L_1$ and $L_2$  less than one-third?
\end{enumerate}
Note that the first method has a computational complexity of $O(1)$,  and the second method has a complexity of $O(len(L_1) + len(L_2))$. Thus, these methods can significantly improve the speed of calculation.
Clearly, if either of the above two questions is positive,
the lines cannot be considered similar.
Otherwise,
we calculate $\text{isSimilar}(L_1, L_2)$ to obtain the similarity between  $L_1$ and $L_2$.

\begin{figure}[htbp]
    \centering
      \includegraphics[width=0.95\textwidth]{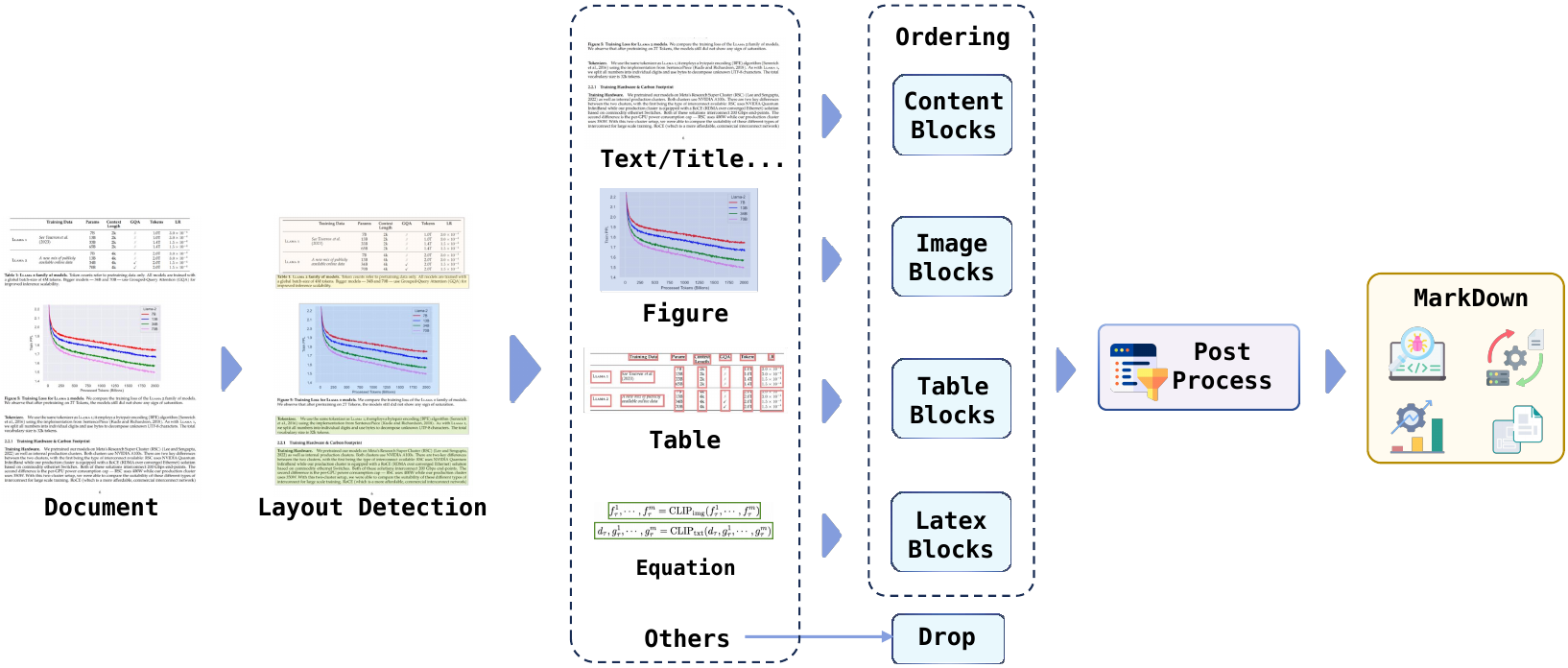}
  \caption{The document conversion framework is composed of various sub-models for different parts.}
  \label{fig:document_conversion_pipeline}
\end{figure}

\subsection{Document Conversion Pipeline}\label{sec:doc_pipeline}
The documents are usually better formatted, in concentrated topics, and with more consistent expressions compared to noisy web content. However, it seems to be a gold mine of high-quality corpus except that the golds lie deeply under the digital dirt.
Such digital documents are mostly stored as standard PDFs with diverse layouts or scanned images with inconsistent quality, making it challenging to build datasets upon.
We observe two core issues in designing an effective conversion pipeline to extract plain text from documents: 
i) analyzing layout information and identifying different layout elements including text, titles, captions, images, tables, and formulas, and 
ii) recognizing the relationships among these layout components.

We survey the existing open-source solutions for document conversion and find some distinguished projects with good performances: PP-StructureV2~\citep{li2022pp}, Marker\footnote{\url{https://github.com/VikParuchuri/marker}}, Vary~\citep{wei2023vary}, and Nougat~\citep{blecher2023nougat}.
However, along with their merits, each of them exhibits limitations that could be addressed to further enhance performance:
PP-StructureV2 cannot recognize LaTeX format content and necessary post-processing stages; 
Marker and Texify\footnote{\url{https://github.com/VikParuchuri/texify}} support few languages and do not process figures effectively; 
Nougat has limited support for multi-column data and recognized languages; 
Vary and Vary-toy require considerable computational resources. 
Therefore, we propose a framework consisting of disentangled processing components, allowing us to leverage the strengths of these models together.
For example, we utilize Marker for enhanced language support and PP-StructureV2 for efficient layout parsing.
As illustrated in Fig.~\ref{fig:document_conversion_pipeline}, our document conversion framework is comprised of four parts: Layout Detection, Element Recognition, Ordering, and Post Process. The decoupling between each module enhances interpretability and simplifies the upgrade, addition, and replacement of various components.

\paragraph{Layout Detection} segments the document into multiple parts such as formulas, text, headers, and footers. The Pipeline employs a lightweight target detection model provided by PP-StructureV2, which is computationally efficient and performs exceptionally well. This model's performance is further enhanced by employing the FGD (Feature Gradient Descent) algorithm, which optimizes feature extraction for more accurate layout detection. 

\paragraph{Element Recognition }incorporates various models to identify different elements. For formula recognition, the TrOCR model trained through Pix2Text outperforms other formula recognition models such as Latex-OCR and Taxify, supporting recognition of formulas embedded within paragraphs and non-conventional formulas, thus effectively addressing most formula recognition scenarios. Text recognition employs PP-OCRv4, Text recognition employs PP-OCRv4, notable for its compatibility with multiple computing devices and boasts strong recognition capabilities; approximately one hundred language recognition models have been publicly released, applicable to a broader range of document recognition tasks. Figures are saved as images and inserted in the subsequent merging phase. Table reconstruction is achieved using SLANet, which represents tables in HTML format. Other regions, such as headers, footers, and page numbers, are discarded and do not proceed to the post-processing and reconstruction stages.

\paragraph{Ordering} In document conversion tasks, correctly handling the relationships between blocks is of paramount importance. To acquire high-quality conversion data, we need to properly handle complex layout scenarios such as multi-column and cross-page conditions. In the ordering stage, we use LayoutLMv3~\citep{huang2022layoutlmv3} for column detection and sorting different areas according to specific rules. This strategy not only enhances the accuracy of the task but also significantly optimizes the readability.

\paragraph{Post-processing.} The texts extracted by OCR usually could not be directly used and require additional processing as follows:
\begin{enumerate}
    \item  Broken-up sentences: In text extracted from images, sentences may be fragmented across different lines or pages, resulting in a single sentence being divided into multiple segments. Effective OCR text extraction necessitates the identification and rejoining of these fragmented sentences to reconstruct coherent, complete sentences.
    \item Hyphenated words: 
    Certain words may be split into two parts within the text due to formatting constraints, connected by hyphens (e.g., network-ing). Text extraction must recognize these hyphenated words and merge them back into a single, complete word (e.g., networking).
    \item Broken math formulas: OCRed mathematical formulas in Markdown may experience issues such as missing elements, incorrect symbols, or fragmented expressions. To address this issue, we fine-tune a 7-billion parameter open-source pre-trained language model~\cite{bai2023qwen} on supervised learning data pairs $(x_{i}, y_{i})$. Here, $x_{i}$ represents the instruction for detecting and correcting errors in the given texts, and $y_{i}$ represents the corrected output texts. 
    We adopt vLLM to enable faster inference through quantization and efficient memory management of attention keys and values using PagedAttention, among other optimizations.
    The prompt templates used for processing both both languages are provided in Appendix \ref{ocrpost}.
\end{enumerate}
By incorporating these strategies, we can significantly improve the quality and coherence of OCR-ed texts, mitigating common errors and enhancing the overall readability and usability of extracted content.
We use FastDeploy\footnote{\url{https://github.com/PaddlePaddle/FastDeploy}}, a highly efficient AI inference deployment tool, as the codebase of our implementation, 
which can fully exploit the advantages of multithreading to optimize inference speed and computational overhead. 
Overall, while maintaining performance and deployment efficiency, we provide a framework for document conversion that covers comprehensive scenarios, including recognizing layout information, supporting table reconstruction, and formula recognition.

\subsection{High-Quality Supplement Data Collection}\label{sec:crawl_pipeline}
In this section, we present our method for High-Quality Supplement Data Collection, which applies to a diverse range of topics and enhances the robustness of datasets.
Inspired by \cite{shao2024deepseekmath}, 
which adopts an iterative pipeline to facilitate the acquisition of large-scale, high-quality data from Common Crawl,
we propose to select high-quality data for mathematics, scientific exam synthetic data, and wiki-based content in our Matrix.

The procedural phases of the iterative pipeline are enumerated as follows:
\begin{itemize}
    \item \textbf{Seed Dataset Collection:} Collect a high-quality seed dataset for the field of interest, like mathematics, code, or wiki-based content.
    \item \textbf{Domain Definition and Sampling:} 
    Define a domain as data entries within the seed dataset sharing the same base URL and extract samples from each domain in the seed dataset as positive samples to enhance format diversity. Correspondingly, acquire an equal amount of data from Common Crawl as negative samples.
    \item \textbf{Model Training:} Employ a FastText model 
 \citep{Joulin_Grave_Bojanowski_Douze_Jégou_Mikolov_2016} for binary classification to discern data relevance to the specified field. Training parameters are set as follows: three epochs, a learning rate of 0.1, an embedding dimension of 256, and an n-gram of 3. The model is quantized to augment operational efficiency within constrained memory capacities, reducing its size to approximately 10\% of its original footprint.
    
    \item \textbf{Data Confidence Assessment:} Utilize the trained FastText model to estimate the confidence of Common Crawl data qualifying as positive. Retain data sequenced from highest to lowest confidence. To streamline the confidence sorting process, initially sample a subset of data to establish a viable threshold that balances data exclusion with retention needs.
    
    \item \textbf{Data Evaluation:} Assess the retained data via ChatGPT 3.5 \cite{achiam2023gpt}, employing the URL to determine field specificity. This stage aims to mitigate the incidence of false positives while maintaining a requisite recall rate.
    
    \item \textbf{Data Recall and Annotation:} Revisit domains where over 10\% of the data was recognized as field-specific. Annotate this data subset using ChatGPT 3.5 \cite{achiam2023gpt} via URL.
    
    \item \textbf{Model Refinement and Iteration:} Integrate unconfirmed positive data from prior iterations into the positive samples to diversify the FastText model’s training base. Subsequently, initiate a new iteration cycle beginning from the training stage.
\end{itemize}

The data selection for Common Crawl focused on the English content of the RedPajama V2 dataset \citep{together2023redpajama}. The seed dataset for the mathematics segment is sourced from OpenWebMath \citep{Azerbayev_Schoelkopf_Paster_Santos_Mcaleer_Jiang_Deng_Biderman_Welleck}, while the science synthetic dataset is from specific domains such as Chemrxiv, biorxiv, and proprietary crawled exercise data from open-source datasets, e.g. wanjuan-exam~\cite{he2023wanjuan}, WebInstruct~\cite{yue2024mammoth2}, Web Of Science~\cite{kowsari2017HDLTex}. Wiki data is procured directly from wiki websites.

\section{Model}
\subsection{Model Architecture}
\label{sec:model architecture}

The MAP-Neo model architecture is grounded on the transformer decoder as outlined by \citet{vaswani2023attention}. The essential parameters defining this architecture are detailed in Table \ref{tab:model_details}. The models are trained with a context length of 8192 tokens, incorporating several enhancements proposed after the original transformer concept. These enhancements are listed below:

\textbf{Multi-Query Attention} \cite{shazeer2019fast}. The 7B model variant employs multi-head attention, whereas the 2B model checkpoints implement multi-query attention, using a single key-value head configuration (\texttt{num\_kv\_heads} $= 1$). This modification is based on ablation studies indicating that multi-query attention is particularly effective at more minor scales \cite{shazeer2019fast}.

\textbf{RoPE Embeddings} \cite{su2023roformer}. Instead of traditional absolute positional embeddings, we utilize rotary positional embeddings at each layer and share these embeddings between the inputs and outputs, minimizing the overall model size.

\textbf{RMSNorm}. To ensure stable training, each transformer sub-layer---including both the attention and feedforward layers---is normalized using RMSNorm \cite{zhang2019root}.

\textbf{Activation Function} We use SwiGLU \cite{shazeer2020glu} as our activation function.

\subsection{Model Scale Hyperparameters}

In this work, we compare two different model scales: 2B and 7B parameters.
Since these models are standard dense Transformers.
These models are constructed using the hyperparameters in Table \ref{tab:model_details}.
The two models are trained identically (except for training data) using the same vocabulary and batch size.
Training details are shown in \S\ref{sec:tokenizer} and  \S\ref{sec:model architecture}.

\begin{table}[h]
\centering
\small{
\caption{Model architecture details. We list the number of layers, \( d_{\text{model}} \), the number of attention heads, and attention head size. The feed-forward size \( d_{\text{ff}} \) is always \( 8 \times d_{\text{model}} \).}
\vspace{0.1cm}
\begin{tabular}{lccccc}
\toprule
Model       & \# Layers & \# Heads & \( d_{\text{model}} \) & \# Feedforward dims & \# KV heads  \\ 
\midrule
MAP-Neo 2B     & 18     & 8          & 2048                   & 16384                           & 1 \\
MAP-Neo 7B   & 28    &  16          & 3072                  & 24576 & 16                        \\
\bottomrule
\end{tabular}
\label{tab:model_details}}
\end{table}

\begin{table}[h]
\centering
\small{
\caption{Model training details.}
\begin{tabular}{lcccc}
\toprule
\textbf{Phases}       & Learning Rate & Weight Decay & Warmup Ratio & Batchsize \\ 
\midrule
 \textbf{Pre-training} (Fundamental Phase)    & 2e-4     & 0.1          & 0.0055                   & 1024                       \\
  \textbf{Pre-training} (Decay Phase)    & 2e-4     & 0.1          & 0                  & 1024                       \\
 \textbf{SFT} (Fundamental Phase)    & 2e-5     & 0          & 0.05                   & 512                       \\
  \textbf{SFT} (Chat Phase)    & 2e-5     & 0          & 0.05                   & 512                      \\
  \textbf{Iterative DPO}   & 5e-6     & 0          & 0.1                   & 256                      \\

\bottomrule
\end{tabular}
\label{tab:training_details}}
\end{table}

\section{Pre-training}
\label{sec:pretraining}
\begin{figure}[!h]
\centering
\includegraphics[width=0.99\linewidth]{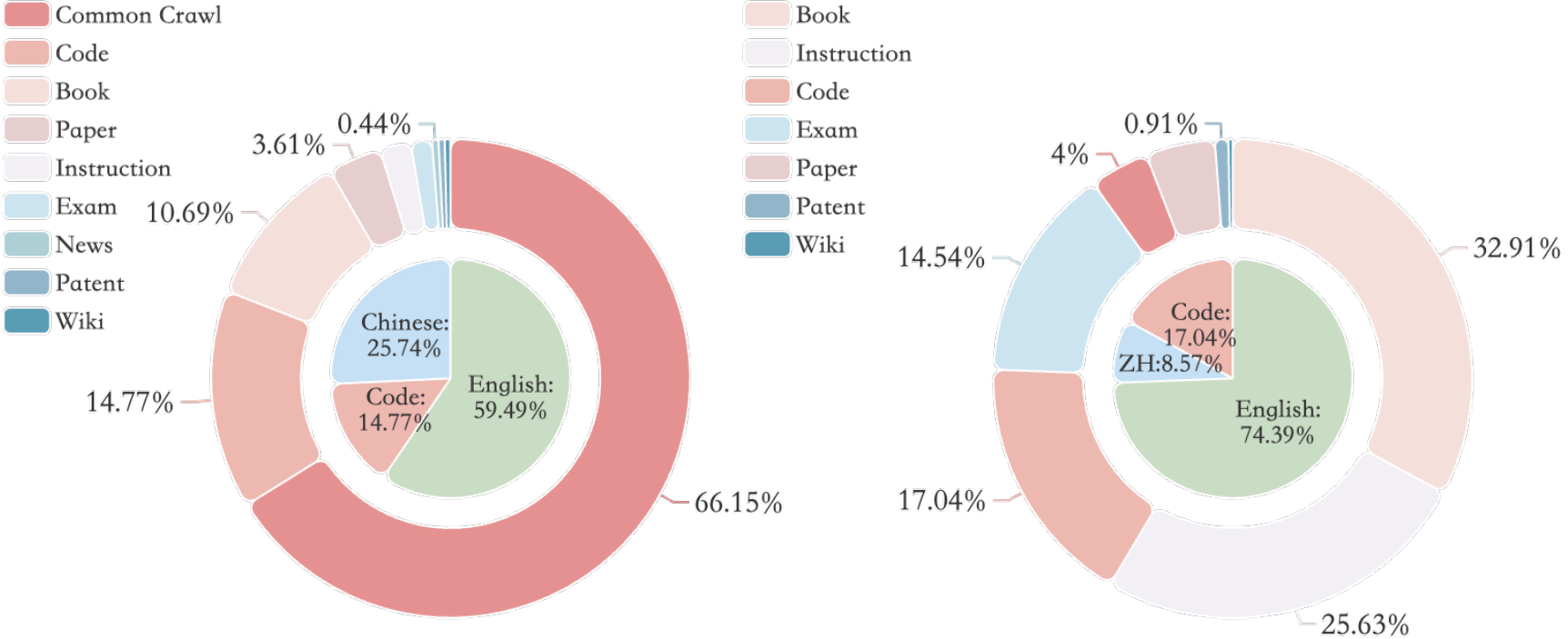}
\caption{
The data mixture ratios in \neo pre-training stage.  
The \textbf{left} is the fundamental phase and the \textbf{right} shows the decay phase.
}
\label{fig:pt_data_ratio}
\end{figure}

In the pre-training process, we employ a two-stage pre-training strategy to train the MAP-Neo model. The first stage termed the fundamental phase, involves training the model on a vast corpus of generic texts to develop its general text generation capability. Subsequently, during the decay phase, we focus on enhancing the reliability of the model's generated content by incorporating high-quality data and mode code data. The distribution of data used across different phases is depicted in Figure~\ref{fig:pt_data_ratio}.
Note that we increase the volume of code data in the decay phase. 
Specifically, during the fundamental phase, since Stack V2~\cite{lozhkov2024starcoder} was not yet available, we utilized Stack V1~\cite{Kocetkov2022TheStack} and repeated the dataset twice to achieve a balanced data ratio. In the decay phase, with the release of Stack V2~\cite{lozhkov2024starcoder}, we incorporated it as the code component for training. 
Moreover, we perform further data distribution tuning including duplicated high-quality data sources, such as books, judicial decisions, and government reports for training, to improve the model's performance.  
The open-source data used for pre-training is shown in Table~\ref{tab:open-source-dataset-list}, the data repetition details are shown in Table~\ref{tab:dataset_round_info} and the training hyperparameters are shown in Table~\ref{tab:training_details}.

\subsection{Fundamental Phase: General Ability Acquisition}
During the fundamental phase, we employ a two-stage learning rate scheduler (LRS) to equip the model with a robust capability for general text generation. 
The LRS is modeled as a piecewise function, consisting of an initial warmup phase where the learning rate linearly ascends from a base rate of \(\eta_{\text{a}} = 2 \times 10^{-5}\) to peak learning rate \(\eta_{\text{max}}= 2\times 10^{-4}\) over $t_{\text{warmup}} = 2k $ steps. This is followed by a cosine decay phase, during which the rate gradually diminishes back to \(\eta_{\text{b}} = 2 \times 10^{-5} \) over about $365k$ steps. 
The learning rate $f(t)$ as a function of time $t$ can be delineated as follows:

\begin{equation}
f(t) = \begin{cases} 
\eta_{\text{a}} + \left(\eta_{\text{max}} - \eta_{\text{a}}\right) \frac{t}{t_{\text{warmup}}} & \text{if } t \leq t_{\text{warmup}} \\
\eta_{\text{b}} + \left(\eta_{\text{max}} - \eta_{\text{b}}\right) \left[\frac{1}{2} \left(1 + \cos\left(\pi \frac{t - t_{\text{warmup}}}{t_{\text{total}} - t_{\text{warmup}}}\right)\right)\right] & \text{if } t_{\text{warmup}} < t \leq t_{\text{total}}
\end{cases},
\end{equation}
where $t$ is the current timestep, $t_{\text{warmup}}$ denotes the duration of the warmup phase, and $t_{\text{total}}$ represents the total number of training timesteps. 
This learning phase processes about $3,726$ billion tokens, ensuring the model's robust training on diverse textual data. This meticulous configuration of learning rates and extensive processing optimize training dynamics and efficiency, fostering a steady maturation of the model’s capabilities.

\subsection{Decay Phase: Improvement and Rectification}
\label{sec:decay_phase}
Owing to the issue in training tokenizer as claimed in \S \ref{sec:tokenizer}, the model encounters test failures in code generation tasks, despite its strong language understanding capabilities acquired during the fundamental phase. To address this issue, we have introduced an additional decay phase specifically designed to utilize a tokenizer of the fixed version. 
The learning rate in this decay phase initiates at \(\eta_{\text{c}} = 2 \times 10^{-4}\) and undergoes exponential decay over $t_{\text{decay}} = 148k $ steps, with a half-life $T$ corresponding to half the $t_{\text{decay}}$ steps,  similar to the decay phase employed by MiniCPM~\cite{hu2024minicpm}, which can be formulated as follows:
\begin{equation}
f(t) = \eta_{\text{c}} \times 0.5^{\frac{t}{T}} \quad \text{if } t \leq t_{\text{delay}},
\end{equation}
where $t$ is the current timestep of the decay phase. 
This strategic adjustment not only rectifies the initial tokenization flaws but also enhances the model’s performance on code generation tasks. During this phase, the model processes a total of about $778$ billion tokens, which primarily consist of high-quality instruction data. We also simultaneously increased the proportion of code in the data from $14.77\%$ to $17.04\%$. 
This adjustment significantly enhances the overall performance of the model. The deliberate enrichment of the dataset with a higher ratio of code, coupled with instructional inputs, ensures a more robust and versatile model, adept at tackling complex coding tasks as well as understanding and generating professional responses in different fields. 

\section{Alignment}
\subsection{Supervised Fine-tuning}
To align with the human behavior of LLMs, the initial step is to perform Supervised Fine-Tuning (SFT).
Our SFT also consists of two phases. In the first phase, we collect a large amount of instruction data to enhance the foundational abilities of LLMs. In the second phase, we build upon the capabilities established in the first phase and propose to improve the chat abilities of MAP-Neo. This process finetunes a pre-trained LLM on chat-style data, including both queries and responses. 
We illustrate the details of data construction and training strategies.

\subsubsection{Data}

\paragraph{Foundational Phase: Enhancing Instruction Following Abilities}
In the first phase, our focus is to significantly boost the model's foundational abilities (e.g., code and math skills),
where we utilize over 2 million instructional data points during this phase. 
Specifically,
the first phase includes the entire OpenHermes 2.5~\citep{OpenHermes2.5}, 
where we exclude segments related to the TheoremQA benchmark~\cite{chen2023theoremqa} to prevent benchmark data leakage. Additionally, we incorporate the complete Code-Feedback ~\citep{zheng2024opencodeinterpreter} dataset and a subset of WebInstructSub~\citep{yue2024mammoth2} data.

\paragraph{Chat Phase: Enhancing Chat Abilities}
In the second phase, we focus on improving the model's chat abilities while maintaining the foundational skills acquired in the first phase. For this purpose, we collect over 100k multi-turn dialogue data sourced from real user conversations. To ensure the model retains its foundational capabilities, we include 5k math and code-related data points extracted from the first phase. Our experiments have demonstrated that this additional phase of SFT significantly boosts the model's performance on chat benchmarks, such as MT-Bench~\citep{zheng2024judging} and AlpacaEval~\citep{alpaca_eval}, without compromising its foundational abilities.

By following this two-phase approach, we ensure that our model can not only maintain a strong foundation in essential skills but also generate natural, helpful, and contextually accurate responses.

\subsubsection{Training}

Consistent with pre-training, we also apply the next-token prediction objective as the training task for SFT. Note that we apply the loss masks for the system and user inputs. The model’s training process utilizes the AdamW optimizer with the hyperparameters in Table~\ref{tab:training_details}.

The sequence length is limited to 8192, and the batch size is 512. The training process consists of two phases using the same hyperparameters. In the first phase, the model is trained for 3 epochs using over 2 million instructional data points, focusing on enhancing foundational abilities. In the second phase, the model is trained for 1 epoch using over 100k multi-turn dialogue data to enhance its chat abilities while maintaining the foundational skills acquired in the first phase.

\subsection{Iterative DPO}
\paragraph{DPO}
Direct Preference Optimization (DPO)~\citep{rafailov2023direct} is a straightforward and effective method for aligning language models with human feedback. It converts the preference loss \citep{btmodel} into a loss function over the language model, thereby bypassing the need for explicit reward modeling~\citep{btmodel} and reinforcement learning~\citep{christiano2017deep,schulman2017proximal}. Starting with a supervised fine-tuned language model, denoted as $\pisft$, DPO collects a dataset $\mathcal{D} = \{(x,y_w,y_l)^{i}\}$, which consists of human preferences between two responses generated by $\pisft$: $y_w$ (preferred) and $y_l$ (dispreferred) to the same prompt $x$. Using this dataset, DPO parameterizes a language model $\pi_\theta$ and directly estimates its parameters through maximum likelihood estimation on the human preference dataset $\mathcal{D}$ as follows:

\begin{equation}\label{eq:dpo}
    \mathcal{L}_{\text{DPO}}(\pi_\theta; \pisft, \mathcal{D}) = - \E_{(x, y_w, y_l) \sim \mathcal{D}}
\left[\log 
    \sigma
    \left(
        \beta \log\frac{\pi_\theta(y_w|x)}{\pisft(y_w|x)}
        - 
        \beta \log\frac{\pi_\theta(y_l|x)}{\pisft(y_l|x)}
    \right)
\right].
\end{equation}

\paragraph{Iterative DPO.}
We follow Storm-7B~\citep{liu2024iterative} to use the Iterative DPO~\citep{xu2023some} pipeline to develop our chat model. Specifically, we employ three iterations, with each iteration consisting of three stages: 1) generating paired responses, 2) labeling responses using reward models, and 3) training the LLM with DPO loss as described in Eq.~\ref{eq:dpo}. We utilize Nectar\footnote{\url{https://huggingface.co/datasets/berkeley-nest/Nectar}} as our prompt dataset and Starling-RM-34B\footnote{\url{https://huggingface.co/Nexusflow/Starling-RM-34B}}~\citep{starling2023} as our reward model. This model is finetuned from Yi-34B-Chat~\citep{young2024yi} and generates a scalar output for any given prompt and response. To preserve the multilingual capabilities of our model, we also adopt a preference dataset\footnote{\url{https://huggingface.co/datasets/llm-wizard/alpaca-gpt4-data-zh}} in Chinese in the 3-rd iteration.

We report the length-controlled win rate of AlpacaEval2.0~\citep{dubois2024length} to demonstrate the performance progress of our model in Table~\ref{tab:idpo}. The results show that performance improves with each iteration, indicating that our model becomes increasingly aligned with human values.

\begin{table}[h]
\centering
\caption{The length-controlled win rate of MAP-Neo at different iterations on the AlpacaEval2.0 leaderboard. For ``SFT", we report the performance of our model using two-phase SFT.}
\vspace{0.1cm}
\label{tab:idpo}
\begin{tabular}{cccccc}
\toprule
Model & SFT & Iteration 1  & Iteration 2 & Iteration 3 \\ 
\midrule
LC Win Rate (\%) &9.77 & 10.02 & 15.59 &  16.65 \\ 
\bottomrule
\end{tabular}
\end{table}

\section{Scaling Law of MAP-Neo}
\subsection{Problem Definition}
The scaling laws are capable of predicting training configuration for the training of LLMs. This principle emphasizes the importance of the ratio between the amount of training data $D$ (measured in tokens) and the size of the model $N$ (in terms of parameters). 
In this section, we applied the Chinchilla Law in Eq.~\ref{eq-chinchilla} \cite{hoffmann2022training}, OpenAI Law in Eq.~\ref{eq-openai} \cite{kaplan2020scaling}, a derivation of Symbolic Music Scaling law in Eq.~\ref{eq-1/ND} \cite{qu2024mupt} and our proposed method on our dataset to fit our models, where $A$, $B$, $E$, $\alpha$, $\beta$, $\alpha_c$, $D_c$, $\alpha_N$, $N_c$ and $d$ are hyperparameters to be optimized.
\begin{equation}\label{eq-chinchilla}
    L(N, D) = \frac{A}{N^\alpha} + \frac{B}{{D}^\beta} + E
\end{equation}
\begin{equation}\label{eq-openai}
    L(N, D) = \left(\left(\frac{N_c}{N}\right)^\frac{\alpha_N}{\alpha_D} + \frac{D_c}{{D}} \right)^{\alpha_D}
\end{equation}
\begin{equation}\label{eq-1/ND}
L(N, D) = \frac{d}{N^\alpha \cdot D^\beta} + \frac{A}{N^\alpha} + \frac{B}{D^\beta} + E.
\end{equation}
The original SMS scaling law introduces two modifications to the Chinchilla law. The first modification addresses the repetition of training data, which is not considered in our study. The second modification concerns the interaction between the number of model parameters, $N$, and the dataset size, $D$. Specifically, it posits that the loss curve as a function of $D$, represented as $\frac{B}{D^\beta}$, is influenced by $N$. 
This interaction between the number of model parameters and dataset size is also reflected in the OpenAI scaling law. However, our version of SMS law, as detailed in Eq. \ref{eq-1/ND}, is simpler and yields superior results compared to the corresponding model in the OpenAI framework.

The motivation for fitting scaling laws is to optimize the loss under the bounds of computational resources. This process is formalized as minimizing the validation cross-entropy loss $L$, subject to constraints imposed by available computational resources ($C$), specifically floating-point operations per second (FLOPs), as denoted below:

\begin{equation}
\arg\min_{N,D} L(N, D) \quad \text{s.t.} \quad \text{FLOPs}(N, D) = C
\end{equation}

Given that our model is trained on almost non-repetitive and high-quality data, we utilize the training loss instead of the validation loss for the scaling law application.

\subsection{NEO Scaling Law}
We train models with sizes of 250M, 460M, and 980M parameters using 1000B tokens of training data. These models are then used to predict the scaling law, which guides the training of a model with 7.8B parameters on 3.07T (3065B) tokens during phase 1.
To evaluate the fit of the scaling law, we employ the Huber loss (
$\delta=1e-3$) between the actual log\-loss and the predicted log\-loss, along with the $R^2$ value between the true loss and predicted loss. Optimization of the scaling law is performed using the L\-BFGS algorithm. This approach is applied consistently across the Chinchilla law and the symbolic music scaling law.
By leveraging these methods, we aim to ensure the accuracy and reliability of our scaling law predictions, enabling efficient training of large-scale language models.

\begin{figure}
    \centering
    \includegraphics[width=1\textwidth]{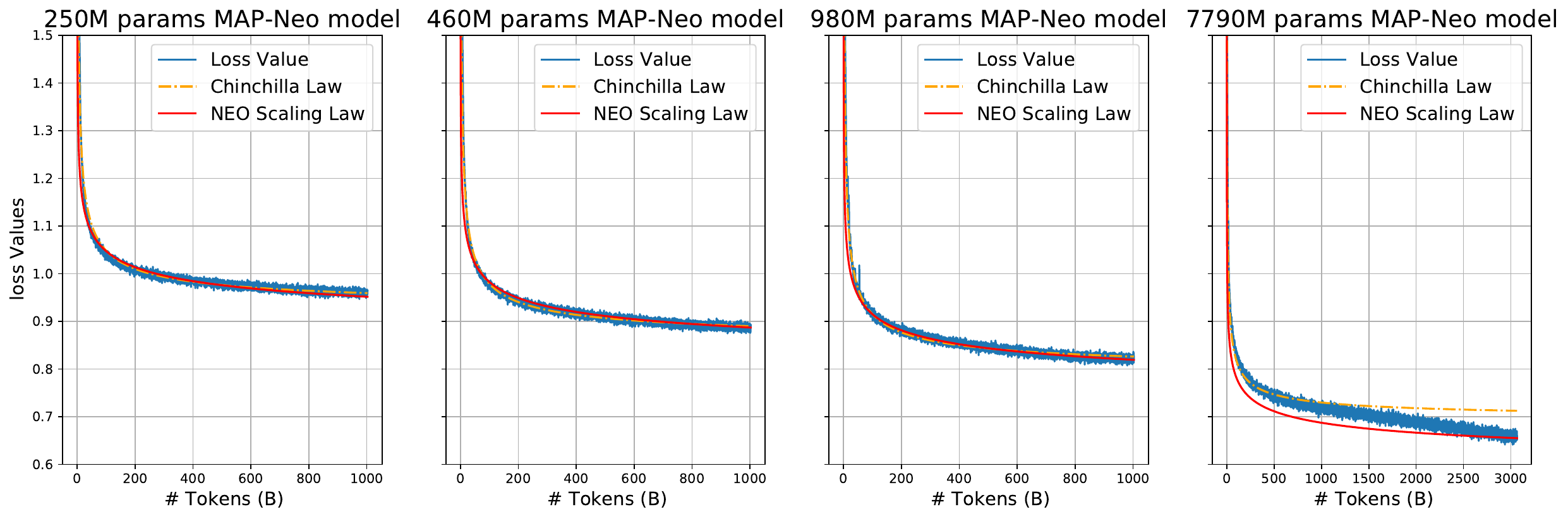}
    \caption{The training loss value is represented by the blue line. The Chinchilla law prediction is shown in yellow, and the NEO scaling law prediction is depicted in green. We fit the Chinchilla law and NEO law on 250M, 460M, and 980M and predict the model behavior on both training samples and samples from the 7B model.}
    \label{fig:NEO-scaling}
\end{figure}

Figure \ref{fig:NEO-scaling} illustrates the training loss values alongside the Chinchilla law predictions. Although the Chinchilla law fits well, with the predicted loss curve falling within the fluctuations of the actual loss curve, its trend appears flatter compared to the actual loss curve. The actual loss decreases more rapidly than predicted by the Chinchilla formula (i.e. $\frac{B}{D^\beta}$), suggesting our dataset with diverse high-quality corpora can further decrease the loss value when $D$ is large. 
To address this discrepancy between Chinchilla prediction and observation, we introduce the following equation, denoted as NEO scaling law, which includes one additional regularization term $log(D)$ for datasets containing several trillion tokens across various corpora:
\begin{equation}
    L(N, D) = \frac{A}{N^\alpha} + \frac{B}{{D}^\beta} + E -d\cdot log(D)
\end{equation}

Note that although the regularization term $-d\cdot \log(D)$ theoretically results in no lower bound on loss as $D$ approaches negative infinity suggesting potential imperfection of the formula, the value of $d$ typically ranges in our experiments between 1e-2 and 3e-2. Therefore, for a dataset size less than hundreds of trillion tokens, the loss remains within a reasonable range.

From the following Table \ref{tab: NEO scaling}, we observe that the NEO scaling law equation yields significantly better results on the training set and testing set.

\begin{table}[htbp]
\centering
\caption{Comparison of parametric fitting on $R^2$ and Huber Loss of different scaling laws.}\label{tab: NEO scaling}
\resizebox{0.95\columnwidth}{!}{
\begin{tabular}{c|cccc}
\toprule
\textbf{Paramatic fit} &  \textbf{$R^2$ Value (train) ↑} & \textbf{Huber Loss (train) ↓} &  \textbf{$R^2$ Value (test) ↑} & \textbf{Huber Loss (test) ↓} \\
\midrule
    Chinchilla Law & 0.2483 &\textbf{ 0.1665 } &  0.4308 & \textbf{0.3372 }\\
    OpenAI Law & 0.2268  & 1.0424 &  -0.2916 & 0.6023\\
    SMS Law & 0.2484 &\textbf{ 0.1665} & 0.4306 & 0.3375 \\
    NEO Scaling Law & \textbf{0.7361} & 0.2961 & \textbf{0.6720 } & 0.2081\\
\bottomrule 
\end{tabular}
}

\end{table}

Under the prediction of the NEO scaling law and the computational resource constraint of $1.5\times 10^{23}$ FLOPs, the optimal configuration is to train a 10B parameter model with 2.5T tokens, providing a predicted loss value of 0.6597. To ensure comparability with baseline models, we choose to keep our model size at 7.8B parameters, similar to the Llama-base model.
This configuration with a 7.8B parameter model with 3.07T tokens requires slightly fewer computational resources and results in a similar prediction loss value (0.6618).
Meanwhile,
after training,
We observe that the real training loss in this configuration is 0.6591, which is close to the prediction loss value and demonstrates the effectiveness of the NEO scaling law.

\subsection{Generalization of NEO Scaling Law}

\begin{figure}[htbp]
    \centering
    \includegraphics[width=1.0\textwidth]{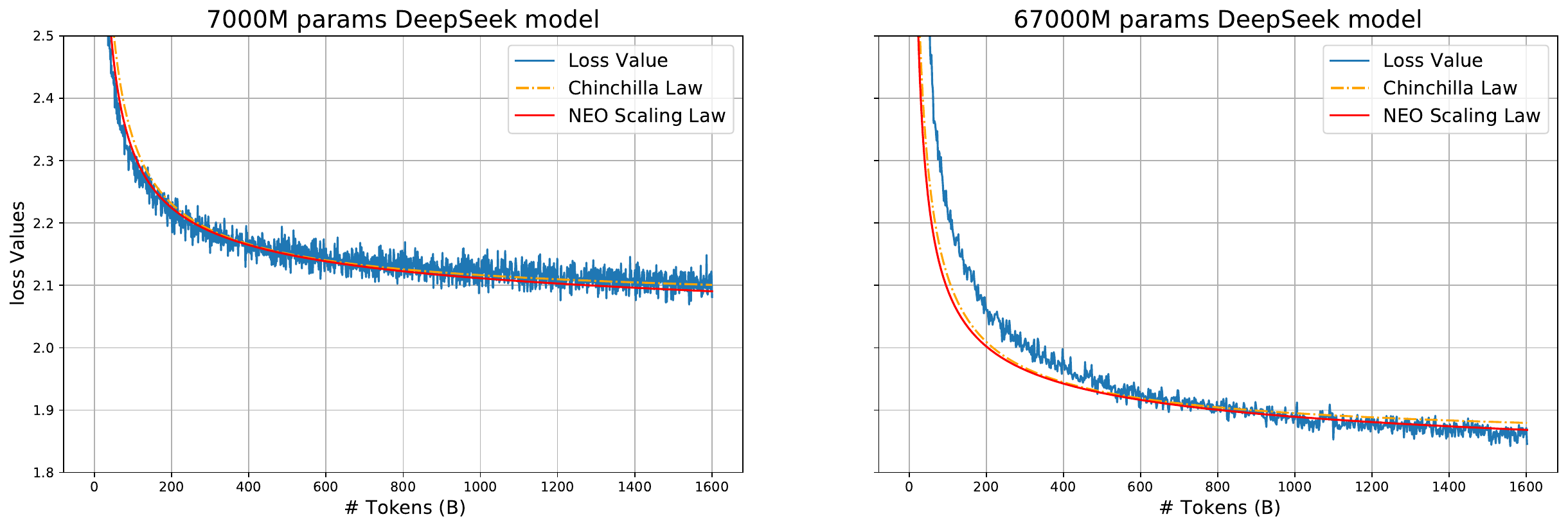}
    \caption{The loss curve of Chinchilla Law prediction and the NEO Scaling law prediction for the DeepSeek LLM. We use loss values from both 7B and 67B for fitting and prediction.}
    \label{fig:deepseek}
\end{figure}

The NEO scaling law can be applicable to a broader range of models beyond MAP-Neo. Specifically,
in Figure \ref{fig:deepseek}, we illustrate the fit results of the Chinchilla scaling law (yellow dashed line) and the NEO scaling law (red solid line) to the DeepSeek LLM~\cite{deepseek-llm} with the 7B and 67B parameters, which also pre-trained on a dataset with multiple corpura including Chinese, English and codes.

We observe that for the largest model sizes (i.e.  MAP-Neo-7B and DeepSeek-67B),
the predictions of Chinchilla Law tend to underestimate the actual loss when the dataset size ($D$) is small and overestimate the actual loss as model parameters and training data scale up.
In contrast,
our predictions of our NEO Scaling Law produce better fitting results when compared with the results of Chinchilla Law for MAP-Neo-7B and DeepSeek-67B.

We further suggest NEO Scaling law might be more suitable for the situation with a large diverse pre-training dataset with multiple high-quality dataset sources. For more discussion on NEO scaling law on other models, please refer to Appendix \ref{scaling-appendix}.

\section{Infrastructure}
Our advanced infrastructure consists of two primary components: a data processing system and a training system. The training system is designed to support both pre-training and fine-tuning stages, enabling comprehensive model development.

Our infrastructure is designed to handle extensive data processing tasks for both English and Chinese datasets. We utilize robust systems to ensure efficient and scalable processing capabilities across different languages. Spark~\citep{zaharia2012resilient} is used for distributed computing, and object storage is used to save the data. Each machine is configured with a 64-core CPU, 256GB of memory, and 1TB of local disk. There are a total of 94 machines.
For the Chinese data processing, there are a total of 14 machines. Among them, 6 machines have a 96-core CPU and 180GB of memory, while the other 8 machines have a 48-core CPU and 190GB of memory. Network File System (NFS)\citep{sandberg1985design} is used as the distributed file storage system.

In the pre-training stage, the Megatron-Core toolkit is utilized for its capacity to train large-scale language models, featuring up to hundreds of billions of parameters. Compared to the tokens per second (TPS) metric, the usage of     Megatron-core achieves a rate of 7200 TPS when training a 7B model, which surpasses the performance of 6400 TPS observed under the same settings without employing Megatron-core. This is accomplished using both model and data parallelism techniques.
We implement several strategies to manage our large datasets and model complexities effectively. 
Firstly, we introduce programs to identify and temporarily remove tainted computing nodes from the resource pool due to software or hardware errors by automatic inspection, prediction, and labeling. 
Secondly, we make modifications to Megatron-LM to specifically prevent overflow issues detailed in \ref{sec_appendix:overflow} when processing large data corpora. 
Lastly, we implement task recovery mechanisms that utilize strategically selected checkpoint iterations to safeguard against potential failures during training. These enhancements ensure optimal performance and reliability in our large-scale training operations.

To ensure optimal utilization of our computational resources, our infrastructure design incorporates a sophisticated network topology and hardware configuration, facilitating efficient workload distribution and data transfer for complex model training tasks. Our infrastructure utilizes distributed computing techniques to optimize the training of our models. 
Specifically, our 7B model is trained using an H800 configuration with 512 GPUs across 64 nodes and employs NCCL for backend distribution with \texttt{ibp} as the network interface and \texttt{mlx5} of InfiniBand hardware to enhance inter-GPU communication. Tensor model parallelism is configured to utilize 2 GPUs, distributing the execution of a single transformer module across these units to enhance efficiency. For our 2B models, we utilize all 256 GPUs with tensor model parallelism set to 1 to ensure effective data replication. We further amplify scalability and efficiency by employing techniques similar to ZeRO-1 for sharding the optimizer state. This approach enables the management of more extensive datasets and more complex model training with significantly reduced memory overhead.

Our cluster consists of machines with dual Intel Xeon CPUs and eight NVIDIA H800 GPUs. The architecture facilitates high-speed data transfer, with each CPU socket interfacing with two PCIe Gen4 x16 lanes connected to dedicated PCIe switches. These switches manage the connections to a local NVMe SSD, an RDMA-capable Network Interface Card (NIC), and two GPUs. Inter-CPU communication is facilitated by Intel's Ultra Path Interconnect (UPI), with both CPUs linked to a dual-port TCP NIC supporting 100 Gbps. Each machine's network configuration includes four RDMA NICs, each offering 200 Gbps of full duplex bandwidth and integrated GPU Direct RDMA capabilities. Notably, the GPU array is interconnected through four NVIDIA NVSwitches, enabling robust intra-GPU communication with a bandwidth of 400 Gbps. This advanced configuration underscores the cluster's capability to handle large-scale model training with exceptional efficiency and speed.

Regarding the inter-machine connections of our data center, we implement a dual-layer Clos network architecture wherein each minipod accommodates at least 512 H800 servers interconnected via a high-speed RDMA network. Within this architecture, each S0 switch is equipped with 64 ports, each supporting a bandwidth of 400 Gbps. This arrangement ensures a network convergence ratio of 1:1, a critical factor in maintaining optimal data flow and reducing bottlenecks. Connectivity within this structure is meticulously organized such that every two S0 switches serve 32 servers, with a total of 32 S0 switches networking within each minipod. This setup exemplifies an advanced implementation designed to maximize throughput and minimize latency in data center environments.

\begin{table}[ht]
\centering
\setlength{\tabcolsep}{3pt}
\caption{Performance comparison of various base models on different benchmarks. The best results are in \lightblue{blue}, the second-best results are \underline{underline}, and the third-best results are in \fbox{fbox}.}
\vspace{0.1cm}
\label{tab:base_model_compare}
\resizebox{\textwidth}{!}{
\begin{tabular}{ccccccccccc}
\toprule
                     \textbf{Dataset} & \textbf{LLama3-8B} & \textbf{Mistral-7B} & \textbf{LLama2-7B} & \textbf{Amber-7B} & \textbf{OLMo-7B} & \textbf{OLMo-1.7-7B} & \textbf{Pythia-6.9B} & \textbf{Pythia-12B} & \textbf{MAP-Neo-7B} \\\midrule            
\multicolumn{11}{c}{\textbf{ Standard Benchmarks }}\\\midrule
    \textbf{BoolQ}&66.82&64.1&\underline{70.67}&63.52&68.41&\fbox{70.49}&62.45&61.07&\lightblue{81.07} \\\midrule
    \textbf{PIQA}&\underline{81.12}&\lightblue{81.18}&78.18&76.82&79&\fbox{80.25}&75.52&76.17&76.55 \\\midrule
    \textbf{SIQA}&\fbox{47.34}&47.13&45.50&42.89&44.11&\underline{54.71}&42.32&44.32&\lightblue{68.22} \\\midrule
    \textbf{HellaSwag}&\underline{74.52}&\lightblue{76.49}&71.27&66.76&70.32&\fbox{72.37}&59.6&63.04&70.74 \\\midrule
    \textbf{WinoGrande}&\underline{72.38}&\lightblue{75.3}&\fbox{69.53}&64.64&66.54&69.22&60.85&63.69&59.83 \\\midrule
    \textbf{ARC-c}&\lightblue{79.66}&\underline{71.53}&35.93&24.41&24.07&49.83&22.71&25.08&\fbox{68.14} \\\midrule
    \textbf{OpenBookQA-Fact}&\fbox{69.0}&\underline{81.0}&42.60&26.6&24.6&64.4&25&28.6&\lightblue{82.0} \\\midrule
    \textbf{CommonsensQA}&\underline{69.7}&67.57&66.50&57&60.44&\fbox{69.04}&55.45&54.79&\lightblue{69.94} \\\midrule
    \textbf{MMLU-AVG}&\lightblue{66.52}&\underline{64.04}&46.80&28.07&28.51&53.52&26.39&27.06&\fbox{58.14} \\\midrule
    \textbf{*-humanities}&\lightblue{70.41}&\underline{68.04}&51.47&30.17&25.52&55.03&26.87&27.39&\fbox{60.7} \\\midrule
    \textbf{*-stem}&\lightblue{56.22}&\underline{53.21}&38.02&27.66&28.68&44.17&26.77&28.13&\fbox{49.84} \\\midrule
    \textbf{*-social-science}&\lightblue{76.0}&\underline{73.65}&52.20&27.18&30.05&62.19&24.32&26.26&\fbox{66.78} \\\midrule
    \textbf{*-other}&\lightblue{68.94}&\underline{67.0}&49.99&27.37&29.86&57.67&27.25&25.91&\fbox{59.73} \\\midrule
\multicolumn{11}{c}{\textbf{ Code Generation } }\\\midrule
    \textbf{Humaneval}&\lightblue{33.5}&\underline{28.0}&13.4&13.4&11.6&17.1&9.1&8.5&\fbox{23.8} \\\midrule
    \textbf{Humaneval-Plus}&\lightblue{29.3}&\underline{23.2}&11.6&12.2&9.8&15.2&8.5&7.3&\fbox{20.1} \\\midrule
    \textbf{MBPP}&\lightblue{61.4}&\underline{46.8}&29.1&22.8&27&32.3&16.1&15.6&\fbox{34.9} \\\midrule
    \textbf{MBPP-Plus}&\lightblue{51.6}&\underline{38.9}&22.8&18.5&21.2&25.7&13.2&11.1&\fbox{29.9} \\\midrule
\multicolumn{11}{c}{\textbf{ World Knowledge } }\\\midrule
    \textbf{NQ}&\lightblue{10.14}&\fbox{9.31}&5.07&3.1&0.66&1.02&0.86&1.83&\underline{9.97} \\\midrule
    \textbf{TriviaQA}&\fbox{51.94}&\lightblue{56.47}&\underline{52.44}&26.65&31.97&45.16&16.97&24.31&42.36 \\\midrule
\multicolumn{11}{c}{\textbf{ Reading Comprehension } }\\\midrule
    \textbf{SQuAD2.0}&\underline{40.88}&12.53&\lightblue{41.32}&\fbox{31.15}&27.05&30.43&22.54&23.11&30.98 \\\midrule
\multicolumn{11}{c}{\textbf{ Exams } }\\\midrule
    \textbf{MATH}&\lightblue{20.76}&\fbox{15.74}&6.14&3.88&1.6&4.86&3.82&4.54&\underline{20.7} \\\midrule
    \textbf{GSM8K}&\lightblue{54.74}&\fbox{47.46}&16.22&3.64&5.84&28.43&3.41&3.94&\underline{53.68} \\\midrule
\multicolumn{11}{c}{\textbf{ Chinese } }\\\midrule
    \textbf{C-EVAL-AVG}&\underline{49.83}&\fbox{47.54}&32.37&23.82&27.39&35.21&24.64&24.82&\lightblue{57.68} \\\midrule
    \textbf{*-stem}&\underline{45.26}&\fbox{44.74}&28.28&22.36&25.75&32.36&23.94&27.27&\lightblue{50.35} \\\midrule
    \textbf{*-social-science}&\underline{58.09}&\fbox{54.8}&39.22&25.95&31.87&40.43&26.34&23.78&\lightblue{70.23} \\\midrule
    \textbf{*-humanities}&\fbox{50.6}&\underline{51.52}&37.11&21.19&26.29&35.5&21.7&20.05&\lightblue{63.49} \\\midrule
    \textbf{*-other}&\underline{49.84}&\fbox{42.06}&28.84&27.16&27.4&35.36&27.28&26.08&\lightblue{53.78} \\\midrule
    \textbf{*-hard}&\fbox{32.41}&\underline{33.97}&25.21&19.63&27.12&29.16&22.99&27.05&\lightblue{41.07} \\\midrule
    \textbf{CMMLU-AVG}&\underline{50.72}&\fbox{44.63}&31.85&25.77&25.53&36.74&25.34&24.88&\lightblue{55.1} \\\midrule
    \textbf{*-humanities}&\underline{53.1}&\fbox{44.59}&32.50&24.86&26.65&37.04&25.81&25.41&\lightblue{62.24} \\\midrule
    \textbf{*-stem}&\underline{43.59}&\fbox{37.82}&29.05&25.61&25.24&31.94&24.29&23.7&\lightblue{45.62} \\\midrule
    \textbf{*-social-science}&\underline{52.59}&\fbox{46.37}&32.60&25.83&25.17&38.14&25.78&25.17&\lightblue{59.39} \\\midrule
    \textbf{*-other}&\lightblue{53.98}&\fbox{49.83}&33.35&26.65&25.43&39.88&25.47&25.33&\underline{53.39} \\\midrule
    \textbf{*-china-specific}&\underline{44.81}&\fbox{40.84}&29.27&24.96&24.97&34.91&26.5&25.34&\lightblue{55.84} \\
\bottomrule
\end{tabular}}

\end{table}

\begin{table}[htbp]
\centering
\caption{Performance comparison of various aligned models on different benchmarks. The best results are in \lightblue{blue}, the second-best results are \underline{underline}, and the third-best results are in \fbox{fbox}.}
\vspace{0.1cm}
\label{tab:aligned_model_compare}
\resizebox{0.98\textwidth}{!}
{
\begin{tabular}{ccccccccc}
\toprule


\multirow{2}{*}{{\textbf{Dataset}}} & \textbf{LLama-3-8B} & \textbf{Mistral-7B} & \textbf{LLama-2-7B} & \textbf{Amber-7B} & \textbf{OLMo-7B} & \textbf{MAP-Neo-7B}\ & \textbf{MAP-Neo-7B} \\
& \textbf{(Instruct)} & \textbf{(Instruct-v0.2)}& \textbf{(Chat)} & \textbf{(Chat)} &  \textbf{(Instruct)} & \textbf{(SFT)} & \textbf{(Instruct)} \\


\midrule
\multicolumn{9}{c}{\textbf{Chat Benchmarks}} \\
\midrule
\textbf{AlignBench} & \lightblue{6.17} & \underline{5.27} & 4.33 & 2.85 & 3.2 & 4.63 & \fbox{5.04} \\
\midrule
\textbf{AlpacaEval} & \lightblue{22.9} & \underline{17.1} & 5.4 & 1.21 & 3.64 & 9.77 & \fbox{16.65} \\
\midrule
\textbf{Arena-Hard} & \lightblue{20.6} & \underline{12.6} & 4.6 & 1.2 & 1.7 & 10 & \fbox{11.5} \\
\midrule
\textbf{CHC-Bench} & 5.53 & \underline{6.86} & 4.7 & 3.13 & 3.91 & \fbox{6.14} & \lightblue{7.42} \\
\midrule
\textbf{MT-Bench} & \lightblue{8.1} & \underline{7.5} & 6.6 & 5.2 & 5.3 & \fbox{7.1} & \fbox{7.1} \\
\midrule
\multicolumn{9}{c}{\textbf{Standard Benchmarks}} \\
\midrule
\textbf{BoolQ} & 75.05 & \underline{82.87} & 74.77 & 66.51 & 72.2 & \lightblue{84.59} & \fbox{81.28} \\
\midrule
\textbf{PIQA} & \underline{80.09} & \lightblue{82.43} & 76.01 & \fbox{77.48} & 75.3 & 76.06 & 75.24 \\
\midrule
\textbf{SIQA} & \fbox{51.23} & 50.41 & 48.72 & 44.88 & 48.41 & \underline{51.69} & \lightblue{52.25} \\
\midrule
\textbf{HellaSwag} & \fbox{71.39} & \lightblue{80.11} & 71.32 & 67.84 & \underline{75.18} & 68.5 & 68.7 \\
\midrule
\textbf{WinoGrande} & \underline{71.9} & \lightblue{73.4} & \fbox{68.35} & 64.96 & 66.69 & 65.19 & 66.06 \\
\midrule
\textbf{ARC-c} & \lightblue{81.36} & \fbox{73.56} & 55.59 & 37.29 & 57.63 & \underline{80} & \underline{80} \\
\midrule
\textbf{OpenBookQA-Fact} & \lightblue{87} & \underline{85.4} & \fbox{74.4} & 36.6 & 74 & \underline{85.4} & \underline{85.4} \\
\midrule
\textbf{CommonsenseQA} & \underline{73.55} & \lightblue{75.84} & 70.11 & 60.28 & 63.47 & 68.39 &  \fbox{70.35} \\
\midrule
\textbf{MMLU-Pro} & \lightblue{38.12} & \underline{30.86} & 21.61 & 14.65 & 16.27 & 28.08 & \fbox{28.74} \\
\midrule
\textbf{MMLU} & \lightblue{67.1} & \underline{60.81} & 48.22 & 38.8 & 47.47 & \fbox{58.28} & \fbox{58.28} \\
\midrule
\textbf{*-humanities} & \lightblue{70.67} & \underline{66.58} & 52.71 & 39.19 & 48.33 & 60.4 & \fbox{60.85} \\
\midrule
\textbf{*-stem} & \lightblue{56.97} & 50.01 & 37.98 & 33.78 & 38 & \fbox{51.86} & \underline{52.29} \\
\midrule
\textbf{*-social-science} & \lightblue{76.9} & \underline{69.75} & 55.81 & 42.85 & 56.57 & \fbox{66.19} & 65.6 \\
\midrule
\textbf{*-other} & \lightblue{69.3} & \underline{62.55} & 51.69 & 42.03 & 52.06 & \fbox{58.26} & 57.68 \\
\midrule
\multicolumn{9}{c}{\textbf{Code Generation}} \\
\midrule
\textbf{HumanEval} & \lightblue{48.8} & \fbox{42.1} & 14 & 17.7 & 14.63 & 34.1 & \underline{45.1} \\
\midrule
\textbf{HumanEval-Plus} & \lightblue{44.5} & \fbox{36.0} & 12.2 & 14 & 12.8 & 31.7 & \underline{37.8} \\
\midrule
\textbf{MBPP} & \lightblue{70.1} & \fbox{39.7} & 29.1 & 28.0 & 20.1 & \underline{44.4} & \underline{44.4} \\
\midrule
\textbf{MBPP-Plus} & \lightblue{59.3} & 33.3 & 22.8 & 23.5 & 16.7 & \underline{38.1} & \fbox{36} \\
\midrule
\multicolumn{9}{c}{\textbf{World Knowledge}} \\
\midrule
\textbf{NQ} & \lightblue{8.25} & 1.14 & 1.5 & \fbox{3.02} & 0.53 & \underline{3.8} & 2.41 \\
\midrule
\textbf{Triviaqa} & \lightblue{56.32} & \fbox{45.06} & \underline{46.79} & 30.95 & 27.91 & 38.77 & 27.09 \\
\midrule
\multicolumn{9}{c}{\textbf{Reading Comprehension}} \\
\midrule
\textbf{SQuAD2.0} & \lightblue{66.99} & 15.01 & 19.61 & 13.12 & \fbox{42.13} & \underline{44.58} & 25.2 \\
\midrule
\multicolumn{9}{c}{\textbf{Exams}} \\
\midrule
\textbf{MATH} & \fbox{29.28} & 13.14 & 6.9 & 4.2 & 1.8 & \underline{35.36} & \lightblue{35.58} \\
\midrule
\textbf{GSM8K} & \lightblue{79.23} & 49.2 & 26 & 7.59 & 13.5 & \fbox{72.02} & \underline{73.16} \\
\midrule
\multicolumn{9}{c}{\textbf{Chinese}} \\
\midrule
\textbf{C-Eval} & \fbox{50.76} & 43.72 & 35.67 & 26.29 & 35.18 & \underline{55.42} & \lightblue{56.97} \\
\midrule
\textbf{*-stem} & \underline{47.47} & 41.35 & 32.59 & 23.99 & 31.43 & \fbox{47.37} & \lightblue{49.08} \\
\midrule
\textbf{*-social-science} & \fbox{57.05} & 47.75 & 40.04 & 26.77 & 42.13 & \underline{69.21} & \lightblue{70.75} \\
\midrule
\textbf{*-humanities} & \fbox{48.32} & 47.33 & 36.96 & 28.26 & 34.03 & \lightblue{63.17} & \underline{63.14} \\
\midrule
\textbf{*-other} & \lightblue{53.48} & 40.74 & 36.01 & 28.06 & 36.81 & \fbox{49.78} & \underline{52.63} \\
\midrule
\textbf{*-hard} & \fbox{31.04} & 27.32 & 28.45 & 22.77 & 26.33 & \underline{38.41} & \lightblue{39.55} \\
\midrule
\textbf{CMMLU} & \fbox{51.68} & 42.67 & 33.9 & 30.09 & 35.55 & \lightblue{55.27} & \underline{55.01} \\
\midrule
\textbf{*-humanities} & \fbox{52.55} & 42.01 & 35.45 & 30.48 & 34.78 & \lightblue{63.4} & \underline{62.99} \\
\midrule
\textbf{*-stem} & \fbox{44.09} & 36.82 & 29.33 & 26.76 & 30.36 & \lightblue{47.29} & \underline{46.69} \\
\midrule
\textbf{*-social-science} & \fbox{53.02} & 44.41 & 34.55 & 30.97 & 38.04 & \underline{57.55} & \lightblue{57.79} \\
\midrule
\textbf{*-other} & \lightblue{57.58} & 47.3 & 36.77 & 32.25 & 38.45 & \underline{53.93} & \fbox{53.44} \\
\midrule
\textbf{*-china-specific} & \fbox{45.86} & 39.22 & 32.64 & 28.38 & 33.97 & \underline{55.69} & \lightblue{55.9} \\

\bottomrule
\end{tabular}
}
\end{table}

\section{Evaluations}
\label{evaluations}
The thorough evaluation demonstrates that the MAP-Neo model family achieves inspiring performance both on automatic benchmarks of base models and chat models.
Compared to the previous transparent LLM series, we underline MAP-Neo's distinctive performance on code, math, and instruction following abilities, which not only endows the MAP-Neo with academic and practical value.

\subsection{Base Model Performance}
\subsubsection{Main Results}

We present the results of our base models compared to several well-known LLMs, e.g. LLama3-8B and Mistral-7B, across standard academic benchmarks. 
All our evaluation metrics are derived from our assessments, ensuring consistency and transparency. 
We do not perform any post-processing on the evaluation content, maintaining the integrity of the raw outputs.

Our evaluation spans a comprehensive suite of public benchmarks in both English and Chinese, leveraging an internal evaluation framework designed for rigorous assessment. These benchmarks include a diverse range of datasets catering to multiple disciplines and aspects of language understanding and reasoning. 
Our evaluation strategy encompasses various metrics, including language modeling, specialized knowledge, and code generation. For datasets requiring multiple-choice selection, we employ a perplexity-based evaluation. 
For generation-based datasets, we generate free text and parse the results accordingly. 
The detailed results of our comparison with other base models are shown in Table~\ref{tab:base_model_compare}.

\textbf{Standard Benchmarks} We include Boolean Questions(BoolQ)~\citep{clark2019boolq}, Physical Interaction QA(PIQA)~\citep{bisk2020piqa}, Social Interaction QA(SIQA)~\citep{sap2019socialiqa}, HellaSwag~\citep{zellers2019hellaswag}, WinoGrande~\citep{sakaguchi2021winogrande}, ARC-Challenge(ARC-c)~\citep{clark2018think}, OpenBookQA-Fact~\citep{mihaylov2018can}, CommonsenseQA~\citep{talmor2018commonsenseqa}, and MMLU~\citep{hendrycks2020measuring} to assess general reasoning capabilities. All these benchmarks are tested with a 0-shot configuration, except for MMLU, which is evaluated with a 5-shot setup.

\textbf{Code Generation} We report the pass@1 scores of the evaluated models on HumanEval~\citep{chen2021evaluating}, HumanEval-Plus, MBPP~\citep{austin2021program}, and MBPP-Plus, all with a 0-shot configuration, following the EvalPlus framework~\citep{evalplus}.

\textbf{World Knowledge} We include NaturalQuestions(NQ)~\citep{kwiatkowski2019natural} and TriviaQA~\citep{joshi2017triviaqa} to assess world knowledge. Both benchmarks are tested with a 0-shot configuration.

\textbf{Reading Comprehension} We report the 0-shot average on SQuAD2.0~\citep{rajpurkar2018know}.

\textbf{Exams} We report the average scores for MATH~\citep{hendrycks2021measuring} and GSM8K~\citep{cobbe2021training}, both with a 4-shot configuration. For GSM8K, we employ a simple Chain-of-Thought prompting strategy: "Let's think step by step." For both datasets, we use the MAmmoTH evaluation framework~\citep{yue2023mammoth}. 

\textbf{Chinese} We use CMMLU~\citep{li2023cmmlu} and CEval~\citep{huang2024c} to assess performance on Chinese language tasks. Both benchmarks are tested with a 5-shot configuration.

\subsubsection{Discussions}
\paragraph{Data Quality}
MAP-Neo demonstrates significantly better performance on math, code, and complex reasoning by incorporating high-quality data, compared to previous transparent LLMs, e.g. Amber~\citep{liu2023llm360} and Pythia~\citep{biderman2023pythia}, adopting (presumably) lower quality data.

\paragraph{Gap between our MAP-Neo and other transparent LLMs}
In Table~\ref{tab:base_model_compare}, we note that transparent LLMs still significantly lag behind the performance of frontier industrial Open-weight LLMs with similar sizes (e.g. LLama3-8B, Mistral-7B). 
In contrast,
our MAP-Neo can match or even surpass them on part of the automatic benchmarks about math, code, and Chinese knowledge.
We call for increased participation in the development of transparent LLMs to further advance the LLM democratization.

\subsection{Aligned Model Performance}
\subsubsection{Main Results}

To accurately evaluate the realistic conversational performance of our aligned models, we selected several benchmarks that measure various aspects of model capabilities. These benchmarks were chosen for their ability to comprehensively assess key abilities such as alignment, instruction-following, real-world performance, and alignment with human preferences. Below are the specific benchmarks we used and the unique capabilities they evaluate:

\textbf{AlignBench~\citep{liu2023alignbench}} AlignBench evaluates the alignment capabilities of Chinese LLMs, ensuring high reliability and interpretability through a comprehensive, multi-dimensional benchmark and human-in-the-loop data curation.

\textbf{AlpacaEval~\citep{alpaca_eval,dubois2024length,dubois2023alpacafarm}} AlpacaEval measures instruction-following models' performance efficiently and reliably through an LLM-based automatic evaluation, validated against extensive human annotations.

\textbf{Arena-Hard~\citep{arenahard2024}} Arena-Hard evaluates LLMs' real-world performance and ability to reflect human preferences by constructing benchmarks from live data and ensuring robust model capability separation.

\textbf{CHC-Bench~\citep{du2024chinese}} CHC-Bench evaluates LLMs on their proficiency in Chinese culture, history, and language, with tasks like composing poetry, understanding ancient Chinese, and explaining Chinese terms, emphasizing the challenges for models trained mainly on English datasets.

\textbf{MT-Bench~\citep{zheng2024judging}} MT-Bench assesses LLM-based chat assistants' alignment with human preferences using strong LLMs as judges, achieving high agreement with human evaluations. 

\textbf{MMLU-Pro~\citep{wang2024mmlupro}} For the aligned models, we further evaluate {MMLU-Pro~\citep{wang2024mmlupro}} with a 5-shot configuration to reflect the model's capabilities more comprehensively.

\subsubsection{Discussions}
\paragraph{The effectiveness of Iterative DPO} 
In Table ~\ref{tab:aligned_model_compare}, when compared to Neo-7B-SFT, Neo-7B-Instruct shows significant improvement on the chat-related benchmark datasets (e.g., {AlignBench}, {AlpacaEval}, {Arena-Hard}, and {CHC-Bench}),
which further demonstrates the effectiveness of our Iterative DPO.

\paragraph{The performance of the chat model} 
Table ~\ref{tab:aligned_model_compare} shows that Amber-7B-Chat and OLMo-7B-Instruct perform poorly on Chat Benchmarks. 
We assume that the limited capabilities of the base model may severely limit the performance of corresponding instruction-tuned models on chat benchmarks.

\section{Societal Impact}
Data Colonialism is a deep concern when firms decide to exploit an algorithm product. \cite{couldry2019data} conceptualize the data colonialism framework and argue that Big Tech Giants, particularly in the U.S., use their massive data power to manipulate human behaviors and judgments and track people's traces continuously, forming a new social order. This suggests that controlling and owning data benefits firms' market status and generates large returns. So, making LLMs as firms' proprietary models is a common practice in the industry. \cite{ahmed2020democratization} discuss the barriers to AI democratization, such as the concentration of AI capabilities in large tech firms and elite universities. They underscore the importance of democratizing access to AI resources to mitigate the risks of data colonialism and promote equitable access to AI technologies across all institutions. \cite{shashidhar2023democratizing} discuss the dominance of proprietary LLMs and the need for high-performing open-source alternatives. They propose methods to enhance open-source models to compete with proprietary models while addressing privacy and resource-constrained concerns. They also point out how important the open-source model is in the LLMs community and acknowledge that firms with fewer resources and sensitive information are hesitant to trust the proprietary models. However, most LLMs are the product of a massive English corpus and are trained from English scratch \cite{zhang2023don}. How the open-source model can benefit the non-English language community and its data democratization remains unclear. 

Additionally, most open-source models are not thoroughly transparent. Open-source large language models (LLMs) often claim to be transparent and accessible, but many critical aspects of their development, such as data cleaning processes and pre-training code, remain undisclosed. This lack of transparency hampers reproducibility and the ability to fully understand and trust these models \cite{xu2022systematic}. For firms with financial constraints and privacy concerns, it is not economical to train their LLMs. Even though most open-source models give open access to the final and some intermediate checkpoints, they keep data sources, data pre-training code, and data processing methods opaque, those of which are the most costly parts of setting up an LLM. That is the key issue we want to tackle and then hope to promote full transparency in our community.

In our report, the MAP-Neo model might complement the current scarcity of Chinese corpus in LLMs. Importantly, our bi-lingual language model is a "thorough" open-source model--disclosing all key processes from sources of searching original data, and data cleaning to pre-training code base. Those disclosures significantly reduce the cost of deploying and customizing a LLM, especially for a Chinese LLM. It might have potential societal impacts. Firms with the need for a Chinese version of LLM but face constraints can be more able to leverage benefits from LLMs by using or referencing our MAP-Neo Model. It might improve social welfare in total and make a more vivid and diversified Chinese LLMs community \cite{colombo2014open}. Our advocates for thorough open-source action may attract more Chinese LLM researchers or relevant firms to fully disclose their models because thorough transparent open-source models can bring them sizable benefits from more constructive feedback and criticism. Those might make their models better and eventually accelerate the iterations of Chinese LLMs and empower the local community \cite{ricaurte2019data}. Overall, open innovation practices like disclosing the MAP-Neo model might alleviate the dominance of English LLMs and improve the inclusivity of the international LLMs community. 

Those open innovation practices may also benefit Small and Medium enterprises (SME) to introduce new products effectively \cite{spithoven2013open} and efficiently with easier implementation of their own customized LLMs, which may partially mitigate the threats of data colonialism from Big Tech Giants. Our Map-Neo model's open and economical attributes give an optimistic outlook for researchers in academia. Those attributes suggest that it is not hard and costly to set up the university's own AI without depending on specific Big Tech Giants' help. If universities have independent and decentralized control over their data and AI processes, it will prevent large companies from AI monopolization and promote data and AI democratization.

\section{Conclusion}
In this paper, we introduce MAP-Neo, which makes strides toward enhancing the transparency and accessibility of large language models (LLMs) by offering a fully open-source bilingual LLM suite. 
By sharing thoroughly detailed processes, from data curation, pre-training corpus (\textit{i.e.}, Matrix Data Pile), and model training to evaluation, we aim to support the academic and open-source communities in advancing transparent NLP research. 
Moreover, MAP-Neo narrows the gap with industry-level models (typically closed-source) with enhanced reasoning, instruction-following, and coding abilities. 
We hope that our work provides a valuable resource for researchers and developers, contributing to a broader effort to democratize access to advanced LLM technologies.

\section{Contributions and Acknowledgments}
\paragraph{Team Leaders:}
\begin{itemize}
    \item Ge Zhang, M-A-P, University of Waterloo, 01.AI, \textit{Data \& Pretrain \& Evaluation \& Model Architecture \& Codebase \& Alignment}
    \item Scott Qu, M-A-P, University of Manchester, 01.AI, \textit{Codebase \& Model Architecture \& Infra \& Pretrain}
    \item Jiaheng Liu, M-A-P, \textit{Scaling Law \& Alignment} 
\end{itemize}
\paragraph{Core Contributors: (Alphabet Order)}
\begin{itemize}
    \item Chenchen Zhang, Independent Researcher, \textit{Pretrain}
    \item Chenghua Lin. M-A-P, University of Manchester, \textit{Data}
    \item Chou Leuang Yu, CUHK-Shenzhen, \textit{Alignment \& Data}   
    \item Danny Pan, Peking University, \textit{Data \& Codebase}
    \item Esther Cheng, Peking University, \textit{Data}
    \item Jie Liu, The Chinese University of Hong Kong, \textit{Alignment}
    \item Qunshu Lin, 2077AI, \textit{Data}
    \item Raven Yuan, M-A-P, \textit{Pretrain \& Infra}
    \item Tuney Zheng, M-A-P, 01.AI, University of Waterloo, \textit{Pretrain \& Evaluation \& Alignment}
    \item Wei Pang, University of Waterloo, \textit{Data}
    \item Xinrun Du, M-A-P, 01.AI, \textit{Codebase \& Pretrain \& Alignment \& Evaluation}
    \item Yiming Liang, Institute of Automation, Chinese Academy of Sciences, \textit{Alignment \& Evaluation}
    \item Yinghao Ma, M-A-P, Queen Mary University of London, \textit{Scaling Law}
    \item Yizhi Li, M-A-P, University of Manchester, \textit{Data}
    \item Ziyang Ma, M-A-P, Shanghai Jiao Tong University, \textit{Alignment}
\end{itemize}
\paragraph{Contributors: (Alphabet Order)}
\begin{itemize}
    \item Bill Lin, University of Southern California, \textit{Alignment}
    \item Emmanouil Benetos, Queen Mary University of London, \textit{Scaling Law}
    \item Huan Yang, University of Warwick
, \textit{Ethics \& Societal Impact}
    \item Junting Zhou, Peking University, \textit{Data \& Scaling Law}
    \item Kaijing Ma, Tongji University, \textit{Data}
    \item Minghao Liu, 2077AI, \textit{Data}
    \item Morry Niu, 01.AI, \textit{Codebase}
    \item Noah Wang, 01.AI, \textit{Alignment}
    \item Quehry Que, Independent Researcher, \textit{Data}
    \item Ruibo Liu, Dartmouth University, \textit{Pretrain \& Model Architecture}
    \item Sine Liu, Independent Researcher, \textit{Infra}
    \item Shawn Guo, 01.AI, \textit{Data}    
    \item Soren Gao, Fudan University, \textit{Tokenization}
    \item Wangchunshu Zhou, M-A-P \& AIWaves Inc., \textit{Data}
    \item Xinyue Zhang, Unity, \textit{Ethics \& Data}
    \item Yizhi Zhou, Nanjing University, \textit{Data} 
    \item Yubo Wang, University of Waterloo, \textit{Pretrain}
    \item Yuelin Bai, M-A-P, Shenzhen Institute of Advanced Technology, CAS, \textit{Data}
    \item Yuhan Zhang, M-A-P, \textit{Data}
    \item Yuxiang Zhang, M-A-P, Waseda University, \textit{Codebase \& Evaluation \& Data}
    \item Zenith Wang, Independent Researcher, \textit{Data}
    \item Zhenzhu Yang, China University of Geosciences Beijing, \textit{Ethics \& Data}
    \item Zijian Zhao, 2077AI, \textit{Data}
\end{itemize}
\paragraph{Advisors:}
\begin{itemize}
    \item Jiajun Zhang, Wuhan AI Research, Institute of Automation, Chinese Academy of Sciences
\item Wanli Ouyang, The Chinese University of Hong Kong, Shanghai AI Lab
    \item Wenhao Huang, 01.AI
        \item Wenhu Chen, University of Waterloo

\end{itemize}

\section{Multimodal Art Projection}
Multimodal Art Projection (M-A-P) is an open-source research community. The community members are working on Artificial Intelligence-Generated Content (AIGC) topics, including text, audio, and vision modalities. We aim to prompt open research on large language/music/multimodal models (LLMs/LMMs) training, data collection, and development of fun applications.

\clearpage
\bibliography{colm2024_conference}

\begin{thebibliography}{128}
\providecommand{\natexlab}[1]{#1}
\providecommand{\url}[1]{\texttt{#1}}
\expandafter\ifx\csname urlstyle\endcsname\relax
  \providecommand{\doi}[1]{doi: #1}\else
  \providecommand{\doi}{doi: \begingroup \urlstyle{rm}\Url}\fi

\bibitem[Achiam et~al.(2023)Achiam, Adler, Agarwal, Ahmad, Akkaya, Aleman, Almeida, Altenschmidt, Altman, Anadkat, et~al.]{achiam2023gpt}
Josh Achiam, Steven Adler, Sandhini Agarwal, Lama Ahmad, Ilge Akkaya, Florencia~Leoni Aleman, Diogo Almeida, Janko Altenschmidt, Sam Altman, Shyamal Anadkat, et~al.
\newblock Gpt-4 technical report.
\newblock \emph{arXiv preprint arXiv:2303.08774}, 2023.

\bibitem[Ahmed \& Wahed(2020)Ahmed and Wahed]{ahmed2020democratization}
Nur Ahmed and Muntasir Wahed.
\newblock The de-democratization of ai: Deep learning and the compute divide in artificial intelligence research.
\newblock \emph{arXiv preprint arXiv:2010.15581}, 2020.

\bibitem[AI@Meta(2024)]{llama3modelcard}
AI@Meta.
\newblock Llama 3 model card.
\newblock 2024.
\newblock URL \url{https://github.com/meta-llama/llama3/blob/main/MODEL_CARD.md}.

\bibitem[Anthropic(2024)]{anthropic2024claude}
AI~Anthropic.
\newblock The claude 3 model family: Opus, sonnet, haiku.
\newblock \emph{Claude-3 Model Card}, 2024.

\bibitem[Austin et~al.(2021)Austin, Odena, Nye, Bosma, Michalewski, Dohan, Jiang, Cai, Terry, Le, et~al.]{austin2021program}
Jacob Austin, Augustus Odena, Maxwell Nye, Maarten Bosma, Henryk Michalewski, David Dohan, Ellen Jiang, Carrie Cai, Michael Terry, Quoc Le, et~al.
\newblock Program synthesis with large language models.
\newblock \emph{arXiv preprint arXiv:2108.07732}, 2021.

\bibitem[Azerbayev et~al.()Azerbayev, Schoelkopf, Paster, Santos, Mcaleer, Jiang, Deng, Biderman, and Welleck]{Azerbayev_Schoelkopf_Paster_Santos_Mcaleer_Jiang_Deng_Biderman_Welleck}
Zhangir Azerbayev, Hailey Schoelkopf, Keiran Paster, MarcoDos Santos, Stephen Mcaleer, AlbertQ Jiang, Jia Deng, Stella Biderman, and Sean Welleck.
\newblock Llemma: An open language model for mathematics.

\bibitem[Bai et~al.(2023)Bai, Bai, Chu, Cui, Dang, Deng, Fan, Ge, Han, Huang, et~al.]{bai2023qwen}
Jinze Bai, Shuai Bai, Yunfei Chu, Zeyu Cui, Kai Dang, Xiaodong Deng, Yang Fan, Wenbin Ge, Yu~Han, Fei Huang, et~al.
\newblock Qwen technical report.
\newblock \emph{arXiv preprint arXiv:2309.16609}, 2023.

\bibitem[Ben~Allal et~al.(2024)Ben~Allal, Lozhkov, Penedo, Wolf, and von Werra]{benallal2024cosmopedia}
Loubna Ben~Allal, Anton Lozhkov, Guilherme Penedo, Thomas Wolf, and Leandro von Werra.
\newblock Cosmopedia, 2024.
\newblock URL \url{https://huggingface.co/datasets/HuggingFaceTB/cosmopedia}.

\bibitem[Biderman et~al.(2023)Biderman, Schoelkopf, Anthony, Bradley, O’Brien, Hallahan, Khan, Purohit, Prashanth, Raff, et~al.]{biderman2023pythia}
Stella Biderman, Hailey Schoelkopf, Quentin~Gregory Anthony, Herbie Bradley, Kyle O’Brien, Eric Hallahan, Mohammad~Aflah Khan, Shivanshu Purohit, USVSN~Sai Prashanth, Edward Raff, et~al.
\newblock Pythia: A suite for analyzing large language models across training and scaling.
\newblock In \emph{International Conference on Machine Learning}, pp.\  2397--2430. PMLR, 2023.

\bibitem[Bisk et~al.(2020)Bisk, Zellers, Gao, Choi, et~al.]{bisk2020piqa}
Yonatan Bisk, Rowan Zellers, Jianfeng Gao, Yejin Choi, et~al.
\newblock Piqa: Reasoning about physical commonsense in natural language.
\newblock In \emph{Proceedings of the AAAI conference on artificial intelligence}, volume~34, pp.\  7432--7439, 2020.

\bibitem[Blecher et~al.(2023)Blecher, Cucurull, Scialom, and Stojnic]{blecher2023nougat}
Lukas Blecher, Guillem Cucurull, Thomas Scialom, and Robert Stojnic.
\newblock Nougat: Neural optical understanding for academic documents, 2023.

\bibitem[Bradley \& Terry(1952)Bradley and Terry]{btmodel}
Ralph~Allan Bradley and Milton~E. Terry.
\newblock Rank analysis of incomplete block designs: I. the method of paired comparisons.
\newblock \emph{Biometrika}, 39\penalty0 (3/4):\penalty0 324--345, 1952.
\newblock ISSN 00063444.
\newblock URL \url{http://www.jstor.org/stable/2334029}.

\bibitem[Broder(1997)]{broder1997resemblance}
Andrei~Z Broder.
\newblock On the resemblance and containment of documents.
\newblock In \emph{Proceedings. Compression and Complexity of SEQUENCES 1997 (Cat. No. 97TB100171)}, pp.\  21--29. IEEE, 1997.

\bibitem[Chen et~al.(2023{\natexlab{a}})Chen, Jian, Xi, Yi, Du, Ding, Zhu, Zong, Wang, and Zhang]{chen2023chinesewebtext}
Jianghao Chen, Pu~Jian, Tengxiao Xi, Dongyi Yi, Qianlong Du, Chenglin Ding, Guibo Zhu, Chengqing Zong, Jinqiao Wang, and Jiajun Zhang.
\newblock Chinesewebtext: Large-scale high-quality chinese web text extracted with effective evaluation model, 2023{\natexlab{a}}.

\bibitem[Chen et~al.(2021)Chen, Tworek, Jun, Yuan, Pinto, Kaplan, Edwards, Burda, Joseph, Brockman, et~al.]{chen2021evaluating}
Mark Chen, Jerry Tworek, Heewoo Jun, Qiming Yuan, Henrique Ponde de~Oliveira Pinto, Jared Kaplan, Harri Edwards, Yuri Burda, Nicholas Joseph, Greg Brockman, et~al.
\newblock Evaluating large language models trained on code.
\newblock \emph{arXiv preprint arXiv:2107.03374}, 2021.

\bibitem[Chen et~al.(2023{\natexlab{b}})Chen, Yin, Ku, Lu, Wan, Ma, Xu, Wang, and Xia]{chen2023theoremqa}
Wenhu Chen, Ming Yin, Max Ku, Pan Lu, Yixin Wan, Xueguang Ma, Jianyu Xu, Xinyi Wang, and Tony Xia.
\newblock Theoremqa: A theorem-driven question answering dataset.
\newblock In \emph{The 2023 Conference on Empirical Methods in Natural Language Processing}, 2023{\natexlab{b}}.

\bibitem[Chen et~al.(2024)Chen, Liu, Wang, Liu, Zhang, Lin, Chen, and Zhao]{chen2024agent}
Zehui Chen, Kuikun Liu, Qiuchen Wang, Jiangning Liu, Wenwei Zhang, Dahua Lin, Kai Chen, and Feng Zhao.
\newblock Agent-flan: Designing data and methods of effective agent tuning for large language models.
\newblock \emph{arXiv preprint arXiv:2403.12881}, 2024.

\bibitem[Chevalier et~al.(2024)Chevalier, Geng, Wettig, Chen, Mizera, Annala, Aragon, Fanlo, Frieder, Machado, Prabhakar, Thieu, Wang, Wang, Wu, Xia, Jia, Yu, Zhu, Ren, Arora, and Chen]{llm-science-tutor}
Alexis Chevalier, Jiayi Geng, Alexander Wettig, Howard Chen, Sebastian Mizera, Toni Annala, Max~Jameson Aragon, Arturo~Rodríguez Fanlo, Simon Frieder, Simon Machado, Akshara Prabhakar, Ellie Thieu, Jiachen~T. Wang, Zirui Wang, Xindi Wu, Mengzhou Xia, Wenhan Jia, Jiatong Yu, Jun-Jie Zhu, Zhiyong~Jason Ren, Sanjeev Arora, and Danqi Chen.
\newblock Language models as science tutors.
\newblock \emph{arXiv preprint arXiv: 2402.11111}, 2024.

\bibitem[Christiano et~al.(2017)Christiano, Leike, Brown, Martic, Legg, and Amodei]{christiano2017deep}
Paul~F Christiano, Jan Leike, Tom Brown, Miljan Martic, Shane Legg, and Dario Amodei.
\newblock Deep reinforcement learning from human preferences.
\newblock \emph{Advances in neural information processing systems}, 30, 2017.

\bibitem[Chung et~al.(2024)Chung, Hou, Longpre, Zoph, Tay, Fedus, Li, Wang, Dehghani, Brahma, et~al.]{chung2024scaling}
Hyung~Won Chung, Le~Hou, Shayne Longpre, Barret Zoph, Yi~Tay, William Fedus, Yunxuan Li, Xuezhi Wang, Mostafa Dehghani, Siddhartha Brahma, et~al.
\newblock Scaling instruction-finetuned language models.
\newblock \emph{Journal of Machine Learning Research}, 25\penalty0 (70):\penalty0 1--53, 2024.

\bibitem[Clark et~al.(2019)Clark, Lee, Chang, Kwiatkowski, Collins, and Toutanova]{clark2019boolq}
Christopher Clark, Kenton Lee, Ming-Wei Chang, Tom Kwiatkowski, Michael Collins, and Kristina Toutanova.
\newblock Boolq: Exploring the surprising difficulty of natural yes/no questions.
\newblock \emph{arXiv preprint arXiv:1905.10044}, 2019.

\bibitem[Clark et~al.(2018)Clark, Cowhey, Etzioni, Khot, Sabharwal, Schoenick, and Tafjord]{clark2018think}
Peter Clark, Isaac Cowhey, Oren Etzioni, Tushar Khot, Ashish Sabharwal, Carissa Schoenick, and Oyvind Tafjord.
\newblock Think you have solved question answering? try arc, the ai2 reasoning challenge.
\newblock \emph{arXiv preprint arXiv:1803.05457}, 2018.

\bibitem[Cobbe et~al.(2021)Cobbe, Kosaraju, Bavarian, Chen, Jun, Kaiser, Plappert, Tworek, Hilton, Nakano, et~al.]{cobbe2021training}
Karl Cobbe, Vineet Kosaraju, Mohammad Bavarian, Mark Chen, Heewoo Jun, Lukasz Kaiser, Matthias Plappert, Jerry Tworek, Jacob Hilton, Reiichiro Nakano, et~al.
\newblock Training verifiers to solve math word problems.
\newblock \emph{arXiv preprint arXiv:2110.14168}, 2021.

\bibitem[Colombo et~al.(2014)Colombo, Piva, and Rossi-Lamastra]{colombo2014open}
Massimo~G Colombo, Evila Piva, and Cristina Rossi-Lamastra.
\newblock Open innovation and within-industry diversification in small and medium enterprises: The case of open source software firms.
\newblock \emph{Research policy}, 43\penalty0 (5):\penalty0 891--902, 2014.

\bibitem[Computer(2023)]{together2023redpajama}
Together Computer.
\newblock Redpajama: an open dataset for training large language models, 2023.
\newblock URL \url{https://github.com/togethercomputer/RedPajama-Data}.

\bibitem[Contributors(2023)]{2023opencompass}
OpenCompass Contributors.
\newblock Opencompass: A universal evaluation platform for foundation models.
\newblock \url{https://github.com/open-compass/opencompass}, 2023.

\bibitem[Couldry \& Mejias(2019)Couldry and Mejias]{couldry2019data}
Nick Couldry and Ulises~A Mejias.
\newblock Data colonialism: Rethinking big data’s relation to the contemporary subject.
\newblock \emph{Television \& New Media}, 20\penalty0 (4):\penalty0 336--349, 2019.

\bibitem[DeepSeek-AI(2024)]{deepseek-llm}
DeepSeek-AI.
\newblock Deepseek llm: Scaling open-source language models with longtermism.
\newblock \emph{arXiv preprint arXiv:2401.02954}, 2024.
\newblock URL \url{https://github.com/deepseek-ai/DeepSeek-LLM}.

\bibitem[Deng et~al.(2024)Deng, Yang, Yuan, Huang, Wang, Liu, Tian, Pan, Zhang, Lin, et~al.]{deng2024composerx}
Qixin Deng, Qikai Yang, Ruibin Yuan, Yipeng Huang, Yi~Wang, Xubo Liu, Zeyue Tian, Jiahao Pan, Ge~Zhang, Hanfeng Lin, et~al.
\newblock Composerx: Multi-agent symbolic music composition with llms.
\newblock \emph{arXiv preprint arXiv:2404.18081}, 2024.

\bibitem[Du et~al.(2024)Du, Yu, Gao, Pan, Cheng, Ma, Yuan, Qu, Liu, Zheng, Luo, Zhou, Yuan, Chen, Fu, and Zhang]{du2024chinese}
Xinrun Du, Zhouliang Yu, Songyang Gao, Ding Pan, Yuyang Cheng, Ziyang Ma, Ruibin Yuan, Xingwei Qu, Jiaheng Liu, Tianyu Zheng, Xinchen Luo, Guorui Zhou, Binhang Yuan, Wenhu Chen, Jie Fu, and Ge~Zhang.
\newblock Chinese tiny llm: Pretraining a chinese-centric large language model, 2024.

\bibitem[Dubois et~al.(2023)Dubois, Li, Taori, Zhang, Gulrajani, Ba, Guestrin, Liang, and Hashimoto]{dubois2023alpacafarm}
Yann Dubois, Xuechen Li, Rohan Taori, Tianyi Zhang, Ishaan Gulrajani, Jimmy Ba, Carlos Guestrin, Percy Liang, and Tatsunori~B. Hashimoto.
\newblock Alpacafarm: A simulation framework for methods that learn from human feedback, 2023.

\bibitem[Dubois et~al.(2024)Dubois, Galambosi, Liang, and Hashimoto]{dubois2024length}
Yann Dubois, Bal{\'a}zs Galambosi, Percy Liang, and Tatsunori~B Hashimoto.
\newblock Length-controlled alpacaeval: A simple way to debias automatic evaluators.
\newblock \emph{arXiv preprint arXiv:2404.04475}, 2024.

\bibitem[Gabriel et~al.(2018)Gabriel, Kuo, McAuley, and Hsu]{gabriel2018identifying}
Rodney~A Gabriel, Tsung-Ting Kuo, Julian McAuley, and Chun-Nan Hsu.
\newblock Identifying and characterizing highly similar notes in big clinical note datasets.
\newblock \emph{Journal of biomedical informatics}, 82:\penalty0 63--69, 2018.

\bibitem[Geng \& Liu(2023)Geng and Liu]{openlm2023openllama}
Xinyang Geng and Hao Liu.
\newblock Openllama: An open reproduction of llama, May 2023.
\newblock URL \url{https://github.com/openlm-research/open_llama}.

\bibitem[Gionis et~al.(1999)Gionis, Indyk, Motwani, et~al.]{gionis1999similarity}
Aristides Gionis, Piotr Indyk, Rajeev Motwani, et~al.
\newblock Similarity search in high dimensions via hashing.
\newblock In \emph{Vldb}, volume~99, pp.\  518--529, 1999.

\bibitem[Groeneveld et~al.(2024)Groeneveld, Beltagy, Walsh, Bhagia, Kinney, Tafjord, Jha, Ivison, Magnusson, Wang, et~al.]{groeneveld2024olmo}
Dirk Groeneveld, Iz~Beltagy, Pete Walsh, Akshita Bhagia, Rodney Kinney, Oyvind Tafjord, Ananya~Harsh Jha, Hamish Ivison, Ian Magnusson, Yizhong Wang, et~al.
\newblock Olmo: Accelerating the science of language models.
\newblock \emph{arXiv preprint arXiv:2402.00838}, 2024.

\bibitem[Gyawali et~al.(2020)Gyawali, Anastasiou, and Knoth]{gyawali2020deduplication}
Bikash Gyawali, Lucas Anastasiou, and Petr Knoth.
\newblock Deduplication of scholarly documents using locality sensitive hashing and word embeddings.
\newblock 2020.

\bibitem[He et~al.(2023)He, Jin, Xu, Qiu, Wang, Li, Yan, Wang, and Lin]{he2023wanjuan}
Conghui He, Zhenjiang Jin, Chao Xu, Jiantao Qiu, Bin Wang, Wei Li, Hang Yan, Jiaqi Wang, and Dahua Lin.
\newblock Wanjuan: A comprehensive multimodal dataset for advancing english and chinese large models, 2023.

\bibitem[Henderson* et~al.(2022)Henderson*, Krass*, Zheng, Guha, Manning, Jurafsky, and Ho]{hendersonkrass2022pileoflaw}
Peter Henderson*, Mark~S. Krass*, Lucia Zheng, Neel Guha, Christopher~D. Manning, Dan Jurafsky, and Daniel~E. Ho.
\newblock Pile of law: Learning responsible data filtering from the law and a 256gb open-source legal dataset, 2022.
\newblock URL \url{https://arxiv.org/abs/2207.00220}.

\bibitem[Hendrycks et~al.(2020)Hendrycks, Burns, Basart, Zou, Mazeika, Song, and Steinhardt]{hendrycks2020measuring}
Dan Hendrycks, Collin Burns, Steven Basart, Andy Zou, Mantas Mazeika, Dawn Song, and Jacob Steinhardt.
\newblock Measuring massive multitask language understanding.
\newblock \emph{arXiv preprint arXiv:2009.03300}, 2020.

\bibitem[Hendrycks et~al.(2021)Hendrycks, Burns, Kadavath, Arora, Basart, Tang, Song, and Steinhardt]{hendrycks2021measuring}
Dan Hendrycks, Collin Burns, Saurav Kadavath, Akul Arora, Steven Basart, Eric Tang, Dawn Song, and Jacob Steinhardt.
\newblock Measuring mathematical problem solving with the math dataset.
\newblock \emph{arXiv preprint arXiv:2103.03874}, 2021.

\bibitem[Hernandez et~al.(2022)Hernandez, Brown, Conerly, DasSarma, Drain, El-Showk, Elhage, Hatfield-Dodds, Henighan, Hume, et~al.]{hernandez2022scaling}
Danny Hernandez, Tom Brown, Tom Conerly, Nova DasSarma, Dawn Drain, Sheer El-Showk, Nelson Elhage, Zac Hatfield-Dodds, Tom Henighan, Tristan Hume, et~al.
\newblock Scaling laws and interpretability of learning from repeated data.
\newblock \emph{arXiv preprint arXiv:2205.10487}, 2022.

\bibitem[Hoffmann et~al.(2022)Hoffmann, Borgeaud, Mensch, Buchatskaya, Cai, Rutherford, Casas, Hendricks, Welbl, Clark, et~al.]{hoffmann2022training}
Jordan Hoffmann, Sebastian Borgeaud, Arthur Mensch, Elena Buchatskaya, Trevor Cai, Eliza Rutherford, Diego de~Las Casas, Lisa~Anne Hendricks, Johannes Welbl, Aidan Clark, et~al.
\newblock Training compute-optimal large language models.
\newblock \emph{arXiv preprint arXiv:2203.15556}, 2022.

\bibitem[Hu et~al.(2024)Hu, Tu, Han, He, Cui, Long, Zheng, Fang, Huang, Zhao, et~al.]{hu2024minicpm}
Shengding Hu, Yuge Tu, Xu~Han, Chaoqun He, Ganqu Cui, Xiang Long, Zhi Zheng, Yewei Fang, Yuxiang Huang, Weilin Zhao, et~al.
\newblock Minicpm: Unveiling the potential of small language models with scalable training strategies.
\newblock \emph{arXiv preprint arXiv:2404.06395}, 2024.

\bibitem[Huang et~al.(2022)Huang, Lv, Cui, Lu, and Wei]{huang2022layoutlmv3}
Yupan Huang, Tengchao Lv, Lei Cui, Yutong Lu, and Furu Wei.
\newblock Layoutlmv3: Pre-training for document ai with unified text and image masking.
\newblock In \emph{Proceedings of the 30th ACM International Conference on Multimedia}, pp.\  4083--4091, 2022.

\bibitem[Huang et~al.(2024)Huang, Bai, Zhu, Zhang, Zhang, Su, Liu, Lv, Zhang, Fu, et~al.]{huang2024c}
Yuzhen Huang, Yuzhuo Bai, Zhihao Zhu, Junlei Zhang, Jinghan Zhang, Tangjun Su, Junteng Liu, Chuancheng Lv, Yikai Zhang, Yao Fu, et~al.
\newblock C-eval: A multi-level multi-discipline chinese evaluation suite for foundation models.
\newblock \emph{Advances in Neural Information Processing Systems}, 36, 2024.

\bibitem[Jaccard(1912)]{jaccard1912distribution}
Paul Jaccard.
\newblock The distribution of the flora in the alpine zone. 1.
\newblock \emph{New phytologist}, 11\penalty0 (2):\penalty0 37--50, 1912.

\bibitem[Jiang et~al.(2023)Jiang, Sablayrolles, Mensch, Bamford, Chaplot, Casas, Bressand, Lengyel, Lample, Saulnier, et~al.]{jiang2023mistral}
Albert~Q Jiang, Alexandre Sablayrolles, Arthur Mensch, Chris Bamford, Devendra~Singh Chaplot, Diego de~las Casas, Florian Bressand, Gianna Lengyel, Guillaume Lample, Lucile Saulnier, et~al.
\newblock Mistral 7b.
\newblock \emph{arXiv preprint arXiv:2310.06825}, 2023.

\bibitem[Joshi et~al.(2017)Joshi, Choi, Weld, and Zettlemoyer]{joshi2017triviaqa}
Mandar Joshi, Eunsol Choi, Daniel~S Weld, and Luke Zettlemoyer.
\newblock Triviaqa: A large scale distantly supervised challenge dataset for reading comprehension.
\newblock \emph{arXiv preprint arXiv:1705.03551}, 2017.

\bibitem[Joulin et~al.(2016)Joulin, Grave, Bojanowski, Douze, Jégou, and Mikolov]{Joulin_Grave_Bojanowski_Douze_Jégou_Mikolov_2016}
Armand Joulin, Edouard Grave, Piotr Bojanowski, Matthijs Douze, Hervé Jégou, and Tomas Mikolov.
\newblock Fasttext.zip: Compressing text classification models.
\newblock \emph{arXiv: Computation and Language,arXiv: Computation and Language}, Nov 2016.

\bibitem[Kaddour(2023)]{kaddour2023minipile}
Jean Kaddour.
\newblock The minipile challenge for data-efficient language models.
\newblock \emph{arXiv preprint arXiv:2304.08442}, 2023.

\bibitem[Kaplan et~al.(2020)Kaplan, McCandlish, Henighan, Brown, Chess, Child, Gray, Radford, Wu, and Amodei]{kaplan2020scaling}
Jared Kaplan, Sam McCandlish, Tom Henighan, Tom~B Brown, Benjamin Chess, Rewon Child, Scott Gray, Alec Radford, Jeffrey Wu, and Dario Amodei.
\newblock Scaling laws for neural language models.
\newblock \emph{arXiv preprint arXiv:2001.08361}, 2020.

\bibitem[Kiveris et~al.(2014)Kiveris, Lattanzi, Mirrokni, Rastogi, and Vassilvitskii]{kiveris2014connected}
Raimondas Kiveris, Silvio Lattanzi, Vahab Mirrokni, Vibhor Rastogi, and Sergei Vassilvitskii.
\newblock Connected components in mapreduce and beyond.
\newblock In \emph{Proceedings of the ACM Symposium on Cloud Computing}, pp.\  1--13, 2014.

\bibitem[Kocetkov et~al.(2022)Kocetkov, Li, Ben~Allal, Li, Mou, Muñoz~Ferrandis, Jernite, Mitchell, Hughes, Wolf, Bahdanau, von Werra, and de~Vries]{Kocetkov2022TheStack}
Denis Kocetkov, Raymond Li, Loubna Ben~Allal, Jia Li, Chenghao Mou, Carlos Muñoz~Ferrandis, Yacine Jernite, Margaret Mitchell, Sean Hughes, Thomas Wolf, Dzmitry Bahdanau, Leandro von Werra, and Harm de~Vries.
\newblock The stack: 3 tb of permissively licensed source code.
\newblock \emph{Preprint}, 2022.

\bibitem[Kowsari et~al.(2017)Kowsari, Brown, Heidarysafa, Jafari~Meimandi, , Gerber, and Barnes]{kowsari2017HDLTex}
Kamran Kowsari, Donald~E Brown, Mojtaba Heidarysafa, Kiana Jafari~Meimandi, , Matthew~S Gerber, and Laura~E Barnes.
\newblock Hdltex: Hierarchical deep learning for text classification.
\newblock In \emph{Machine Learning and Applications (ICMLA), 2017 16th IEEE International Conference on}. IEEE, 2017.

\bibitem[Kudo \& Richardson(2018)Kudo and Richardson]{kudo2018sentencepiece}
Taku Kudo and John Richardson.
\newblock Sentencepiece: A simple and language independent subword tokenizer and detokenizer for neural text processing.
\newblock \emph{arXiv preprint arXiv:1808.06226}, 2018.

\bibitem[Kwiatkowski et~al.(2019)Kwiatkowski, Palomaki, Redfield, Collins, Parikh, Alberti, Epstein, Polosukhin, Devlin, Lee, et~al.]{kwiatkowski2019natural}
Tom Kwiatkowski, Jennimaria Palomaki, Olivia Redfield, Michael Collins, Ankur Parikh, Chris Alberti, Danielle Epstein, Illia Polosukhin, Jacob Devlin, Kenton Lee, et~al.
\newblock Natural questions: a benchmark for question answering research.
\newblock \emph{Transactions of the Association for Computational Linguistics}, 7:\penalty0 453--466, 2019.

\bibitem[Lee et~al.(2021)Lee, Ippolito, Nystrom, Zhang, Eck, Callison-Burch, and Carlini]{lee2021deduplicating}
Katherine Lee, Daphne Ippolito, Andrew Nystrom, Chiyuan Zhang, Douglas Eck, Chris Callison-Burch, and Nicholas Carlini.
\newblock Deduplicating training data makes language models better.
\newblock \emph{arXiv preprint arXiv:2107.06499}, 2021.

\bibitem[Li et~al.(2022)Li, Guo, Zhou, An, Du, Zhu, Liu, Hu, and Yu]{li2022pp}
Chenxia Li, Ruoyu Guo, Jun Zhou, Mengtao An, Yuning Du, Lingfeng Zhu, Yi~Liu, Xiaoguang Hu, and Dianhai Yu.
\newblock Pp-structurev2: A stronger document analysis system.
\newblock \emph{arXiv preprint arXiv:2210.05391}, 2022.

\bibitem[Li et~al.(2023{\natexlab{a}})Li, Zhang, Koto, Yang, Zhao, Gong, Duan, and Baldwin]{li2023cmmlu}
Haonan Li, Yixuan Zhang, Fajri Koto, Yifei Yang, Hai Zhao, Yeyun Gong, Nan Duan, and Timothy Baldwin.
\newblock Cmmlu: Measuring massive multitask language understanding in chinese.
\newblock \emph{arXiv preprint arXiv:2306.09212}, 2023{\natexlab{a}}.

\bibitem[Li* et~al.(2024)Li*, Chiang, Frick, Dunlap, Zhu, Gonzalez, and Stoica]{arenahard2024}
Tianle Li*, Wei-Lin Chiang, Evan Frick, Lisa Dunlap, Banghua Zhu, Joseph~E. Gonzalez, and Ion Stoica.
\newblock From live data to high-quality benchmarks: The arena-hard pipeline, April 2024.
\newblock URL \url{https://lmsys.org/blog/2024-04-19-arena-hard/}.

\bibitem[Li et~al.(2023{\natexlab{b}})Li, Zhang, Dubois, Taori, Gulrajani, Guestrin, Liang, and Hashimoto]{alpaca_eval}
Xuechen Li, Tianyi Zhang, Yann Dubois, Rohan Taori, Ishaan Gulrajani, Carlos Guestrin, Percy Liang, and Tatsunori~B. Hashimoto.
\newblock Alpacaeval: An automatic evaluator of instruction-following models.
\newblock \url{https://github.com/tatsu-lab/alpaca_eval}, 2023{\natexlab{b}}.

\bibitem[Liu et~al.(2023{\natexlab{a}})Liu, Xia, Wang, and Zhang]{evalplus}
Jiawei Liu, Chunqiu~Steven Xia, Yuyao Wang, and Lingming Zhang.
\newblock Is your code generated by chat{GPT} really correct? rigorous evaluation of large language models for code generation.
\newblock In \emph{Thirty-seventh Conference on Neural Information Processing Systems}, 2023{\natexlab{a}}.
\newblock URL \url{https://openreview.net/forum?id=1qvx610Cu7}.

\bibitem[Liu et~al.(2024)Liu, Zhou, Liu, Bu, Yang, Han-Sen, and Ouyang]{liu2024iterative}
Jie Liu, Zhanhui Zhou, Jiaheng Liu, Xingyuan Bu, Chao Yang, Zhong Han-Sen, and Wanli Ouyang.
\newblock Iterative length-regularized direct preference optimization: A case study on improving 7b language models to gpt-4 level.
\newblock \emph{arXiv preprint arXiv:2406.11817}, 2024.

\bibitem[Liu et~al.(2023{\natexlab{b}})Liu, Lei, Wang, Huang, Feng, Wen, Cheng, Ke, Xu, Tam, Zhang, Sun, Wang, Zhang, Huang, Dong, and Tang]{liu2023alignbench}
Xiao Liu, Xuanyu Lei, Shengyuan Wang, Yue Huang, Zhuoer Feng, Bosi Wen, Jiale Cheng, Pei Ke, Yifan Xu, Weng~Lam Tam, Xiaohan Zhang, Lichao Sun, Hongning Wang, Jing Zhang, Minlie Huang, Yuxiao Dong, and Jie Tang.
\newblock Alignbench: Benchmarking chinese alignment of large language models, 2023{\natexlab{b}}.

\bibitem[Liu et~al.(2023{\natexlab{c}})Liu, Qiao, Neiswanger, Wang, Tan, Tao, Li, Wang, Sun, Pangarkar, Fan, Gu, Miller, Zhuang, He, Li, Koto, Tang, Ranjan, Shen, Ren, Iriondo, Mu, Hu, Schulze, Nakov, Baldwin, and Xing]{liu2023llm360}
Zhengzhong Liu, Aurick Qiao, Willie Neiswanger, Hongyi Wang, Bowen Tan, Tianhua Tao, Junbo Li, Yuqi Wang, Suqi Sun, Omkar Pangarkar, Richard Fan, Yi~Gu, Victor Miller, Yonghao Zhuang, Guowei He, Haonan Li, Fajri Koto, Liping Tang, Nikhil Ranjan, Zhiqiang Shen, Xuguang Ren, Roberto Iriondo, Cun Mu, Zhiting Hu, Mark Schulze, Preslav Nakov, Tim Baldwin, and Eric~P. Xing.
\newblock Llm360: Towards fully transparent open-source llms, 2023{\natexlab{c}}.

\bibitem[Longpre et~al.(2023)Longpre, Hou, Vu, Webson, Chung, Tay, Zhou, Le, Zoph, Wei, et~al.]{longpre2023flan}
Shayne Longpre, Le~Hou, Tu~Vu, Albert Webson, Hyung~Won Chung, Yi~Tay, Denny Zhou, Quoc~V Le, Barret Zoph, Jason Wei, et~al.
\newblock The flan collection: Designing data and methods for effective instruction tuning.
\newblock \emph{arXiv preprint arXiv:2301.13688}, 2023.

\bibitem[Lozhkov et~al.(2024)Lozhkov, Li, Allal, Cassano, Lamy-Poirier, Tazi, Tang, Pykhtar, Liu, Wei, Liu, Tian, Kocetkov, Zucker, Belkada, Wang, Liu, Abulkhanov, Paul, Li, Li, Risdal, Li, Zhu, Zhuo, Zheltonozhskii, Dade, Yu, Krauß, Jain, Su, He, Dey, Abati, Chai, Muennighoff, Tang, Oblokulov, Akiki, Marone, Mou, Mishra, Gu, Hui, Dao, Zebaze, Dehaene, Patry, Xu, McAuley, Hu, Scholak, Paquet, Robinson, Anderson, Chapados, Patwary, Tajbakhsh, Jernite, Ferrandis, Zhang, Hughes, Wolf, Guha, von Werra, and de~Vries]{lozhkov2024starcoder}
Anton Lozhkov, Raymond Li, Loubna~Ben Allal, Federico Cassano, Joel Lamy-Poirier, Nouamane Tazi, Ao~Tang, Dmytro Pykhtar, Jiawei Liu, Yuxiang Wei, Tianyang Liu, Max Tian, Denis Kocetkov, Arthur Zucker, Younes Belkada, Zijian Wang, Qian Liu, Dmitry Abulkhanov, Indraneil Paul, Zhuang Li, Wen-Ding Li, Megan Risdal, Jia Li, Jian Zhu, Terry~Yue Zhuo, Evgenii Zheltonozhskii, Nii Osae~Osae Dade, Wenhao Yu, Lucas Krauß, Naman Jain, Yixuan Su, Xuanli He, Manan Dey, Edoardo Abati, Yekun Chai, Niklas Muennighoff, Xiangru Tang, Muhtasham Oblokulov, Christopher Akiki, Marc Marone, Chenghao Mou, Mayank Mishra, Alex Gu, Binyuan Hui, Tri Dao, Armel Zebaze, Olivier Dehaene, Nicolas Patry, Canwen Xu, Julian McAuley, Han Hu, Torsten Scholak, Sebastien Paquet, Jennifer Robinson, Carolyn~Jane Anderson, Nicolas Chapados, Mostofa Patwary, Nima Tajbakhsh, Yacine Jernite, Carlos~Muñoz Ferrandis, Lingming Zhang, Sean Hughes, Thomas Wolf, Arjun Guha, Leandro von Werra, and Harm de~Vries.
\newblock Starcoder 2 and the stack v2: The next generation, 2024.

\bibitem[Luo et~al.(2023)Luo, Kong, Xu, Cao, Hao, Qu, Chen, Zhu, Zhao, Zhang, et~al.]{luo2023yayi}
Yin Luo, Qingchao Kong, Nan Xu, Jia Cao, Bao Hao, Baoyu Qu, Bo~Chen, Chao Zhu, Chenyang Zhao, Donglei Zhang, et~al.
\newblock Yayi 2: Multilingual open-source large language models.
\newblock \emph{arXiv preprint arXiv:2312.14862}, 2023.

\bibitem[Mihaylov et~al.(2018)Mihaylov, Clark, Khot, and Sabharwal]{mihaylov2018can}
Todor Mihaylov, Peter Clark, Tushar Khot, and Ashish Sabharwal.
\newblock Can a suit of armor conduct electricity? a new dataset for open book question answering.
\newblock \emph{arXiv preprint arXiv:1809.02789}, 2018.

\bibitem[{Nam Pham}(2024)]{nam_pham_2024}
{Nam Pham}.
\newblock tiny-strange-textbooks (revision 6f304f1), 2024.
\newblock URL \url{https://huggingface.co/datasets/nampdn-ai/tiny-strange-textbooks}.

\bibitem[Nguyen et~al.(2023)Nguyen, Van~Nguyen, Lai, Man, Ngo, Dernoncourt, Rossi, and Nguyen]{nguyen2023culturax}
Thuat Nguyen, Chien Van~Nguyen, Viet~Dac Lai, Hieu Man, Nghia~Trung Ngo, Franck Dernoncourt, Ryan~A Rossi, and Thien~Huu Nguyen.
\newblock Culturax: A cleaned, enormous, and multilingual dataset for large language models in 167 languages.
\newblock \emph{arXiv preprint arXiv:2309.09400}, 2023.

\bibitem[Paster et~al.(2023)Paster, Santos, Azerbayev, and Ba]{paster2023openwebmath}
Keiran Paster, Marco~Dos Santos, Zhangir Azerbayev, and Jimmy Ba.
\newblock Openwebmath: An open dataset of high-quality mathematical web text, 2023.

\bibitem[Penedo et~al.(2023)Penedo, Malartic, Hesslow, Cojocaru, Cappelli, Alobeidli, Pannier, Almazrouei, and Launay]{refinedweb}
Guilherme Penedo, Quentin Malartic, Daniel Hesslow, Ruxandra Cojocaru, Alessandro Cappelli, Hamza Alobeidli, Baptiste Pannier, Ebtesam Almazrouei, and Julien Launay.
\newblock The {R}efined{W}eb dataset for {F}alcon {LLM}: outperforming curated corpora with web data, and web data only.
\newblock \emph{arXiv preprint arXiv:2306.01116}, 2023.
\newblock URL \url{https://arxiv.org/abs/2306.01116}.

\bibitem[Qu et~al.(2024)Qu, Bai, Ma, Zhou, Lo, Liu, Yuan, Min, Liu, Zhang, et~al.]{qu2024mupt}
Xingwei Qu, Yuelin Bai, Yinghao Ma, Ziya Zhou, Ka~Man Lo, Jiaheng Liu, Ruibin Yuan, Lejun Min, Xueling Liu, Tianyu Zhang, et~al.
\newblock Mupt: A generative symbolic music pretrained transformer.
\newblock \emph{arXiv preprint arXiv:2404.06393}, 2024.

\bibitem[Rae et~al.(2022)Rae, Borgeaud, Cai, Millican, Hoffmann, Song, Aslanides, Henderson, Ring, Young, Rutherford, Hennigan, Menick, Cassirer, Powell, van~den Driessche, Hendricks, Rauh, Huang, Glaese, Welbl, Dathathri, Huang, Uesato, Mellor, Higgins, Creswell, McAleese, Wu, Elsen, Jayakumar, Buchatskaya, Budden, Sutherland, Simonyan, Paganini, Sifre, Martens, Li, Kuncoro, Nematzadeh, Gribovskaya, Donato, Lazaridou, Mensch, Lespiau, Tsimpoukelli, Grigorev, Fritz, Sottiaux, Pajarskas, Pohlen, Gong, Toyama, de~Masson~d'Autume, Li, Terzi, Mikulik, Babuschkin, Clark, de~Las~Casas, Guy, Jones, Bradbury, Johnson, Hechtman, Weidinger, Gabriel, Isaac, Lockhart, Osindero, Rimell, Dyer, Vinyals, Ayoub, Stanway, Bennett, Hassabis, Kavukcuoglu, and Irving]{rae2022scaling}
Jack~W. Rae, Sebastian Borgeaud, Trevor Cai, Katie Millican, Jordan Hoffmann, Francis Song, John Aslanides, Sarah Henderson, Roman Ring, Susannah Young, Eliza Rutherford, Tom Hennigan, Jacob Menick, Albin Cassirer, Richard Powell, George van~den Driessche, Lisa~Anne Hendricks, Maribeth Rauh, Po-Sen Huang, Amelia Glaese, Johannes Welbl, Sumanth Dathathri, Saffron Huang, Jonathan Uesato, John Mellor, Irina Higgins, Antonia Creswell, Nat McAleese, Amy Wu, Erich Elsen, Siddhant Jayakumar, Elena Buchatskaya, David Budden, Esme Sutherland, Karen Simonyan, Michela Paganini, Laurent Sifre, Lena Martens, Xiang~Lorraine Li, Adhiguna Kuncoro, Aida Nematzadeh, Elena Gribovskaya, Domenic Donato, Angeliki Lazaridou, Arthur Mensch, Jean-Baptiste Lespiau, Maria Tsimpoukelli, Nikolai Grigorev, Doug Fritz, Thibault Sottiaux, Mantas Pajarskas, Toby Pohlen, Zhitao Gong, Daniel Toyama, Cyprien de~Masson~d'Autume, Yujia Li, Tayfun Terzi, Vladimir Mikulik, Igor Babuschkin, Aidan Clark, Diego de~Las~Casas, Aurelia Guy, Chris Jones,
  James Bradbury, Matthew Johnson, Blake Hechtman, Laura Weidinger, Iason Gabriel, William Isaac, Ed~Lockhart, Simon Osindero, Laura Rimell, Chris Dyer, Oriol Vinyals, Kareem Ayoub, Jeff Stanway, Lorrayne Bennett, Demis Hassabis, Koray Kavukcuoglu, and Geoffrey Irving.
\newblock Scaling language models: Methods, analysis \& insights from training gopher, 2022.

\bibitem[Rafailov et~al.(2023)Rafailov, Sharma, Mitchell, Ermon, Manning, and Finn]{rafailov2023direct}
Rafael Rafailov, Archit Sharma, Eric Mitchell, Stefano Ermon, Christopher~D. Manning, and Chelsea Finn.
\newblock Direct preference optimization: Your language model is secretly a reward model, 2023.

\bibitem[{Raffel} et~al.(2019){Raffel}, {Shazeer}, {Roberts}, {Lee}, {Narang}, {Matena}, {Zhou}, {Li}, and {Liu}]{2019arXiv191010683R}
Colin {Raffel}, Noam {Shazeer}, Adam {Roberts}, Katherine {Lee}, Sharan {Narang}, Michael {Matena}, Yanqi {Zhou}, Wei {Li}, and Peter~J. {Liu}.
\newblock {Exploring the Limits of Transfer Learning with a Unified Text-to-Text Transformer}.
\newblock \emph{arXiv e-prints}, art. arXiv:1910.10683, October 2019.
\newblock \doi{10.48550/arXiv.1910.10683}.

\bibitem[Rajpurkar et~al.(2018)Rajpurkar, Jia, and Liang]{rajpurkar2018know}
Pranav Rajpurkar, Robin Jia, and Percy Liang.
\newblock Know what you don't know: Unanswerable questions for squad.
\newblock \emph{arXiv preprint arXiv:1806.03822}, 2018.

\bibitem[Reid et~al.(2024)Reid, Savinov, Teplyashin, Lepikhin, Lillicrap, Alayrac, Soricut, Lazaridou, Firat, Schrittwieser, et~al.]{reid2024gemini}
Machel Reid, Nikolay Savinov, Denis Teplyashin, Dmitry Lepikhin, Timothy Lillicrap, Jean-baptiste Alayrac, Radu Soricut, Angeliki Lazaridou, Orhan Firat, Julian Schrittwieser, et~al.
\newblock Gemini 1.5: Unlocking multimodal understanding across millions of tokens of context.
\newblock \emph{arXiv preprint arXiv:2403.05530}, 2024.

\bibitem[Ricaurte(2019)]{ricaurte2019data}
Paola Ricaurte.
\newblock Data epistemologies, the coloniality of power, and resistance.
\newblock \emph{Television \& New Media}, 20\penalty0 (4):\penalty0 350--365, 2019.

\bibitem[Ronsor(2023)]{ryokoai2023-bigknow2022}
Ronsor.
\newblock Bigknow2022: Bringing language models up to speed.
\newblock \url{https://github.com/RyokoAI/BigKnow2022}, 2023.

\bibitem[Sakaguchi et~al.(2021)Sakaguchi, Bras, Bhagavatula, and Choi]{sakaguchi2021winogrande}
Keisuke Sakaguchi, Ronan~Le Bras, Chandra Bhagavatula, and Yejin Choi.
\newblock Winogrande: An adversarial winograd schema challenge at scale.
\newblock \emph{Communications of the ACM}, 64\penalty0 (9):\penalty0 99--106, 2021.

\bibitem[Sandberg et~al.(1985)Sandberg, Goldberg, Kleiman, Walsh, and Lyon]{sandberg1985design}
Russel Sandberg, David Goldberg, Steve Kleiman, Dan Walsh, and Bob Lyon.
\newblock Design and implementation of the sun network filesystem.
\newblock In \emph{Proceedings of the summer 1985 USENIX conference}, pp.\  119--130, 1985.

\bibitem[Sap et~al.(2019)Sap, Rashkin, Chen, LeBras, and Choi]{sap2019socialiqa}
Maarten Sap, Hannah Rashkin, Derek Chen, Ronan LeBras, and Yejin Choi.
\newblock Socialiqa: Commonsense reasoning about social interactions.
\newblock \emph{arXiv preprint arXiv:1904.09728}, 2019.

\bibitem[Scao et~al.(2023)Scao, Fan, Akiki, Pavlick, Ili{\'c}, Hesslow, Castagn{\'e}, Luccioni, Yvon, Gall{\'e}, Tow, Rush, Biderman, Webson, Ammanamanchi, Wang, Sagot, Muennighoff, del Moral, Ruwase, Bawden, Bekman, Mcmillan-Major, Beltagy, Nguyen, Saulnier, Tan, Ortiz~Suarez, Sanh, Lauren{\c c}on, Jernite, Launay, Mitchell, Raffel, Gokaslan, Simhi, Soroa, Aji, Alfassy, Rogers, Nitzav, Xu, Mou, Emezue, Klamm, Leong, van Strien, Adelani, Radev, Ponferrada, Levkovizh, Kim, Natan, de~Toni, Dupont, Kruszewski, Pistilli, Elsahar, Benyamina, Tran, Yu, Abdulmumin, Johnson, Gonzalez-Dios, de~la Rosa, Chim, Dodge, Zhu, Chang, Frohberg, Tobing, Bhattacharjee, Almubarak, Chen, Lo, von Werra, Weber, Phan, Allal, Tanguy, Dey, Mu{\~n}oz, Masoud, Grandury, {\v S}a{\v s}ko, Huang, Coavoux, Singh, Jiang, Vu, Jauhar, Ghaleb, Subramani, Kassner, Khamis, Nguyen, Espejel, de~Gibert, Villegas, Henderson, Colombo, Amuok, Lhoest, Harliman, Bommasani, L{\'o}pez, Ribeiro, Osei, Pyysalo, Nagel, Bose, Muhammad, Sharma, Longpre,
  Nikpoor, Silberberg, Pai, Zink, Torrent, Schick, Thrush, Danchev, Nikoulina, Laippala, Lepercq, Prabhu, Alyafeai, Talat, Raja, Heinzerling, Si, Salesky, Mielke, Lee, Sharma, Santilli, Chaffin, Stiegler, Datta, Szczechla, Chhablani, Wang, Pandey, Strobelt, Fries, Rozen, Gao, Sutawika, Bari, Al-Shaibani, Manica, Nayak, Teehan, Albanie, Shen, Ben-David, Bach, Kim, Bers, Fevry, Neeraj, Thakker, Raunak, Tang, Yong, Sun, Brody, Uri, Tojarieh, Roberts, Chung, Tae, Phang, Press, Li, Narayanan, Bourfoune, Casper, Rasley, Ryabinin, Mishra, Zhang, Shoeybi, Peyrounette, Patry, Tazi, Sanseviero, von Platen, Cornette, Lavall{\'e}e, Lacroix, Rajbhandari, Gandhi, Smith, Requena, Patil, Dettmers, Baruwa, Singh, Cheveleva, Ligozat, Subramonian, N{\'e}v{\'e}ol, Lovering, Garrette, Tunuguntla, Reiter, Taktasheva, Voloshina, Bogdanov, Winata, Schoelkopf, Kalo, Novikova, Forde, Clive, Kasai, Kawamura, Hazan, Carpuat, Clinciu, Kim, Cheng, Serikov, Antverg, van~der Wal, Zhang, Zhang, Gehrmann, Pais, Shavrina, Scialom, Yun,
  Limisiewicz, Rieser, Protasov, Mikhailov, Pruksachatkun, Belinkov, Bamberger, Kasner, Rueda, Pestana, Feizpour, Khan, Faranak, Santos, Hevia, Unldreaj, Aghagol, Abdollahi, Tammour, Hajihosseini, Behroozi, Ajibade, Saxena, Ferrandis, Contractor, Lansky, David, Kiela, Nguyen, Tan, Baylor, Ozoani, Mirza, Ononiwu, Rezanejad, Jones, Bhattacharya, Solaiman, Sedenko, Nejadgholi, Passmore, Seltzer, Sanz, Dutra, Samagaio, Elbadri, Mieskes, Gerchick, Akinlolu, Mckenna, Qiu, Ghauri, Burynok, Abrar, Rajani, Elkott, Fahmy, Samuel, An, Kromann, Hao, Alizadeh, Shubber, Wang, Roy, Viguier, Le, Oyebade, Le, Yang, Nguyen, Kashyap, Palasciano, Callahan, Shukla, Miranda-Escalada, Singh, Beilharz, Wang, Brito, Zhou, Jain, Xu, Fourrier, Peri{\~n}{\'a}n, Molano, Yu, Manjavacas, Barth, Fuhrimann, Altay, Bayrak, Burns, Vrabec, Bello, Dash, Kang, Giorgi, Golde, Posada, Sivaraman, Bulchandani, Liu, Shinzato, de~Bykhovetz, Takeuchi, P{\`a}mies, Castillo, Nezhurina, S{\"a}nger, Samwald, Cullan, Weinberg, de~Wolf, Mihaljcic, Liu,
  Freidank, Kang, Seelam, Dahlberg, Broad, Muellner, Fung, Haller, Chandrasekhar, Eisenberg, Martin, Canalli, Su, Su, Cahyawijaya, Garda, Deshmukh, Mishra, Kiblawi, Ott, Sang-Aroonsiri, Kumar, Schweter, Bharati, Laud, Gigant, Kainuma, Kusa, Labrak, Bajaj, Venkatraman, Xu, Xu, Xu, Tan, Xie, Ye, Bras, Belkada, and Wolf]{bloom}
Teven~Le Scao, Angela Fan, Christopher Akiki, Ellie Pavlick, Suzana Ili{\'c}, Daniel Hesslow, Roman Castagn{\'e}, Alexandra~Sasha Luccioni, Fran{\c c}ois Yvon, Matthias Gall{\'e}, Jonathan Tow, Alexander~M. Rush, Stella Biderman, Albert Webson, Pawan~Sasanka Ammanamanchi, Thomas Wang, Beno{\^i}t Sagot, Niklas Muennighoff, Albert~Villanova del Moral, Olatunji Ruwase, Rachel Bawden, Stas Bekman, Angelina Mcmillan-Major, Iz~Beltagy, Huu Nguyen, Lucile Saulnier, Samson Tan, Pedro Ortiz~Suarez, Victor Sanh, Hugo Lauren{\c c}on, Yacine Jernite, Julien Launay, Margaret Mitchell, Colin Raffel, Aaron Gokaslan, Adi Simhi, Aitor Soroa, Alham~Fikri Aji, Amit Alfassy, Anna Rogers, Ariel~Kreisberg Nitzav, Canwen Xu, Chenghao Mou, Chris Emezue, Christopher Klamm, Colin Leong, Daniel van Strien, David~Ifeoluwa Adelani, Dragomir Radev, Eduardo~Gonz{\'a}lez Ponferrada, Efrat Levkovizh, Ethan Kim, Eyal~Bar Natan, Francesco de~Toni, G{\'e}rard Dupont, Germ{\'a}n Kruszewski, Giada Pistilli, Hady Elsahar, Hamza Benyamina, Hieu
  Tran, Ian Yu, Idris Abdulmumin, Isaac Johnson, Itziar Gonzalez-Dios, Javier de~la Rosa, Jenny Chim, Jesse Dodge, Jian Zhu, Jonathan Chang, J{\"o}rg Frohberg, Joseph Tobing, Joydeep Bhattacharjee, Khalid Almubarak, Kimbo Chen, Kyle Lo, Leandro von Werra, Leon Weber, Long Phan, Loubna~Ben Allal, Ludovic Tanguy, Manan Dey, Manuel~Romero Mu{\~n}oz, Maraim Masoud, Mar{\'i}a Grandury, Mario {\v S}a{\v s}ko, Max Huang, Maximin Coavoux, Mayank Singh, Mike Tian-Jian Jiang, Minh~Chien Vu, Mohammad~A. Jauhar, Mustafa Ghaleb, Nishant Subramani, Nora Kassner, Nurulaqilla Khamis, Olivier Nguyen, Omar Espejel, Ona de~Gibert, Paulo Villegas, Peter Henderson, Pierre Colombo, Priscilla Amuok, Quentin Lhoest, Rheza Harliman, Rishi Bommasani, Roberto~Luis L{\'o}pez, Rui Ribeiro, Salomey Osei, Sampo Pyysalo, Sebastian Nagel, Shamik Bose, Shamsuddeen~Hassan Muhammad, Shanya Sharma, Shayne Longpre, Somaieh Nikpoor, Stanislav Silberberg, Suhas Pai, Sydney Zink, Tiago~Timponi Torrent, Timo Schick, Tristan Thrush, Valentin Danchev,
  Vassilina Nikoulina, Veronika Laippala, Violette Lepercq, Vrinda Prabhu, Zaid Alyafeai, Zeerak Talat, Arun Raja, Benjamin Heinzerling, Chenglei Si, Elizabeth Salesky, Sabrina~J. Mielke, Wilson~Y. Lee, Abheesht Sharma, Andrea Santilli, Antoine Chaffin, Arnaud Stiegler, Debajyoti Datta, Eliza Szczechla, Gunjan Chhablani, Han Wang, Harshit Pandey, Hendrik Strobelt, Jason~Alan Fries, Jos Rozen, Leo Gao, Lintang Sutawika, M~Saiful Bari, Maged~S. Al-Shaibani, Matteo Manica, Nihal Nayak, Ryan Teehan, Samuel Albanie, Sheng Shen, Srulik Ben-David, Stephen~H. Bach, Taewoon Kim, Tali Bers, Thibault Fevry, Trishala Neeraj, Urmish Thakker, Vikas Raunak, Xiangru Tang, Zheng-Xin Yong, Zhiqing Sun, Shaked Brody, Yallow Uri, Hadar Tojarieh, Adam Roberts, Hyung~Won Chung, Jaesung Tae, Jason Phang, Ofir Press, Conglong Li, Deepak Narayanan, Hatim Bourfoune, Jared Casper, Jeff Rasley, Max Ryabinin, Mayank Mishra, Minjia Zhang, Mohammad Shoeybi, Myriam Peyrounette, Nicolas Patry, Nouamane Tazi, Omar Sanseviero, Patrick von
  Platen, Pierre Cornette, Pierre~Fran{\c c}ois Lavall{\'e}e, R{\'e}mi Lacroix, Samyam Rajbhandari, Sanchit Gandhi, Shaden Smith, St{\'e}phane Requena, Suraj Patil, Tim Dettmers, Ahmed Baruwa, Amanpreet Singh, Anastasia Cheveleva, Anne-Laure Ligozat, Arjun Subramonian, Aur{\'e}lie N{\'e}v{\'e}ol, Charles Lovering, Dan Garrette, Deepak Tunuguntla, Ehud Reiter, Ekaterina Taktasheva, Ekaterina Voloshina, Eli Bogdanov, Genta~Indra Winata, Hailey Schoelkopf, Jan-Christoph Kalo, Jekaterina Novikova, Jessica~Zosa Forde, Jordan Clive, Jungo Kasai, Ken Kawamura, Liam Hazan, Marine Carpuat, Miruna Clinciu, Najoung Kim, Newton Cheng, Oleg Serikov, Omer Antverg, Oskar van~der Wal, Rui Zhang, Ruochen Zhang, Sebastian Gehrmann, Shani Pais, Tatiana Shavrina, Thomas Scialom, Tian Yun, Tomasz Limisiewicz, Verena Rieser, Vitaly Protasov, Vladislav Mikhailov, Yada Pruksachatkun, Yonatan Belinkov, Zachary Bamberger, Zden{\v e}k Kasner, Alice Rueda, Amanda Pestana, Amir Feizpour, Ammar Khan, Amy Faranak, Ana Santos, Anthony
  Hevia, Antigona Unldreaj, Arash Aghagol, Arezoo Abdollahi, Aycha Tammour, Azadeh Hajihosseini, Bahareh Behroozi, Benjamin Ajibade, Bharat Saxena, Carlos~Mu{\~n}oz Ferrandis, Danish Contractor, David Lansky, Davis David, Douwe Kiela, Duong~A. Nguyen, Edward Tan, Emi Baylor, Ezinwanne Ozoani, Fatima Mirza, Frankline Ononiwu, Habib Rezanejad, Hessie Jones, Indrani Bhattacharya, Irene Solaiman, Irina Sedenko, Isar Nejadgholi, Jesse Passmore, Josh Seltzer, Julio~Bonis Sanz, Livia Dutra, Mairon Samagaio, Maraim Elbadri, Margot Mieskes, Marissa Gerchick, Martha Akinlolu, Michael Mckenna, Mike Qiu, Muhammed Ghauri, Mykola Burynok, Nafis Abrar, Nazneen Rajani, Nour Elkott, Nour Fahmy, Olanrewaju Samuel, Ran An, Rasmus Kromann, Ryan Hao, Samira Alizadeh, Sarmad Shubber, Silas Wang, Sourav Roy, Sylvain Viguier, Thanh Le, Tobi Oyebade, Trieu Le, Yoyo Yang, Zach Nguyen, Abhinav~Ramesh Kashyap, Alfredo Palasciano, Alison Callahan, Anima Shukla, Antonio Miranda-Escalada, Ayush Singh, Benjamin Beilharz, Bo~Wang, Caio
  Brito, Chenxi Zhou, Chirag Jain, Chuxin Xu, Cl{\'e}mentine Fourrier, Daniel~Le{\'o}n Peri{\~n}{\'a}n, Daniel Molano, Dian Yu, Enrique Manjavacas, Fabio Barth, Florian Fuhrimann, Gabriel Altay, Giyaseddin Bayrak, Gully Burns, Helena~U. Vrabec, Imane Bello, Ishani Dash, Jihyun Kang, John Giorgi, Jonas Golde, Jose~David Posada, Karthik~Rangasai Sivaraman, Lokesh Bulchandani, Lu~Liu, Luisa Shinzato, Madeleine~Hahn de~Bykhovetz, Maiko Takeuchi, Marc P{\`a}mies, Maria~A Castillo, Marianna Nezhurina, Mario S{\"a}nger, Matthias Samwald, Michael Cullan, Michael Weinberg, Michiel de~Wolf, Mina Mihaljcic, Minna Liu, Moritz Freidank, Myungsun Kang, Natasha Seelam, Nathan Dahlberg, Nicholas~Michio Broad, Nikolaus Muellner, Pascale Fung, Patrick Haller, Ramya Chandrasekhar, Renata Eisenberg, Robert Martin, Rodrigo Canalli, Rosaline Su, Ruisi Su, Samuel Cahyawijaya, Samuele Garda, Shlok~S Deshmukh, Shubhanshu Mishra, Sid Kiblawi, Simon Ott, Sinee Sang-Aroonsiri, Srishti Kumar, Stefan Schweter, Sushil Bharati, Tanmay Laud,
  Th{\'e}o Gigant, Tomoya Kainuma, Wojciech Kusa, Yanis Labrak, Yash~Shailesh Bajaj, Yash Venkatraman, Yifan Xu, Yingxin Xu, Yu~Xu, Zhe Tan, Zhongli Xie, Zifan Ye, Mathilde Bras, Younes Belkada, and Thomas Wolf.
\newblock {BLOOM: A 176B-Parameter Open-Access Multilingual Language Model}.
\newblock working paper or preprint, November 2023.
\newblock URL \url{https://inria.hal.science/hal-03850124}.

\bibitem[Schulman et~al.(2017)Schulman, Wolski, Dhariwal, Radford, and Klimov]{schulman2017proximal}
John Schulman, Filip Wolski, Prafulla Dhariwal, Alec Radford, and Oleg Klimov.
\newblock Proximal policy optimization algorithms, 2017.

\bibitem[Sennrich et~al.(2015)Sennrich, Haddow, and Birch]{sennrich2015neural}
Rico Sennrich, Barry Haddow, and Alexandra Birch.
\newblock Neural machine translation of rare words with subword units.
\newblock \emph{arXiv preprint arXiv:1508.07909}, 2015.

\bibitem[Shao et~al.(2024)Shao, Wang, Zhu, Xu, Song, Zhang, Li, Wu, and Guo]{shao2024deepseekmath}
Zhihong Shao, Peiyi Wang, Qihao Zhu, Runxin Xu, Junxiao Song, Mingchuan Zhang, YK~Li, Y~Wu, and Daya Guo.
\newblock Deepseekmath: Pushing the limits of mathematical reasoning in open language models.
\newblock \emph{arXiv preprint arXiv:2402.03300}, 2024.

\bibitem[Sharma et~al.(2019)Sharma, Li, and Wang]{DBLP:journals/corr/abs-1906-03741}
Eva Sharma, Chen Li, and Lu~Wang.
\newblock {BIGPATENT:} {A} large-scale dataset for abstractive and coherent summarization.
\newblock \emph{CoRR}, abs/1906.03741, 2019.
\newblock URL \url{http://arxiv.org/abs/1906.03741}.

\bibitem[Shashidhar et~al.(2023)Shashidhar, Chinta, Sahai, Wang, and Ji]{shashidhar2023democratizing}
Sumuk Shashidhar, Abhinav Chinta, Vaibhav Sahai, Zhenhailong Wang, and Heng Ji.
\newblock Democratizing llms: An exploration of cost-performance trade-offs in self-refined open-source models.
\newblock In \emph{The 2023 Conference on Empirical Methods in Natural Language Processing}, 2023.

\bibitem[Shazeer(2019)]{shazeer2019fast}
Noam Shazeer.
\newblock Fast transformer decoding: One write-head is all you need, 2019.

\bibitem[Shazeer(2020)]{shazeer2020glu}
Noam Shazeer.
\newblock Glu variants improve transformer, 2020.

\bibitem[Soboleva et~al.(2023)Soboleva, Al-Khateeb, Myers, Steeves, Hestness, and Dey]{cerebras2023slimpajama}
Daria Soboleva, Faisal Al-Khateeb, Robert Myers, Jacob~R Steeves, Joel Hestness, and Nolan Dey.
\newblock {SlimPajama: A 627B token cleaned and deduplicated version of RedPajama}.
\newblock \url{https://www.cerebras.net/blog/slimpajama-a-627b-token-cleaned-and-deduplicated-version-of-redpajama}, 2023.
\newblock URL \url{https://huggingface.co/datasets/cerebras/SlimPajama-627B}.

\bibitem[Soldaini et~al.(2024)Soldaini, Kinney, Bhagia, Schwenk, Atkinson, Authur, Bogin, Chandu, Dumas, Elazar, Hofmann, Jha, Kumar, Lucy, Lyu, Lambert, Magnusson, Morrison, Muennighoff, Naik, Nam, Peters, Ravichander, Richardson, Shen, Strubell, Subramani, Tafjord, Walsh, Zettlemoyer, Smith, Hajishirzi, Beltagy, Groeneveld, Dodge, and Lo]{dolma}
Luca Soldaini, Rodney Kinney, Akshita Bhagia, Dustin Schwenk, David Atkinson, Russell Authur, Ben Bogin, Khyathi Chandu, Jennifer Dumas, Yanai Elazar, Valentin Hofmann, Ananya~Harsh Jha, Sachin Kumar, Li~Lucy, Xinxi Lyu, Nathan Lambert, Ian Magnusson, Jacob Morrison, Niklas Muennighoff, Aakanksha Naik, Crystal Nam, Matthew~E. Peters, Abhilasha Ravichander, Kyle Richardson, Zejiang Shen, Emma Strubell, Nishant Subramani, Oyvind Tafjord, Pete Walsh, Luke Zettlemoyer, Noah~A. Smith, Hannaneh Hajishirzi, Iz~Beltagy, Dirk Groeneveld, Jesse Dodge, and Kyle Lo.
\newblock {Dolma: An Open Corpus of Three Trillion Tokens for Language Model Pretraining Research}.
\newblock \emph{arXiv preprint}, 2024.
\newblock URL \url{https://arxiv.org/abs/2402.00159}.

\bibitem[Spithoven et~al.(2013)Spithoven, Vanhaverbeke, and Roijakkers]{spithoven2013open}
Andr{\'e} Spithoven, Wim Vanhaverbeke, and Nadine Roijakkers.
\newblock Open innovation practices in smes and large enterprises.
\newblock \emph{Small business economics}, 41:\penalty0 537--562, 2013.

\bibitem[Su et~al.(2023)Su, Lu, Pan, Murtadha, Wen, and Liu]{su2023roformer}
Jianlin Su, Yu~Lu, Shengfeng Pan, Ahmed Murtadha, Bo~Wen, and Yunfeng Liu.
\newblock Roformer: Enhanced transformer with rotary position embedding, 2023.

\bibitem[Talmor et~al.(2018)Talmor, Herzig, Lourie, and Berant]{talmor2018commonsenseqa}
Alon Talmor, Jonathan Herzig, Nicholas Lourie, and Jonathan Berant.
\newblock Commonsenseqa: A question answering challenge targeting commonsense knowledge.
\newblock \emph{arXiv preprint arXiv:1811.00937}, 2018.

\bibitem[Teknium(2023)]{OpenHermes2.5}
Teknium.
\newblock Openhermes 2.5: An open dataset of synthetic data for generalist llm assistants, 2023.
\newblock URL \url{https://huggingface.co/datasets/teknium/OpenHermes-2.5}.

\bibitem[Thuat~Nguyen \& Nguyen(2024)Thuat~Nguyen and Nguyen]{nguyen2024culturay}
Huu~Nguyen Thuat~Nguyen and Thien Nguyen.
\newblock Culturay: A large cleaned multilingual dataset of 75 languages, 2024.

\bibitem[Touvron et~al.(2023{\natexlab{a}})Touvron, Lavril, Izacard, Martinet, Lachaux, Lacroix, Rozière, Goyal, Hambro, Azhar, Rodriguez, Joulin, Grave, and Lample]{touvron2023llama}
Hugo Touvron, Thibaut Lavril, Gautier Izacard, Xavier Martinet, Marie-Anne Lachaux, Timothée Lacroix, Baptiste Rozière, Naman Goyal, Eric Hambro, Faisal Azhar, Aurelien Rodriguez, Armand Joulin, Edouard Grave, and Guillaume Lample.
\newblock Llama: Open and efficient foundation language models.
\newblock \emph{ARXIV}, 2023{\natexlab{a}}.

\bibitem[Touvron et~al.(2023{\natexlab{b}})Touvron, Martin, Stone, Albert, Almahairi, Babaei, Bashlykov, Batra, Bhargava, Bhosale, Bikel, Blecher, Ferrer, Chen, Cucurull, Esiobu, Fernandes, Fu, Fu, Fuller, Gao, Goswami, Goyal, Hartshorn, Hosseini, Hou, Inan, Kardas, Kerkez, Khabsa, Kloumann, Korenev, Koura, Lachaux, Lavril, Lee, Liskovich, Lu, Mao, Martinet, Mihaylov, Mishra, Molybog, Nie, Poulton, Reizenstein, Rungta, Saladi, Schelten, Silva, Smith, Subramanian, Tan, Tang, Taylor, Williams, Kuan, Xu, Yan, Zarov, Zhang, Fan, Kambadur, Narang, Rodriguez, Stojnic, Edunov, and Scialom]{touvron2023llama2}
Hugo Touvron, Louis Martin, Kevin Stone, Peter Albert, Amjad Almahairi, Yasmine Babaei, Nikolay Bashlykov, Soumya Batra, Prajjwal Bhargava, Shruti Bhosale, Dan Bikel, Lukas Blecher, Cristian~Canton Ferrer, Moya Chen, Guillem Cucurull, David Esiobu, Jude Fernandes, Jeremy Fu, Wenyin Fu, Brian Fuller, Cynthia Gao, Vedanuj Goswami, Naman Goyal, Anthony Hartshorn, Saghar Hosseini, Rui Hou, Hakan Inan, Marcin Kardas, Viktor Kerkez, Madian Khabsa, Isabel Kloumann, Artem Korenev, Punit~Singh Koura, Marie-Anne Lachaux, Thibaut Lavril, Jenya Lee, Diana Liskovich, Yinghai Lu, Yuning Mao, Xavier Martinet, Todor Mihaylov, Pushkar Mishra, Igor Molybog, Yixin Nie, Andrew Poulton, Jeremy Reizenstein, Rashi Rungta, Kalyan Saladi, Alan Schelten, Ruan Silva, Eric~Michael Smith, Ranjan Subramanian, Xiaoqing~Ellen Tan, Binh Tang, Ross Taylor, Adina Williams, Jian~Xiang Kuan, Puxin Xu, Zheng Yan, Iliyan Zarov, Yuchen Zhang, Angela Fan, Melanie Kambadur, Sharan Narang, Aurelien Rodriguez, Robert Stojnic, Sergey Edunov, and Thomas
  Scialom.
\newblock Llama 2: Open foundation and fine-tuned chat models.
\newblock \emph{arXiv preprint arXiv: 2307.09288}, 2023{\natexlab{b}}.

\bibitem[Vaswani et~al.(2023)Vaswani, Shazeer, Parmar, Uszkoreit, Jones, Gomez, Kaiser, and Polosukhin]{vaswani2023attention}
Ashish Vaswani, Noam Shazeer, Niki Parmar, Jakob Uszkoreit, Llion Jones, Aidan~N. Gomez, Lukasz Kaiser, and Illia Polosukhin.
\newblock Attention is all you need, 2023.

\bibitem[Versley \& Panchenko(2012)Versley and Panchenko]{versley2012not}
Yannick Versley and Yana Panchenko.
\newblock Not just bigger: Towards better-quality web corpora.
\newblock In \emph{Proceedings of the seventh Web as Corpus Workshop (WAC7)}, pp.\  44--52, 2012.

\bibitem[Wang et~al.(2024{\natexlab{a}})Wang, Chen, Jia, Wang, Fang, Wang, Gao, Xie, Xu, Dai, Liu, Wu, Ding, Li, Huang, Deng, Yu, Ma, Xiao, Chen, Xiang, Wang, Zhu, Xiao, Wang, Wang, Ding, Huang, Xu, Tayier, Hu, Gao, Zheng, Ye, Li, Wan, Jiang, Wang, Cheng, Song, Tang, Xu, Zhang, Chen, Jiang, and Zhou]{wang2024weaver}
Tiannan Wang, Jiamin Chen, Qingrui Jia, Shuai Wang, Ruoyu Fang, Huilin Wang, Zhaowei Gao, Chunzhao Xie, Chuou Xu, Jihong Dai, Yibin Liu, Jialong Wu, Shengwei Ding, Long Li, Zhiwei Huang, Xinle Deng, Teng Yu, Gangan Ma, Han Xiao, Zixin Chen, Danjun Xiang, Yunxia Wang, Yuanyuan Zhu, Yi~Xiao, Jing Wang, Yiru Wang, Siran Ding, Jiayang Huang, Jiayi Xu, Yilihamu Tayier, Zhenyu Hu, Yuan Gao, Chengfeng Zheng, Yueshu Ye, Yihang Li, Lei Wan, Xinyue Jiang, Yujie Wang, Siyu Cheng, Zhule Song, Xiangru Tang, Xiaohua Xu, Ningyu Zhang, Huajun Chen, Yuchen~Eleanor Jiang, and Wangchunshu Zhou.
\newblock Weaver: Foundation models for creative writing.
\newblock \emph{arXiv preprint arXiv: 2401.17268}, 2024{\natexlab{a}}.

\bibitem[Wang et~al.(2024{\natexlab{b}})Wang, Ma, Zhang, Ni, Chandra, Tianle, Guo, Arulraj, He, Ren, Jiang, Zhuang, Wang, Fan, Ku, Yue, and Chen]{wang2024mmlupro}
Yubo Wang, Xueguang Ma, Ge~Zhang, Yuansheng Ni, Abhranil Chandra, Li~Tianle, Shiguang Guo, Aaran Arulraj, Xuan He, Weiming Ren, Ziyan Jiang, Alex Zhuang, Kai Wang, Richard Fan, Max Ku, Xiang Yue, and Wenhu Chen.
\newblock Mmlu-pro: Towards more robust and challenging multi-task language understanding evaluation.
\newblock Manuscript in preparation, 2024{\natexlab{b}}.

\bibitem[Wang et~al.(2023)Wang, Peng, Que, Liu, Zhou, Wu, Guo, Gan, Ni, Zhang, Zhang, Ouyang, Xu, Chen, Fu, and Peng]{wang2023rolellm}
Zekun~Moore Wang, Zhongyuan Peng, Haoran Que, Jiaheng Liu, Wangchunshu Zhou, Yuhan Wu, Hongcheng Guo, Ruitong Gan, Zehao Ni, Man Zhang, Zhaoxiang Zhang, Wanli Ouyang, Ke~Xu, Wenhu Chen, Jie Fu, and Junran Peng.
\newblock Rolellm: Benchmarking, eliciting, and enhancing role-playing abilities of large language models.
\newblock \emph{arXiv preprint arXiv: 2310.00746}, 2023.

\bibitem[Wei et~al.(2023{\natexlab{a}})Wei, Kong, Chen, Zhao, Ge, Yang, Sun, Han, and Zhang]{wei2023vary}
Haoran Wei, Lingyu Kong, Jinyue Chen, Liang Zhao, Zheng Ge, Jinrong Yang, Jianjian Sun, Chunrui Han, and Xiangyu Zhang.
\newblock Vary: Scaling up the vision vocabulary for large vision-language models.
\newblock \emph{arXiv preprint arXiv:2312.06109}, 2023{\natexlab{a}}.

\bibitem[Wei et~al.(2023{\natexlab{b}})Wei, Zhao, Zhang, Zhu, Wang, Yang, Li, Cheng, Lü, Hu, Li, Yang, Luo, Wu, Liu, Cheng, Cheng, Zhang, Zhang, Lin, Wang, Ma, Dong, Sun, Chen, Peng, Liang, Yan, Fang, and Zhou]{wei2023skywork}
Tianwen Wei, Liang Zhao, Lichang Zhang, Bo~Zhu, Lijie Wang, Haihua Yang, Biye Li, Cheng Cheng, Weiwei Lü, Rui Hu, Chenxia Li, Liu Yang, Xilin Luo, Xuejie Wu, Lunan Liu, Wenjun Cheng, Peng Cheng, Jianhao Zhang, Xiaoyu Zhang, Lei Lin, Xiaokun Wang, Yutuan Ma, Chuanhai Dong, Yanqi Sun, Yifu Chen, Yongyi Peng, Xiaojuan Liang, Shuicheng Yan, Han Fang, and Yahui Zhou.
\newblock Skywork: A more open bilingual foundation model, 2023{\natexlab{b}}.

\bibitem[Xu et~al.(2022)Xu, Alon, Neubig, and Hellendoorn]{xu2022systematic}
Frank~F Xu, Uri Alon, Graham Neubig, and Vincent~Josua Hellendoorn.
\newblock A systematic evaluation of large language models of code.
\newblock In \emph{Proceedings of the 6th ACM SIGPLAN International Symposium on Machine Programming}, pp.\  1--10, 2022.

\bibitem[Xu et~al.(2023)Xu, Lee, Sukhbaatar, and Weston]{xu2023some}
Jing Xu, Andrew Lee, Sainbayar Sukhbaatar, and Jason Weston.
\newblock Some things are more cringe than others: Preference optimization with the pairwise cringe loss.
\newblock \emph{arXiv preprint arXiv:2312.16682}, 2023.

\bibitem[Yang et~al.(2024)Yang, Wang, Chen, Wang, Pu, Gao, Huang, Song, and Huang]{psychogat}
Qisen Yang, Zekun Wang, Honghui Chen, Shenzhi Wang, Yifan Pu, Xin Gao, Wenhao Huang, Shiji Song, and Gao Huang.
\newblock Llm agents for psychology: A study on gamified assessments.
\newblock \emph{arXiv preprint arXiv: 2402.12326}, 2024.

\bibitem[Young et~al.(2024)Young, Chen, Li, Huang, Zhang, Zhang, Li, Zhu, Chen, Chang, et~al.]{young2024yi}
Alex Young, Bei Chen, Chao Li, Chengen Huang, Ge~Zhang, Guanwei Zhang, Heng Li, Jiangcheng Zhu, Jianqun Chen, Jing Chang, et~al.
\newblock Yi: Open foundation models by 01. ai.
\newblock \emph{arXiv preprint arXiv:2403.04652}, 2024.

\bibitem[Yu et~al.(2024)Yu, Baker, Chen, Ning, and Sun]{yu2024llasmol}
Botao Yu, Frazier~N. Baker, Ziqi Chen, Xia Ning, and Huan Sun.
\newblock Llasmol: Advancing large language models for chemistry with a large-scale, comprehensive, high-quality instruction tuning dataset.
\newblock \emph{arXiv preprint arXiv:2402.09391}, 2024.

\bibitem[Yuan et~al.(2024)Yuan, Lin, Wang, Tian, Wu, Shen, Zhang, Wu, Liu, Zhou, et~al.]{yuan2024chatmusician}
Ruibin Yuan, Hanfeng Lin, Yi~Wang, Zeyue Tian, Shangda Wu, Tianhao Shen, Ge~Zhang, Yuhang Wu, Cong Liu, Ziya Zhou, et~al.
\newblock Chatmusician: Understanding and generating music intrinsically with llm.
\newblock \emph{arXiv preprint arXiv:2402.16153}, 2024.

\bibitem[Yue et~al.(2023)Yue, Qu, Zhang, Fu, Huang, Sun, Su, and Chen]{yue2023mammoth}
Xiang Yue, Xingwei Qu, Ge~Zhang, Yao Fu, Wenhao Huang, Huan Sun, Yu~Su, and Wenhu Chen.
\newblock Mammoth: Building math generalist models through hybrid instruction tuning.
\newblock \emph{arXiv preprint arXiv:2309.05653}, 2023.

\bibitem[Yue et~al.(2024)Yue, Zheng, Zhang, and Chen]{yue2024mammoth2}
Xiang Yue, Tuney Zheng, Ge~Zhang, and Wenhu Chen.
\newblock Mammoth2: Scaling instructions from the web.
\newblock \emph{arXiv preprint arXiv:2405.03548}, 2024.

\bibitem[Zaharia et~al.(2012)Zaharia, Chowdhury, Das, Dave, Ma, McCauly, Franklin, Shenker, and Stoica]{zaharia2012resilient}
Matei Zaharia, Mosharaf Chowdhury, Tathagata Das, Ankur Dave, Justin Ma, Murphy McCauly, Michael~J Franklin, Scott Shenker, and Ion Stoica.
\newblock Resilient distributed datasets: A $\{$Fault-Tolerant$\}$ abstraction for $\{$In-Memory$\}$ cluster computing.
\newblock In \emph{9th USENIX symposium on networked systems design and implementation (NSDI 12)}, pp.\  15--28, 2012.

\bibitem[Zellers et~al.(2019)Zellers, Holtzman, Bisk, Farhadi, and Choi]{zellers2019hellaswag}
Rowan Zellers, Ari Holtzman, Yonatan Bisk, Ali Farhadi, and Yejin Choi.
\newblock Hellaswag: Can a machine really finish your sentence?
\newblock \emph{arXiv preprint arXiv:1905.07830}, 2019.

\bibitem[Zhang \& Sennrich(2019)Zhang and Sennrich]{zhang2019root}
Biao Zhang and Rico Sennrich.
\newblock Root mean square layer normalization, 2019.

\bibitem[Zhang et~al.(2023{\natexlab{a}})Zhang, Shi, Liu, Yuan, Li, Dong, Shu, Li, Wang, Lin, et~al.]{zhang2023chinese}
Ge~Zhang, Yemin Shi, Ruibo Liu, Ruibin Yuan, Yizhi Li, Siwei Dong, Yu~Shu, Zhaoqun Li, Zekun Wang, Chenghua Lin, et~al.
\newblock Chinese open instruction generalist: A preliminary release.
\newblock \emph{arXiv preprint arXiv:2304.07987}, 2023{\natexlab{a}}.

\bibitem[Zhang et~al.(2023{\natexlab{b}})Zhang, Li, Hauer, Shi, and Kondrak]{zhang2023don}
Xiang Zhang, Senyu Li, Bradley Hauer, Ning Shi, and Grzegorz Kondrak.
\newblock Don’t trust chatgpt when your question is not in english: A study of multilingual abilities and types of llms.
\newblock In \emph{Proceedings of the 2023 Conference on Empirical Methods in Natural Language Processing}, pp.\  7915--7927, 2023{\natexlab{b}}.

\bibitem[Zhang et~al.(2024)Zhang, Luo, Yuan, and Yao]{zhang2024automathtext}
Yifan Zhang, Yifan Luo, Yang Yuan, and Andrew Chi-Chih Yao.
\newblock Automathtext: Autonomous data selection with language models for mathematical texts.
\newblock \emph{arXiv preprint arXiv:2402.07625}, 2024.

\bibitem[Zheng et~al.(2024{\natexlab{a}})Zheng, Chiang, Sheng, Zhuang, Wu, Zhuang, Lin, Li, Li, Xing, et~al.]{zheng2024judging}
Lianmin Zheng, Wei-Lin Chiang, Ying Sheng, Siyuan Zhuang, Zhanghao Wu, Yonghao Zhuang, Zi~Lin, Zhuohan Li, Dacheng Li, Eric Xing, et~al.
\newblock Judging llm-as-a-judge with mt-bench and chatbot arena.
\newblock \emph{Advances in Neural Information Processing Systems}, 36, 2024{\natexlab{a}}.

\bibitem[Zheng et~al.(2024{\natexlab{b}})Zheng, Zhang, Shen, Liu, Lin, Fu, Chen, and Yue]{zheng2024opencodeinterpreter}
Tianyu Zheng, Ge~Zhang, Tianhao Shen, Xueling Liu, Bill~Yuchen Lin, Jie Fu, Wenhu Chen, and Xiang Yue.
\newblock Opencodeinterpreter: Integrating code generation with execution and refinement.
\newblock \emph{arXiv preprint arXiv:2402.14658}, 2024{\natexlab{b}}.

\bibitem[Zhu et~al.(2023)Zhu, Frick, Wu, Zhu, and Jiao]{starling2023}
Banghua Zhu, Evan Frick, Tianhao Wu, Hanlin Zhu, and Jiantao Jiao.
\newblock Starling-7b: Improving llm helpfulness \& harmlessness with rlaif, November 2023.

\bibitem[Zhuang et~al.(2024{\natexlab{a}})Zhuang, Zhang, Zheng, Du, Wang, Ren, Huang, Fu, Yue, and Chen]{zhuang2024structlm}
Alex Zhuang, Ge~Zhang, Tianyu Zheng, Xinrun Du, Junjie Wang, Weiming Ren, Stephen~W. Huang, Jie Fu, Xiang Yue, and Wenhu Chen.
\newblock Structlm: Towards building generalist models for structured knowledge grounding, 2024{\natexlab{a}}.

\bibitem[Zhuang et~al.(2024{\natexlab{b}})Zhuang, Jiang, He, and Wu]{zhuang2024chuxin}
Xiaomin Zhuang, Yufan Jiang, Qiaozhi He, and Zhihua Wu.
\newblock Chuxin: 1.6 b technical report.
\newblock \emph{arXiv preprint arXiv:2405.04828}, 2024{\natexlab{b}}.

\end{thebibliography}
\bibliographystyle{colm2024_conference}

\clearpage
\appendix
\section{Appendix}
\subsection{Details of Heuristic Rules for English Texts}
\label{sec:appendix-heuristic-rules-en}
\begin{table}[!ht]
\centering
\caption{Details of Heuristic Rules for English Texts}
\resizebox{\textwidth}{!}{
\begin{tabular}{ll}
\midrule
\textbf{Rule} & \textbf{Note} \\
\midrule
\multicolumn{2}{c}{\textbf{Document-level Filtering}} \\ 
\midrule
Mean word length [3, 10] & - \\
Lines that end with an ellipsis $\leq0.2$ & Defined as ellipsis: '...', '…', '……' \\
Lines starting with bullet point $\leq0.9$ & Bullet points: \begin{CJK*}{UTF8}{gbsn} "•", "●", "○", "□", "※", "·" \end{CJK*} etc. \\

Words that contain no alphabetical character $\leq0.4$ & - \\
Fraction of Stop words in the document must be $\geq0.06$ & - \\
Number of stop words in the document must be $\geq2$ & - \\
Symbols to words in the content must be $< 0.5$ & - \\
Number of words in the content after normalization [50 ,10000] & - \\
Score of the language identification model must be $>0.8$ & Evaluated by fasttext \\
Number of characters must be $\geq200$ & - \\
Number of lines $>1$ & - \\
Number of sentences $>1$ & - \\
Ratio of '{' or '}' and words in between must be $<0.025$ & - \\
'lorem ipsum' count must be $<3e-08$ & - \\
Number of sentences must be $<7500$ & - \\
Words only consist of uppercase letters $\leq0.1$ & - \\
Fraction of Unique words [0.1, +inf) & - \\
Entropy of the unigram distribution of the content within [3, 6] & - \\
Fraction of lines end with 'readmore' $\leq0.1$ & - \\
Fraction of nonconsecutive hashtags in words $\leq0.1$ & - \\
Fraction of nonconsecutive ellipsis in words $\leq0.1$ & - \\
punctuations in words $>0$ & - \\
Non-alpha words over non-punctuation words $\leq0.2$ & - \\
Digital words over non-punctuation words $\leq0.3$ & - \\
\midrule
\multicolumn{2}{c}{\textbf{Duplicates Filtering}} \\
\midrule
Fraction of characters in duplicate word 10-grams $\leq0.10$ & - \\
Fraction of characters in duplicate word 9-grams
$\leq0.11$ & - \\
Fraction of characters in duplicate word 8-grams
$\leq0.12$ & - \\
Fraction of characters in duplicate word 7-grams
$\leq0.13$ & - \\
Fraction of characters in duplicate word 6-grams
$\leq0.14$ & - \\
Fraction of characters in duplicate word 5-grams
$\leq0.15$ & - \\
Fraction of characters in top word 4-grams
$\leq0.16$ & - \\
Fraction of characters in top word 3-grams
$\leq0.18$ & - \\
Fraction of characters in top word 2-grams
$\leq0.20$ & - \\
Fraction of duplicate sentences
$\leq0.30$ & - \\
Fraction of characters in duplicate sentences
$\leq0.20$ & - \\
\midrule
\multicolumn{2}{c}{\textbf{Prohibited Words Filtering}} \\ 
\midrule
Text should not contain words in the bad words list & Words related to pornography, politics, violence, etc. \\
\bottomrule
\end{tabular}
}
\end{table}


\newpage
\subsection{Details of Heuristic Rules for Chinese Texts}
\label{sec:appendix-heuristic-rules-zh}

\begin{table}[ht]
\centering
\caption{Details of Heuristic Rules for Chinese Texts.}
\resizebox{\textwidth}{!}{
\begin{tabular}{ll}
\toprule
\textbf{Rule} & \textbf{Note} \\
\midrule
\multicolumn{2}{c}{\textbf{Data Format Unification}} \\ 
\midrule
Convert full-angle symbols to half-angle & - \\
\midrule
\multicolumn{2}{c}{\textbf{URL Filtering}} \\
\midrule
The text should not contain blacklisted URLs & Blacklists obtained from \href{http://dsi.ut-capitole.fr/blacklists/}{Blacklists UT1}. \\
Remove links via regular expression & - \\
\midrule
\multicolumn{2}{c}{\textbf{Sentence-level Filtering}} \\
\midrule
Only retain sentences with terminal punctuation & Terminal punctuation: ['.', '!', '?', '……', '…']. \\
Exclude sentences containing "javascript" & - \\
Contain at least 3 words & Word tokenization by jieba. \\
Exclude sentences with "lorem ipsum" & - \\
Exclude sentences with bad words & Words related to pornography, politics, violence, etc. \\
\midrule
\multicolumn{2}{c}{\textbf{Document-level Filtering}} \\
\midrule
Number of sentences $>1$ & - \\
Characters after normalization [50, 10000] & - \\
Mean word length [1.3, 10] & - \\
Fraction of nonconsecutive hashtags $\leq 0.1$ & - \\
Fraction of nonconsecutive ellipsis $\leq 0.1$ & Defined as ellipsis: '...', '…', '……'. \\
Fraction of full brackets \begin{CJK*}{UTF8}{gbsn} 【】 \end{CJK*} $\leq 0.1$ & - \\
Fraction of digital words over non-punctuation words $\leq 0.3$ & - \\
Lines ending with "readmore" etc. $\leq 0.3$ & Endings include: "readmore", \begin{CJK*}{UTF8}{gbsn} "展开", "更多", "。。。" \end{CJK*} \\
Lines starting with bullet point $\leq 0.9$ & Bullet points: \begin{CJK*}{UTF8}{gbsn} "•", "●", "○", "□", "※", "·" \end{CJK*} etc. \\
Fraction of punctuation in words $> 0$ & - \\
Fraction of unique words $> 0.1$ & - \\
Entropy of unigram distribution $\geq 3$ & - \\
Text quality score $> 0.4$ & Evaluated by fasttext \\
\midrule
\multicolumn{2}{c}{\textbf{Duplicates Filtering}} \\
\midrule
Fraction of characters in duplicate word 10-grams $\leq 0.60$ & - \\
Fraction of characters in duplicate word 9-grams $\leq 0.60$ & - \\
Fraction of characters in duplicate word 8-grams $\leq 0.60$ & - \\
Fraction of characters in duplicate word 7-grams $\leq 0.60$ & - \\
Fraction of characters in duplicate word 6-grams $\leq 0.60$ & - \\
Fraction of characters in duplicate word 5-grams $\leq 0.60$ & - \\
Fraction of characters in top word 4-grams $\leq 0.16$ & - \\
Fraction of characters in top word 3-grams $\leq 0.18$ & - \\
Fraction of characters in top word 2-grams $\leq 0.20$ & - \\
Fraction of duplicate sentences $\leq 0.30$ & - \\
Fraction of characters in duplicate sentences $\leq 0.20$ & - \\
\bottomrule
\end{tabular}
}
\end{table}

\clearpage

\subsection{Training Framework Overflow Details}
\label{sec_appendix:overflow}
\begin{figure}[!h]
\centering
\includegraphics[width=0.99\linewidth]{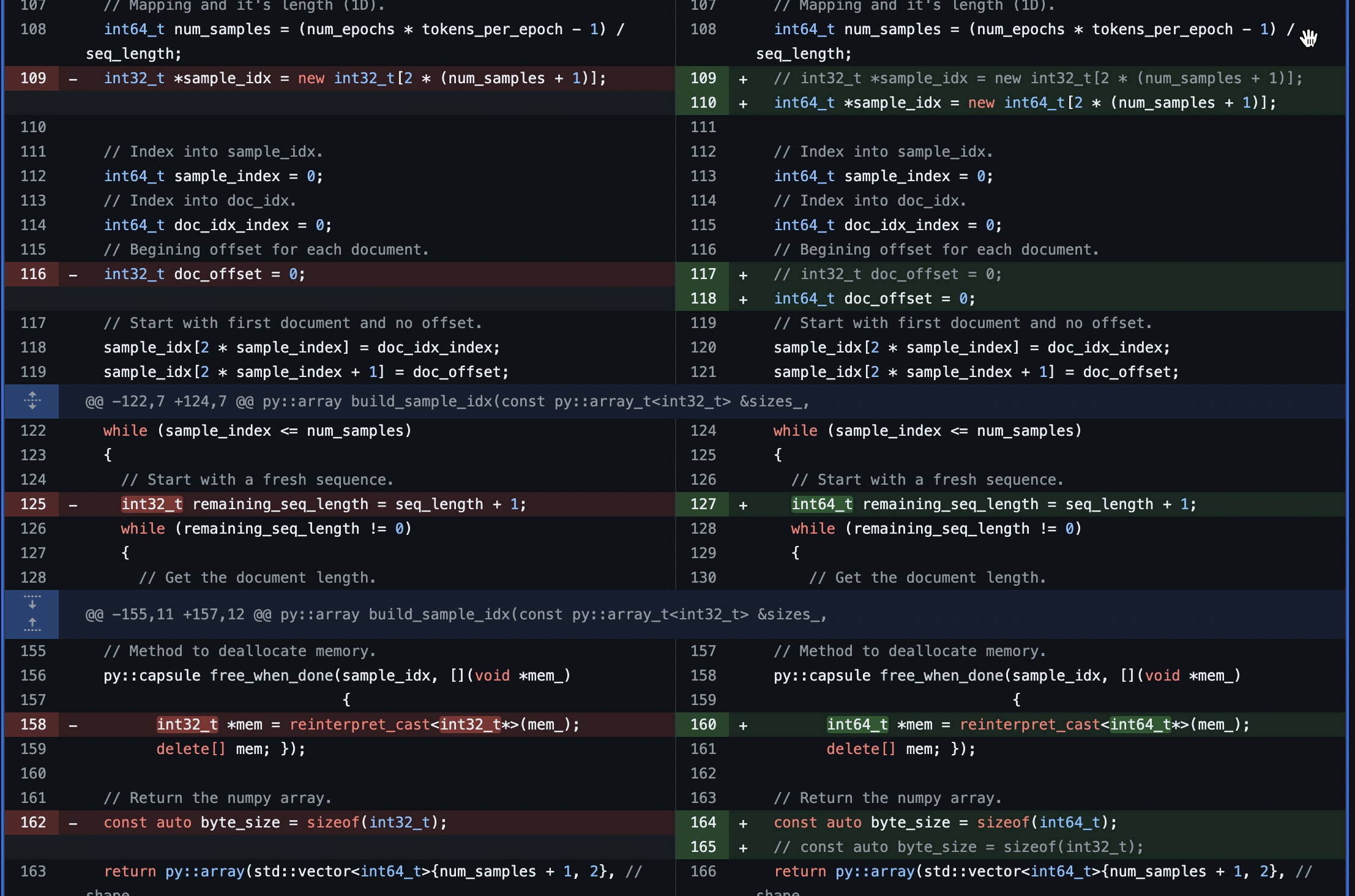}
\caption{
Code modification for our training framework modification.
}
\label{fig:code}
\end{figure}

In this section, we address an overflow problem within the Megatron-core. The issue arises when the variable num\_samples, defined as int64\_t, exceeds the capacity of int32\_t associated with the sample\_idx. This discrepancy can lead to memory leaks and undefined behavior, as illustrated in Fig. \ref{fig:code}.

\subsection{Detailed Prompts in Intermediate Checkpoints Evaluation}
Here we present the prompts used for each dataset in the intermediate checkpoints evaluation. Since the prompts for datasets (BoolQ, PIQA, SIQA, HellaSwag, WinoGrande, ARC easy and challenge, OpenBookQA, CommonsenseQA, MMLU, CEval, CMMLU) are concatenated with the questions and the answers (options), we have not listed them separately. Below, we provide the prompts used for other datasets.

\textbf{BoolQ}
\begin{lstlisting}
HUMAN: Passage: {passage}\nQuestion: {question}?
BOT: Answer: No/Yes
\end{lstlisting}

\textbf{HumanEval}
\begin{lstlisting}
HUMAN: Complete the following python code:\n{prompt}
\end{lstlisting}

\textbf{MBPP}
\begin{lstlisting}
HUMAN: You are an expert Python programmer, and here is your task: Write a function to find the similar elements from the given two tuple lists. Your code should pass these tests:\n\n assert similar_elements((3, 4, 5, 6),(5, 7, 4, 10)) == (4, 5)\n assert similar_elements((1, 2, 3, 4),(5, 4, 3, 7)) == (3, 4) \n assert similar_elements((11, 12, 14, 13),(17, 15, 14, 13)) == (13, 14) \n
BOT: [BEGIN]\n 'def similar_elements(test_tup1, test_tup2):\r\n  res = tuple(set(test_tup1) & set(test_tup2))\r\n  return (res)' \n[DONE] \n\n
HUMAN: You are an expert Python programmer, and here is your task: Write a python function to identify non-prime numbers. Your code should pass these tests:\n\n assert is_not_prime(2) == False \n assert is_not_prime(10) == True \n assert is_not_prime(35) == True \n
BOT: [BEGIN]\n 'import math\r\ndef is_not_prime(n):\r\n    result = False\r\n    for i in range(2,int(math.sqrt(n)) + 1):\r\n        if n % i == 0:\r\n            result = True\r\n    return result' \n[DONE] \n\n
HUMAN: You are an expert Python programmer, and here is your task: Write a function to find the largest integers from a given list of numbers using heap queue algorithm. Your code should pass these tests:\n\n assert heap_queue_largest( [25, 35, 22, 85, 14, 65, 75, 22, 58],3)==[85, 75, 65] \n assert heap_queue_largest( [25, 35, 22, 85, 14, 65, 75, 22, 58],2)==[85, 75] \n assert heap_queue_largest( [25, 35, 22, 85, 14, 65, 75, 22, 58],5)==[85, 75, 65, 58, 35] \n
BOT: [BEGIN]\n 'import heapq as hq\r\ndef heap_queue_largest(nums,n):\r\n  largest_nums = hq.nlargest(n, nums)\r\n  return largest_nums' \n[DONE] \n\n
HUMAN: You are an expert Python programmer, and here is your task: {text} Your code should pass these tests:\n\n {test_list}  \n
BOT: [BEGIN]\n
\end{lstlisting}

\textbf{NaturalQuestions}

\begin{lstlisting}
HUMAN: Answer these questions, your answer should be as simple as possible, start your answer with the prompt \'The answer is \'.\nQ: {question}?
BOT: A:
\end{lstlisting}

\textbf{Triviaqa}
\begin{lstlisting}
HUMAN: Answer these questions, your answer should be as simple as possible, start your answer with the prompt \'The answer is \'.\nQ: {question}?
BOT: A:
\end{lstlisting}

\textbf{SQuAD 2.0}

\begin{lstlisting}
HUMAN: {context}\nAccording to the above passage, answer the following question. If it is impossible to answer according to the passage, answer `impossible to answer`:\nQuestion: {question}
BOT: Answer:
\end{lstlisting}

\textbf{MATH}

\begin{lstlisting}
HUMAN: Problem:\nFind the domain of the expression $\\frac{{\sqrt{{x-2}}}}{{\sqrt{{5-x}}}}$.}}\nSolution:"
BOT: The expressions inside each square root must be non-negative. Therefore, $x-2 \ge 0$, so $x\ge2$, and $5 - x \ge 0$, so $x \le 5$. Also, the denominator cannot be equal to zero, so $5-x>0$, which gives $x<5$. Therefore, the domain of the expression is $\\boxed{{[2,5)}}$.\nFinal Answer: The final answer is $[2,5)$. I hope it is correct.\n
HUMAN: Problem:\nIf $\det \mathbf{{A}} = 2$ and $\det \mathbf{{B}} = 12,$ then find $\det (\mathbf{{A}} \mathbf{{B}}).$\nSolution:
BOT: We have that $\det (\mathbf{{A}} \mathbf{{B}}) = (\det \mathbf{{A}})(\det \mathbf{{B}}) = (2)(12) = \\boxed{{24}}.$\nFinal Answer: The final answer is $24$. I hope it is correct.\n
HUMAN: Problem:\nTerrell usually lifts two 20-pound weights 12 times. If he uses two 15-pound weights instead, how many times must Terrell lift them in order to lift the same total weight?\nSolution:
BOT: If Terrell lifts two 20-pound weights 12 times, he lifts a total of $2\cdot 12\cdot20=480$ pounds of weight. If he lifts two 15-pound weights instead for $n$ times, he will lift a total of $2\cdot15\cdot n=30n$ pounds of weight. Equating this to 480 pounds, we can solve for $n$: \\begin{{align*}} 30n&=480\\\\ \Rightarrow\qquad n&=480/30=\\boxed{{16}} \end{{align*}}\nFinal Answer: The final answer is $16$. I hope it is correct.\n
HUMAN: Problem:\nIf the system of equations: \\begin{{align*}} 6x-4y&=a,\\\\ 6y-9x &=b. \end{{align*}}has a solution $(x, y)$ where $x$ and $y$ are both nonzero, find $\\frac{{a}}{{b}},$ assuming $b$ is nonzero.\nSolution:
BOT: If we multiply the first equation by $-\\frac{{3}}{{2}}$, we obtain $$6y-9x=-\\frac{{3}}{{2}}a.$$Since we also know that $6y-9x=b$, we have $$-\\frac{{3}}{{2}}a=b\Rightarrow\\frac{{a}}{{b}}=\\boxed{{-\\frac{{2}}{{3}}}}.$$\nFinal Answer: The final answer is $-\\frac{{2}}{{3}}$. I hope it is correct.\n
HUMAN: Problem:\n{problem}\nSolution:\n
\end{lstlisting}

\textbf{GSM8K}

\begin{lstlisting}
HUMAN: Question: Angelo and Melanie want to plan how many hours over the next week they should study together for their test next week. They have 2 chapters of their textbook to study and 4 worksheets to memorize. They figure out that they should dedicate 3 hours to each chapter of their textbook and 1.5 hours for each worksheet. If they plan to study no more than 4 hours each day, how many days should they plan to study total over the next week if they take a 10-minute break every hour, include 3 10-minute snack breaks each day, and 30 minutes for lunch each day?\nLet's think step by step\nAnswer: ,

BOT: Angelo and Melanie think they should dedicate 3 hours to each of the 2 chapters, 3 hours x 2 chapters = 6 hours total.\nFor the worksheets they plan to dedicate 1.5 hours for each worksheet, 1.5 hours x 4 worksheets = 6 hours total.\nAngelo and Melanie need to start with planning 12 hours to study, at 4 hours a day, 12 / 4 = 3 days.\nHowever, they need to include time for breaks and lunch. Every hour they want to include a 10-minute break, so 12 total hours x 10 minutes = 120 extra minutes for breaks.\nThey also want to include 3 10-minute snack breaks, 3 x 10 minutes = 30 minutes.\nAnd they want to include 30 minutes for lunch each day, so 120 minutes for breaks + 30 minutes for snack breaks + 30 minutes for lunch = 180 minutes, or 180 / 60 minutes per hour = 3 extra hours.\nSo Angelo and Melanie want to plan 12 hours to study + 3 hours of breaks = 15 hours total.\nThey want to study no more than 4 hours each day, 15 hours / 4 hours each day = 3.75\nThey will need to plan to study 4 days to allow for all the time they need.\nThe answer is 4,

HUMAN :Question: Mark's basketball team scores 25 2 pointers, 8 3 pointers and 10 free throws.  Their opponents score double the 2 pointers but half the 3 pointers and free throws.  What's the total number of points scored by both teams added together?\nLet's think step by step\nAnswer:,

BOT: Mark's team scores 25 2 pointers, meaning they scored 25*2= 50 points in 2 pointers.\nHis team also scores 6 3 pointers, meaning they scored 8*3= 24 points in 3 pointers\nThey scored 10 free throws, and free throws count as one point so they scored 10*1=10 points in free throws.\nAll together his team scored 50+24+10= 84 points\nMark's opponents scored double his team's number of 2 pointers, meaning they scored 50*2=100 points in 2 pointers.\nHis opponents scored half his team's number of 3 pointers, meaning they scored 24/2= 12 points in 3 pointers.\nThey also scored half Mark's team's points in free throws, meaning they scored 10/2=5 points in free throws.\nAll together Mark's opponents scored 100+12+5=117 points\nThe total score for the game is both team's scores added together, so it is 84+117=201 points\nThe answer is 201\n,

HUMAN: Question: Bella has two times as many marbles as frisbees. She also has 20 more frisbees than deck cards. If she buys 2/5 times more of each item, what would be the total number of the items she will have if she currently has 60 marbles?\nLet's think step by step\nAnswer:,

BOT: When Bella buys 2/5 times more marbles, she'll have increased the number of marbles by 2/5*60 = 24\nThe total number of marbles she'll have is 60+24 = 84\nIf Bella currently has 60 marbles, and she has two times as many marbles as frisbees, she has 60/2 = 30 frisbees.\nIf Bella buys 2/5 times more frisbees, she'll have 2/5*30 = 12 more frisbees.\nThe total number of frisbees she'll have will increase to 30+12 = 42\nBella also has 20 more frisbees than deck cards, meaning she has 30-20 = 10 deck cards\nIf she buys 2/5 times more deck cards, she'll have 2/5*10 = 4 more deck cards.\nThe total number of deck cards she'll have is 10+4 = 14\nTogether, Bella will have a total of 14+42+84 = 140 items\nThe answer is 140\n,

HUMAN: Question: A group of 4 fruit baskets contains 9 apples, 15 oranges, and 14 bananas in the first three baskets and 2 less of each fruit in the fourth basket. How many fruits are there?\nLet's think step by step\nAnswer:,

BOT: For the first three baskets, the number of apples and oranges in one basket is 9+15=24\nIn total, together with bananas, the number of fruits in one basket is 24+14=38 for the first three baskets.\nSince there are three baskets each having 38 fruits, there are 3*38=114 fruits in the first three baskets.\nThe number of apples in the fourth basket is 9-2=7\nThere are also 15-2=13 oranges in the fourth basket\nThe combined number of oranges and apples in the fourth basket is 13+7=20\nThe fourth basket also contains 14-2=12 bananas.\nIn total, the fourth basket has 20+12=32 fruits.\nThe four baskets together have 32+114=146 fruits.\nThe answer is 146\n,

HUMAN: Question: {question}\nLet's think step by step\nAnswer:

\end{lstlisting}

\textbf{TheoremQA}

\begin{lstlisting}
HUAMN: You are a mathematician, you are supposed to answer the given question. You need to output the answer in your final sentence like "Therefore, the answer is ...". The answer can only be one of the following forms:\n1. a numerical value like 0.1, no symbol and no unit at all.\n2. a list of number like [2, 3, 4].\n3. True/False.\n4. an option like (a), (b), (c), (d)\nQuestion: {Question}\nLet\'s think step by step.
\end{lstlisting}

\clearpage
\subsection{Detailed Results}
The evaluation results of all intermediate checkpoints are obtained using the OpenCompass framework~\citep{2023opencompass}.

\begin{table}[h]
\centering
\caption{This table showcases evaluation results across a variety of datasets for models trained with different token amounts, ranging from 20B to 1859.86B. Additionally, results for models trained with 2099.84B to 3726.33B tokens can be found in Table~\ref{tab:intermediate-ckpt-part2}. ``Avg'' represents the average over the benchmark. The ``*'' symbol refers to subsets within the MMLU, CMMLU, and C-Eval.}\label{tab:intermediate-ckpt-part1}
\resizebox{\textwidth}{!}{
\begin{tabular}{ccccccccccc}
\midrule

                         \textbf{Dataset} & \textbf{20.00B} & \textbf{60.00B} & \textbf{99.99B} & \textbf{359.97B} & \textbf{599.95B} & \textbf{859.93B} & \textbf{1099.91B} & \textbf{1299.90B} & \textbf{1599.88B} & \textbf{1859.86B} \\ \midrule

\multicolumn{11}{c}{\textbf{ Standard Benchmarks}}\\ \midrule
                         \textbf{BoolQ} &           58.81 &           60.28 &           58.96 &             61.9 &            61.62 &            62.29 &             62.35 &             63.67 &             59.02 &             61.35\\ \midrule
                         \textbf{PIQA} &           67.25 &           70.35 &           73.56 &            76.06 &            76.12 &            77.64 &             77.75 &             77.58 &             77.58 &             77.91\\ \midrule
                         \textbf{SIQA} &           38.33 &           41.04 &           40.43 &            41.71 &             41.4 &            42.48 &             42.99 &             42.99 &             42.43 &             44.06\\ \midrule
                    \textbf{HellaSwag} &           32.53 &           47.07 &           52.03 &            61.32 &            63.61 &            64.83 &             65.75 &             66.11 &             67.35 &             67.69\\ \midrule
                    \textbf{WinoGrande} &           52.09 &           53.12 &            53.2 &            55.25 &            55.41 &            57.38 &             57.93 &             58.09 &             58.09 &             59.67\\ \midrule
                         \textbf{ARC-e} &           35.27 &           43.39 &           51.15 &             57.5 &            57.32 &             57.5 &             58.02 &             58.91 &             62.08 &             60.14\\ \midrule
                         \textbf{ARC-c} &           23.39 &              20 &           23.73 &            29.49 &            28.14 &            31.86 &              32.2 &              32.2 &             32.54 &             33.22\\ \midrule
               \textbf{OpenBookQA-Fact} &            26.2 &            23.8 &            23.8 &             28.8 &             43.6 &             48.6 &              51.8 &              59.6 &                62 &                70\\ \midrule
               \textbf{CommonsenseQA} &           34.32 &           48.32 &           51.43 &            59.54 &            61.43 &            63.72 &             66.09 &             64.95 &             65.19 &             65.44\\ \midrule
                         \textbf{MMLU-AVG} &            24.8 &           24.38 &           26.72 &            36.06 &            43.92 &            47.32 &             47.96 &             50.65 &             51.18 &             51.95\\ \midrule
                         \textbf{*-humanities} &            24.5 &           25.25 &           26.71 &             37.5 &            44.58 &            49.26 &             50.58 &             52.92 &             53.62 &             54.29\\ \midrule
                         \textbf{*-stem} &            24.4 &           23.26 &           26.76 &            30.83 &            36.82 &            39.98 &             40.89 &              42.7 &             42.72 &             44.24\\ \midrule
               \textbf{*-social-science} &            22.8 &           23.58 &            26.9 &             39.7 &            49.07 &            53.77 &             53.71 &             57.75 &             58.93 &             58.73\\ \midrule
                         \textbf{*-other} &           27.52 &           25.87 &           26.52 &            38.89 &             48.9 &            50.15 &             50.37 &             53.42 &             53.94 &             54.62\\ \midrule
                         \multicolumn{11}{c}{\textbf{ Code Generation}}\\ \midrule
               \textbf{HumanEval} &            0.61 &            2.44 &            4.27 &              6.1 &             7.32 &             7.93 &              7.32 &              7.32 &              9.15 &               6.1\\ \midrule
                         \textbf{MBPP} &               0 &             0.4 &               0 &              3.4 &              6.6 &              6.4 &               9.2 &               9.4 &               8.8 &               6.6\\ \midrule
\multicolumn{11}{c}{\textbf{ World Knowledge}}\\ \midrule
                              \textbf{NQ} &            0.08 &            1.55 &             2.8 &              5.1 &             5.79 &             7.51 &              7.84 &              9.34 &              9.03 &              8.01\\ \midrule
                    \textbf{TriviaQA} &             1.2 &             6.9 &            9.54 &            19.64 &            25.97 &            22.24 &              28.6 &             28.22 &             34.19 &             31.31\\ \midrule
\multicolumn{11}{c}{\textbf{ Reading Comprehension}}\\ \midrule
                    \textbf{SQuAD2.0} &            4.54 &           15.94 &            24.2 &            27.06 &            31.05 &            31.48 &             30.68 &             12.56 &             31.35 &             25.76\\ \midrule
\multicolumn{11}{c}{\textbf{ Exams}}\\ \midrule
                         \textbf{MATH} &             0.6 &            1.22 &            1.16 &             2.62 &              2.8 &             3.18 &               3.6 &              3.82 &              3.44 &              4.24\\ \midrule
                         \textbf{GSM8k} &             1.59 &             0.76 &             0.99 &              4.09 &              7.66 &            9.78 &             12.05 &             11.52 &             15.24 &             14.48\\ \midrule
                         \textbf{TheoremQA} &               0 &               0 &             0.5 &             0.75 &             1.38 &              1.5 &              1.38 &              0.75 &              0.62 &              0.38\\ \midrule
\multicolumn{11}{c}{\textbf{ Chinese}}\\ \midrule
                         \textbf{C-EVAL-AVG} &           24.87 &           24.66 &           25.48 &            36.55 &             44.3 &             46.9 &             50.01 &              52.1 &              52.4 &             52.95\\ \midrule
                    \textbf{*-stem} &            26.8 &           24.04 &           24.47 &            31.43 &            35.45 &             38.5 &             39.86 &             42.67 &             45.14 &             45.49\\ \midrule
          \textbf{*-social-science} &           26.99 &           29.16 &           27.19 &            47.15 &            56.78 &             61.6 &             66.94 &             68.71 &             68.29 &             67.01\\ \midrule
               \textbf{*-humanities} &            24.5 &           22.64 &           25.42 &            41.47 &            49.66 &               49 &             53.04 &             58.19 &              56.2 &             58.41\\ \midrule
                    \textbf{*-other} &           19.82 &           23.72 &           25.82 &            31.29 &            43.67 &            46.71 &             50.04 &             48.06 &             47.33 &             48.28\\ \midrule
                    \textbf{*-hard} &           30.97 &           23.78 &           21.87 &            25.69 &            28.04 &             31.1 &             36.06 &              37.5 &             33.66 &             38.08\\ \midrule
                         \textbf{CMMLU-AVG} &           25.11 &           25.18 &           25.96 &            35.48 &            42.93 &            47.54 &             48.85 &             50.14 &             50.94 &             52.18\\ \midrule
               \textbf{*-humanities} &           25.54 &           25.62 &           25.79 &            38.44 &            47.19 &            50.58 &             51.76 &             54.35 &             54.22 &             55.55\\ \midrule
                    \textbf{*-stem} &           24.96 &           24.26 &           25.15 &            28.82 &            34.34 &             38.7 &             39.26 &             39.23 &             40.92 &             42.79\\ \midrule
          \textbf{*-social-science} &           25.05 &           25.91 &           26.78 &            38.72 &            46.14 &            51.96 &             53.18 &             54.53 &             55.26 &             56.85\\ \midrule
                    \textbf{*-other} &           24.99 &           24.76 &           25.83 &            35.69 &            44.29 &            48.44 &             50.83 &             52.41 &             53.11 &             53.05\\ \midrule
          \textbf{*-china-specific} &            24.4 &           25.62 &           25.14 &            38.02 &            43.86 &            48.96 &             50.14 &             52.46 &             53.15 &             54.03\\ \midrule

\end{tabular}
}
\end{table}

\clearpage

\begin{table}[h]
\centering
\caption{This table showcases evaluation results across a variety of datasets for models trained with different token amounts, ranging from 2099.84B to 3726.33B. Additionally, results for models trained with 20B to 1859.86B tokens can be found in Table~\ref{tab:intermediate-ckpt-part1}. ``Avg'' represents the average over the benchmark. The ``*'' symbol refers to subsets within the MMLU, CMMLU, and C-Eval.}\label{tab:intermediate-ckpt-part2}
\resizebox{\textwidth}{!}{
\begin{tabular}{ccccccccccccccccccc}
\midrule

                         \textbf{Dataset} & \textbf{2099.84B} & \textbf{2359.82B} & \textbf{2599.80B} & \textbf{2859.78B} & \textbf{3099.76B} & \textbf{3299.74B} & \textbf{3599.72B} & \textbf{3726.33B} \\ \midrule

\multicolumn{19}{c}{\textbf{ Standard Benchmarks}}\\ \midrule
                         \textbf{BoolQ} &           60.89 &             63.12 &             59.36 &             64.56 &             63.67 &             63.18 &             65.35 &             66.09 \\ \midrule
                         \textbf{PIQA} &           77.91 &             78.02 &             78.35 &             78.56 &             78.94 &             78.67 &             78.13 &             78.29 \\ \midrule
                         \textbf{SIQA} &           44.06 &             44.06 &              43.3 &             43.71 &             44.01 &             43.45 &             44.63 &             43.19 \\ \midrule
                    \textbf{HellaSwag} &           68.6 &             68.52 &             69.04 &             69.78 &             70.06 &             70.02 &             70.46 &             70.17 \\ \midrule
                    \textbf{WinoGrande} &           58.96 &             59.83 &             59.91 &             60.06 &             61.25 &             59.75 &             59.67 &             60.46 \\ \midrule
                         \textbf{ARC-e} &           62.43 &             62.61 &             63.32 &             61.73 &             61.73 &             62.26 &             62.43 &             64.02 \\ \midrule
                         \textbf{ARC-c} &           23.39 &              20 &           37.29 &             35.93 &             35.59 &             36.95 &             34.58 &             34.58 \\ \midrule
               \textbf{OpenBookQA-Fact} &           63.8 &              61.6 &                60 &                66 &              59.6 &              59.4 &                69 &              62.2 \\ \midrule
               \textbf{CommonsenseQA} &           66.42 &             65.77 &             67.32 &             67.98 &             67.57 &             67.57 &             67.81 &             67.73 \\ \midrule
                         \textbf{MMLU-AVG} &            52.72 &             53.25 &             53.93 &             54.71 &             55.34 &              55.8 &             55.42 &             55.91 \\ \midrule
                         \textbf{*-humanities} &            54.18 &             56.75 &              57.2 &             57.25 &             58.34 &             58.19 &             58.71 &             59.22 \\ \midrule
                         \textbf{*-stem} &            44.48 &             44.08 &             44.19 &             45.56 &             45.92 &             46.53 &             46.42 &             46.37 \\ \midrule
               \textbf{*-social-science} &            60.72 &             61.02 &             61.87 &             63.45 &             63.87 &             64.86 &             62.63 &             63.72 \\ \midrule
                         \textbf{*-other} &           55.9 &             55.96 &             57.58 &             57.49 &             58.23 &             58.61 &             58.61 &             59.33 \\ \midrule
                         \multicolumn{19}{c}{\textbf{ Code Generation}}\\ \midrule
               \textbf{HumanEval} &            8.54 &              3.66 &              6.71 &              6.71 &               7.32 &               3.66 &               9.76 &               9.15 \\ \midrule
                         \textbf{MBPP} &          8.4 &               9.4 &               8.8 &               8.6 &               8.8 &               8.8 &               9.2 &                 9 \\ \midrule
\multicolumn{19}{c}{\textbf{ World Knowledge}}\\ \midrule
                              \textbf{NQ} &            10.97 &             10.19 &             10.03 &             11.77 &             10.66 &             12.63 &             11.44 &             11.27 \\ \midrule
                    \textbf{TriviaQA} &             36.53 &             31.06 &              37.9 &             39.29 &             40.81 &             41.27 &             41.08 &             39.54 \\ \midrule
\multicolumn{19}{c}{\textbf{ Reading Comprehension}}\\ \midrule
                    \textbf{SQuAD2.0} &            25.29 &             26.98 &             11.35 &              5.13 &              6.18 &             16.68 &             15.55 &              8.72 \\ \midrule
\multicolumn{19}{c}{\textbf{ Exams}}\\ \midrule
                         \textbf{MATH} &             4.84 &              4.34 &              4.94 &              5.36 &               5.6 &              5.72 &               5.9 &              5.76 \\ \midrule
                         \textbf{GSM8k} &             14.94 &             17.36 &             17.29 &             18.95 &             19.18 &              19.79 &             19.11 &             21.3 \\ \midrule
                         \textbf{TheoremQA} &          1.38 &               0.5 &                 1 &                 3 &              2.38 &                 2 &               1.5 &               2.5 \\ \midrule
\multicolumn{19}{c}{\textbf{ Chinese}}\\ \midrule
                         \textbf{C-EVAL-AVG} &           52.52 &             55.62 &              57.4 &             57.03 &             56.02 &             57.57 &              58.1 &             57.13 \\ \midrule
                    \textbf{*-stem} &            44.77 &             49.52 &             51.84 &             49.08 &             46.52 &             50.26 &             50.26 &             49.47 \\ \midrule
          \textbf{*-social-science} &           66.71 &              70.3 &             71.62 &             71.43 &             70.74 &             73.33 &             73.05 &             72.61 \\ \midrule
               \textbf{*-humanities} &          58.08 &              58.6 &             61.96 &             62.09 &             61.14 &             60.96 &             61.46 &             62.06 \\ \midrule
                    \textbf{*-other} &           48.18 &             50.39 &             50.03 &             53.35 &             54.79 &             53.12 &             55.41 &             52.03 \\ \midrule
                    \textbf{*-hard} &           34.8 &             39.89 &             44.08 &             39.87 &             36.42 &             41.26 &             38.64 &             40.47 \\ \midrule
                         \textbf{CMMLU-AVG} &           52.45 &             54.79 &             56.15 &             56.63 &             57.33 &             58.11 &              57.7 &             58.32 \\ \midrule
               \textbf{*-humanities} &           56.42 &             60.23 &             60.97 &             61.09 &                63 &             63.69 &             63.73 &             65.04 \\ \midrule
                    \textbf{*-stem} &           42.17 &             44.38 &             45.95 &             46.16 &             46.54 &             47.82 &             46.63 &             47.91 \\ \midrule
          \textbf{*-social-science} &           57.34 &             59.26 &             60.42 &             61.79 &             62.27 &             62.93 &             62.83 &             63.25 \\ \midrule
                    \textbf{*-other} &           53.51 &             55.34 &             57.27 &             57.09 &              57.4 &             57.84 &              57.5 &             57.05 \\ \midrule
          \textbf{*-china-specific} &            54.82 &             56.86 &             58.15 &             58.07 &             59.51 &             60.47 &             60.74 &             60.29 \\ \midrule

\end{tabular}
}

\end{table}

\begin{table}[H]
\centering
\caption{This table shows the evaluation results across a variety of datasets for models of different train tokens in the decay phase, from 62.91B to 723.52B. ``Avg'' represents the average over the benchmark. The ``*'' symbol refers to subsets within the MMLU, CMMLU, and C-Eval.}\label{tab:decay-ckpts}
\resizebox{\textwidth}{!}{
\begin{tabular}{ccccccccc}
\midrule

                           \textbf{Dataset} & \textbf{62.91B} & \textbf{104.86B} & \textbf{199.23B} & \textbf{293.60B} & \textbf{419.43B} & \textbf{524.29B} & \textbf{639.63B} & \textbf{723.52B} \\ \midrule

      \multicolumn{9}{c}{\textbf{ Standard Benchmarks}}\\ \midrule
                             \textbf{BoolQ} &           52.51 &            50.73 &               47 &             63.7 &            65.38 &            78.32 &            70.34 &            81.07 \\ \midrule
                              \textbf{PIQA} &           75.41 &            75.41 &            75.73 &            76.71 &            77.04 &             75.9 &             75.3 &            76.55 \\ \midrule
                              \textbf{SIQA} &           48.36 &            49.13 &            50.31 &            50.31 &            51.28 &            69.45 &            68.73 &            68.22 \\ \midrule
                         \textbf{HellaSwag} &           62.79 &            63.98 &            65.19 &            66.17 &            66.43 &            69.57 &               70 &            70.74 \\ \midrule
                        \textbf{WinoGrande} &           62.04 &            63.69 &            64.01 &            65.75 &            66.06 &            59.43 &            59.59 &            59.83 \\ \midrule
                             \textbf{ARC-c} &           58.64 &            60.34 &            63.73 &            61.36 &            68.47 &            45.42 &            63.39 &            68.14 \\ \midrule
                   \textbf{OpenBookQA-Fact} &            75.6 &             76.2 &             77.2 &             74.2 &               79 &             79.6 &             73.4 &               82 \\ \midrule
                     \textbf{CommonsenseQA} &           60.36 &            63.06 &            63.72 &            64.54 &            63.14 &            68.96 &             69.7 &            69.94 \\ \midrule
                              \textbf{MMLU-AVG} &           52.53 &            53.31 &            54.83 &            55.51 &            56.11 &            57.17 &            57.36 &            58.14 \\ \midrule
                              \textbf{*-humanities} &           54.59 &            57.44 &             57.8 &            58.12 &             59.5 &            60.76 &            59.77 &             60.7 \\ \midrule
                         \textbf{*-stem} &           45.68 &            45.58 &            48.37 &            47.29 &            48.48 &            49.82 &            49.31 &            49.84 \\ \midrule
               \textbf{*-social-science} &            59.6 &            60.69 &            61.19 &            63.95 &            64.45 &            64.78 &            65.27 &            66.78 \\ \midrule
                        \textbf{*-other} &           53.94 &            53.65 &            55.43 &            57.14 &            56.17 &            57.31 &            59.42 &            59.73 \\ \midrule
    \multicolumn{9}{c}{\textbf{ Code Generation}}\\  \midrule
                  \textbf{HumanEval} &             6.1 &             7.32 &             3.05 &            11.59 &             0.61 &            21.95 &            20.12 &            24.39 \\ \midrule
                              \textbf{MBPP} &            20.8 &               25 &             24.8 &             28.2 &             28.4 &               27 &             27.8 &               27 \\ \midrule
    \multicolumn{9}{c}{\textbf{ World Knowledge}}\\  \midrule
                                \textbf{NQ} &            4.04 &              6.4 &             5.43 &             5.04 &             3.21 &             9.94 &             8.23 &             9.97 \\ \midrule
                          \textbf{TriviaQA} &           15.27 &            27.31 &            24.76 &            34.32 &            44.03 &             37.5 &            32.66 &            42.36 \\ \midrule
    \multicolumn{9}{c}{\textbf{ Reading Comprehension}}\\  \midrule
                          \textbf{SQuAD2.0} &           33.72 &            13.57 &             27.1 &            30.89 &             16.8 &            29.56 &            19.37 &            30.98 \\ \midrule
    \multicolumn{9}{c}{\textbf{ Exams}}\\  \midrule
                              \textbf{MATH} &            6.62 &             8.62 &            10.08 &            12.88 &            12.24 &            14.06 &            14.12 &            14.66 \\ \midrule
                             \textbf{GSM8K} &           18.35 &            37.83 &            41.85 &            45.03 &            49.43 &            50.64 &            53.37 &            52.01 \\ \midrule
   \multicolumn{9}{c}{\textbf{ Chinese}}\\  \midrule
                             \textbf{C-EVAL-AVG} &           48.96 &            51.29 &            53.66 &            54.96 &            55.71 &            57.58 &            54.25 &            57.68 \\ \midrule
                        \textbf{*-stem} &           43.61 &            45.69 &            49.82 &            47.14 &            49.71 &            52.12 &            45.77 &            50.35 \\ \midrule
              \textbf{*-social-science} &           60.77 &            66.43 &            66.94 &               69 &             67.7 &            70.33 &            71.08 &            70.23 \\ \midrule
                  \textbf{*-humanities} &           50.55 &             51.2 &            53.26 &            60.74 &            59.11 &            63.29 &            59.05 &            63.49 \\ \midrule
                       \textbf{*-other} &           46.37 &            47.77 &            48.95 &            50.63 &             52.3 &            50.22 &            49.55 &            53.78 \\ \midrule
                        \textbf{*-hard} &           35.01 &            38.33 &            42.48 &            39.61 &            41.27 &            46.02 &            37.02 &            41.07 \\ \midrule
                             \textbf{CMMLU-AVG} &           48.03 &             49.1 &            51.37 &            53.26 &            53.32 &            54.48 &            54.59 &             55.1 \\ \midrule
                  \textbf{*-humanities} &           52.86 &            53.77 &            56.23 &            58.01 &            59.12 &            60.08 &            61.75 &            62.24 \\ \midrule
                        \textbf{*-stem} &           39.16 &            40.38 &            43.52 &            43.64 &            44.01 &            45.89 &            45.23 &            45.62 \\ \midrule
              \textbf{*-social-science} &           52.01 &            52.89 &            55.19 &            57.57 &            57.53 &            58.36 &            58.31 &            59.39 \\ \midrule
                       \textbf{*-other} &           48.04 &            49.37 &            50.45 &            53.72 &            52.65 &            53.66 &            53.55 &            53.39 \\ \midrule
              \textbf{*-china-specific} &           47.63 &            48.99 &            51.51 &            52.74 &            53.57 &            54.78 &            54.87 &            55.84 \\ \midrule

\end{tabular}
}

\end{table}

\begin{figure}[htbp]
    \centering    
    \begin{subfigure}[b]{0.3\textwidth}
        \includegraphics[width=\textwidth]{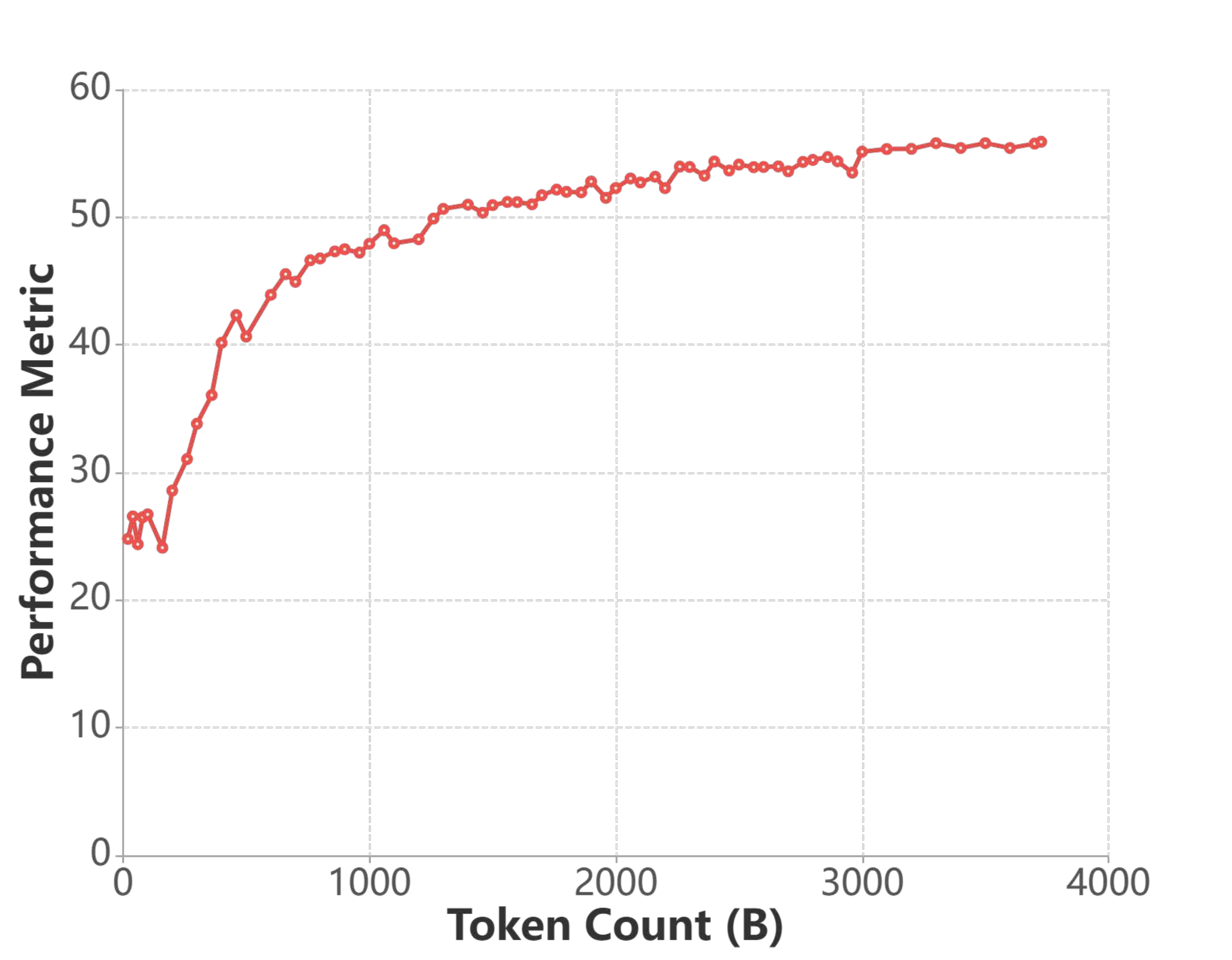}
        \caption{MMLU}
        \label{fig:mmlu}
    \end{subfigure}
    \hfill
    \begin{subfigure}[b]{0.3\textwidth}
        \includegraphics[width=\textwidth]{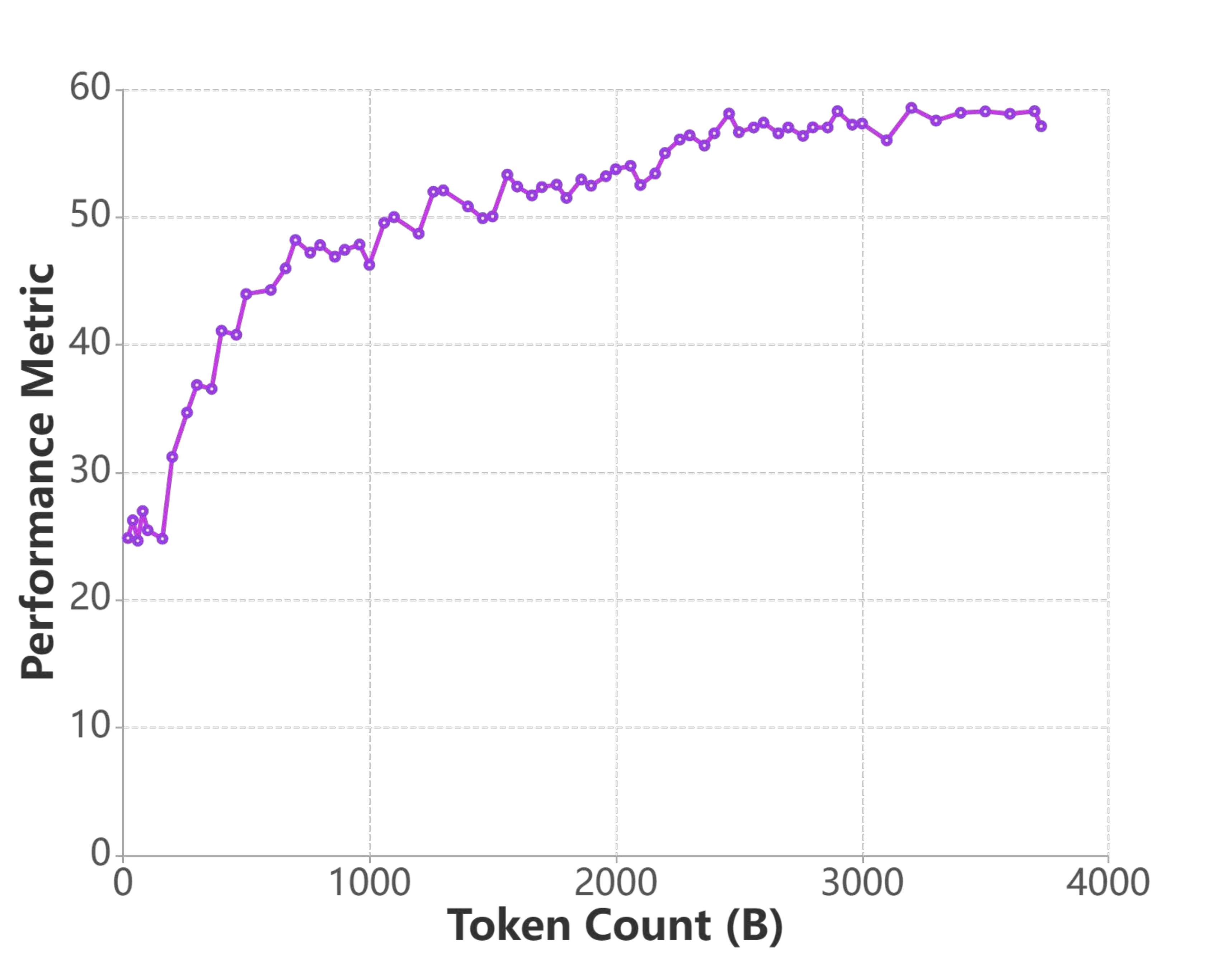}
        \caption{CEval}
        \label{fig:ceval}
    \end{subfigure}
    \hfill
    \begin{subfigure}[b]{0.3\textwidth}
        \includegraphics[width=\textwidth]{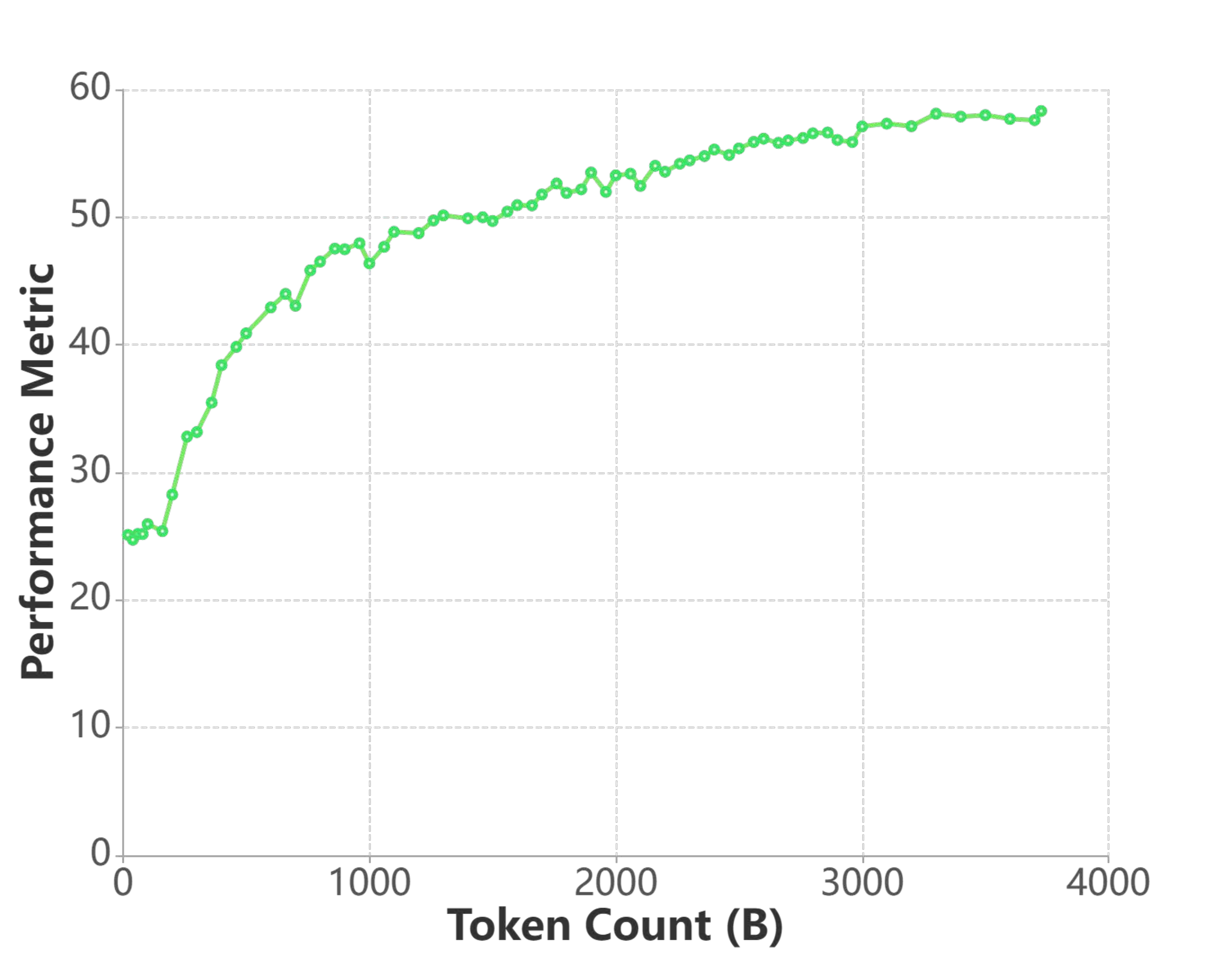}
        \caption{CMMLU}
        \label{fig:cmmlu}
    \end{subfigure}
    
    \begin{subfigure}[b]{0.3\textwidth}
        \includegraphics[width=\textwidth]{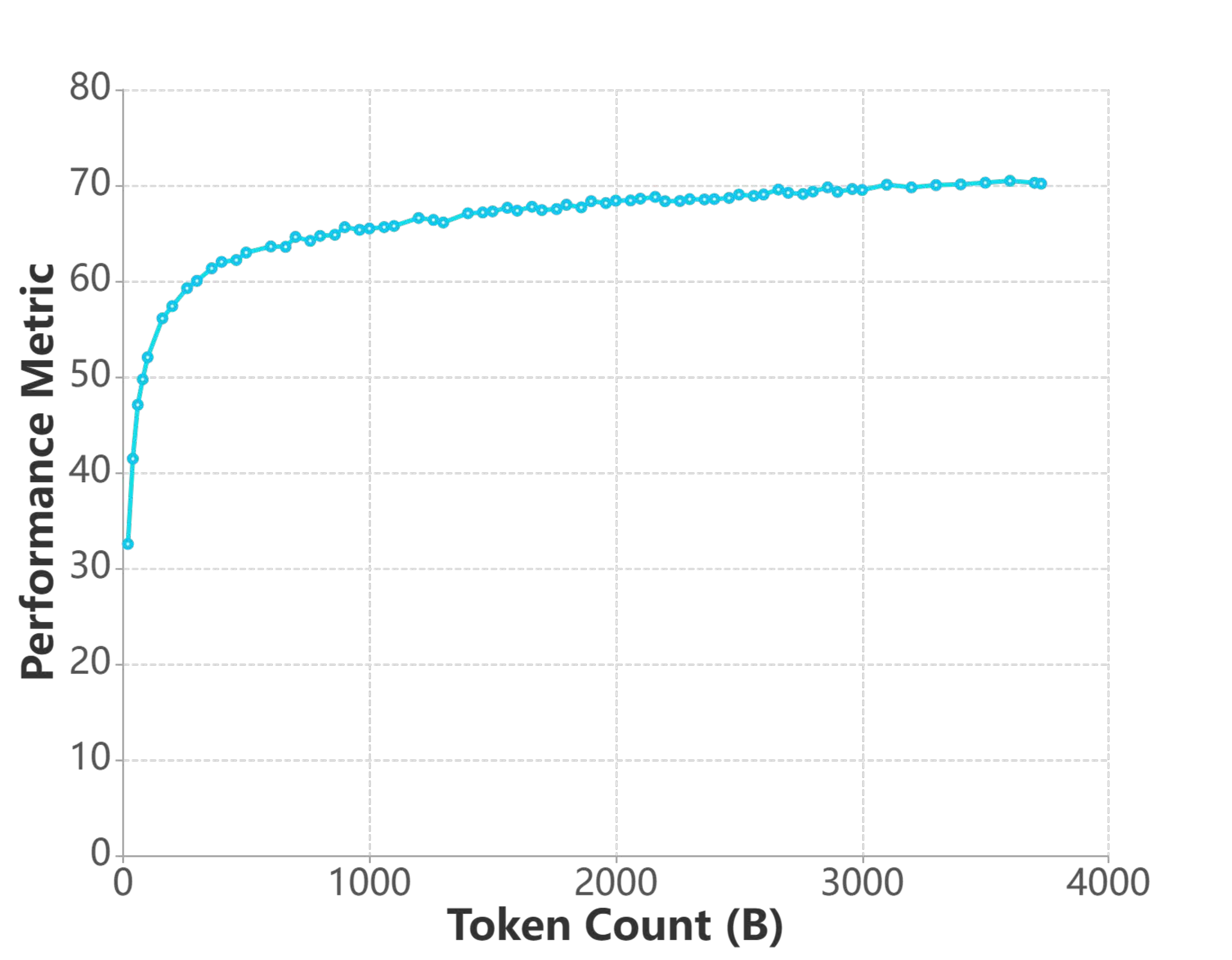}
        \caption{Hellaswag}
        \label{fig:hellaswag}
    \end{subfigure}
    \hfill
    \begin{subfigure}[b]{0.3\textwidth}
        \includegraphics[width=\textwidth]{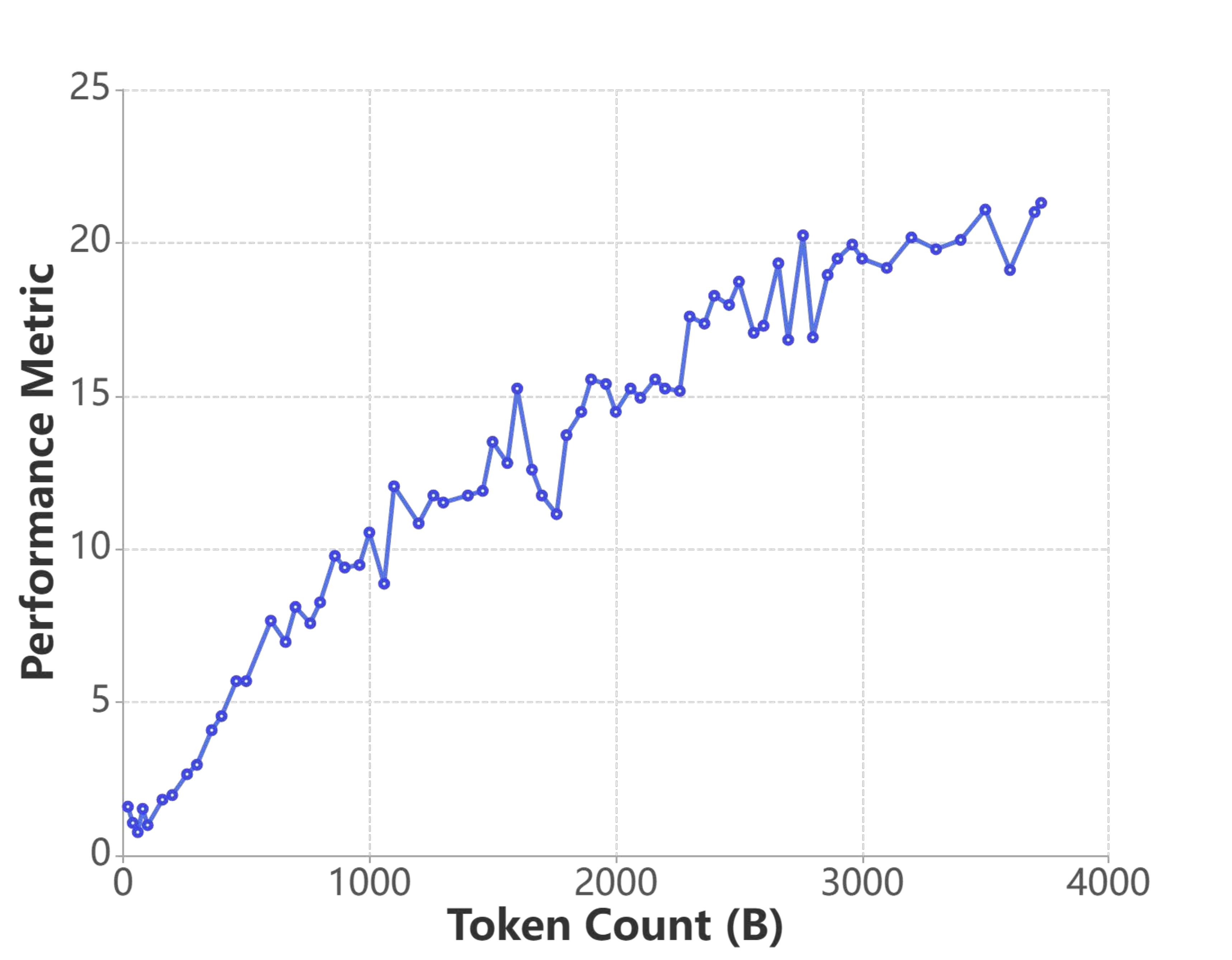}
        \caption{GSM8K}
        \label{fig:gsm8k}
    \end{subfigure}
    \hfill
    \begin{subfigure}[b]{0.3\textwidth}
        \includegraphics[width=\textwidth]{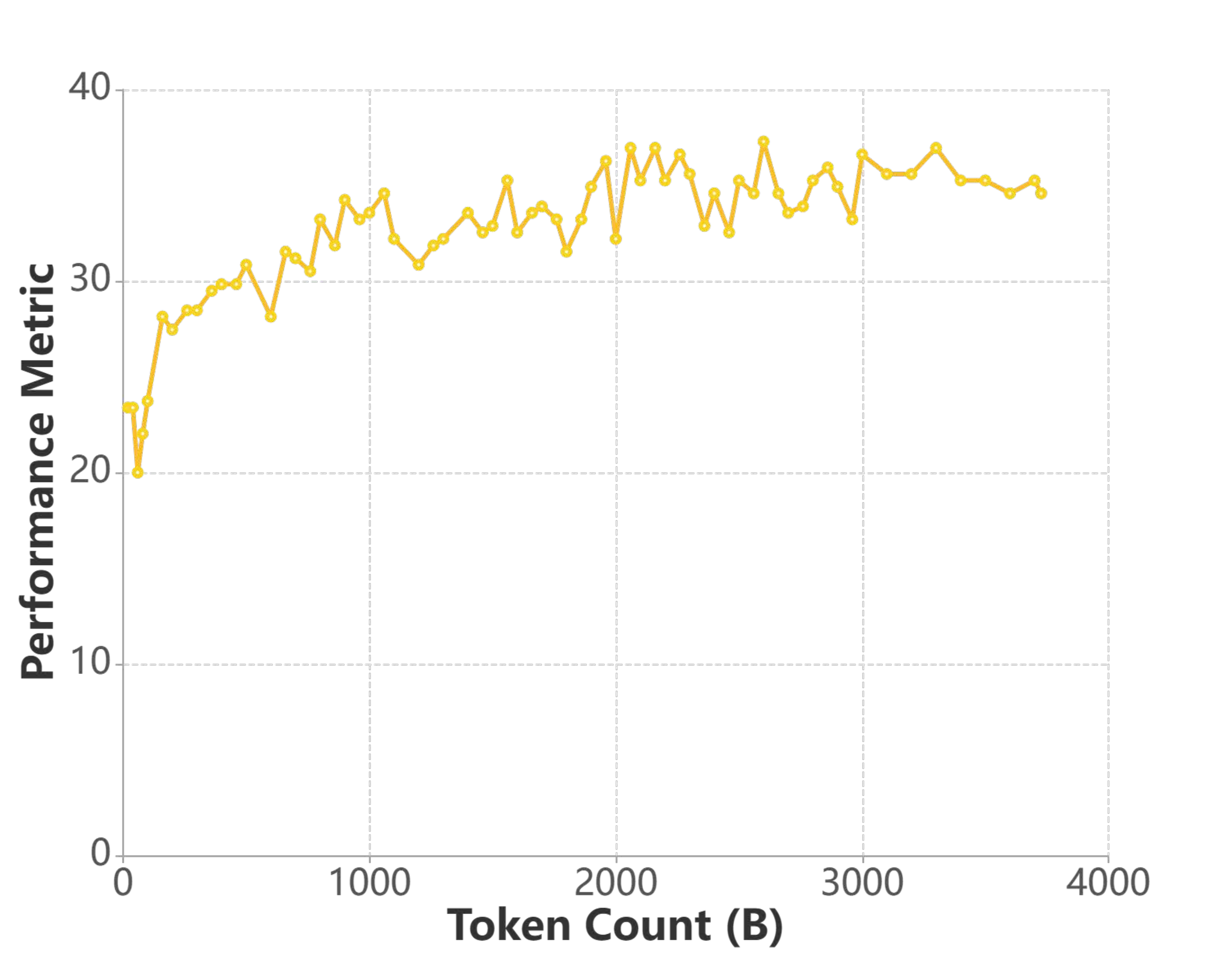}
        \caption{ARC-c}
        \label{fig:ARC-c}
    \end{subfigure}
\caption{Performance of fundamental phase intermediate checkpoints on MMLU, CEvel, CMMLU, Hellaswag, GSM8K, and ARC-c.}
    \label{fig:performance-line}
\end{figure}

\subsection{Details of Open Source Datasets Used in Pre-training}
\begin{table}[ht]
\centering
\caption{List of open-source datasets used during pretraining.}
\label{tab:open-source-dataset-list}
\resizebox{\textwidth}{!}{
\begin{tabular}{ll}
\toprule
                      Dataset &                                                URL \\
\midrule
                     Agent-FLAN~\citep{chen2024agent} & \url{https://huggingface.co/datasets/internlm/Agent-FLAN} \\
ChatDoctor-HealthCareMagic-100k & \url{https://huggingface.co/datasets/lavita/ChatDoctor-HealthCareMagic-100k} \\
                      Fandom23K~\citep{ryokoai2023-bigknow2022} &  \url{https://huggingface.co/datasets/RyokoAI/Fandom23K} \\
                   LoC-PD-Books & \url{https://huggingface.co/datasets/storytracer/LoC-PD-Books} \\
                          MNBVC &         \url{https://huggingface.co/datasets/liwu/MNBVC} \\
             Refined-Anime-Text & \url{https://huggingface.co/datasets/CausalLM/Refined-Anime-Text} \\
           SKGInstruct-skg-only~\citep{zhuang2024structlm} & \url{https://huggingface.co/datasets/TIGER-Lab/SKGInstruct-skg-only} \\
                    US-PD-Books & \url{https://huggingface.co/datasets/storytracer/US-PD-Books} \\
                 UltraTextbooks & \url{https://huggingface.co/datasets/Locutusque/UltraTextbooks} \\
                     big\_patent~\citep{DBLP:journals/corr/abs-1906-03741} &         \url{https://huggingface.co/datasets/big\_patent} \\
       clean\_notebooks\_filtered & \url{https://huggingface.co/datasets/vikp/clean\_notebooks\_filtered} \\
           libre\_chem\_textbooks & \url{https://huggingface.co/datasets/Hack90/libre\_chem\_textbooks} \\
  mental\_health\_chatbot\_dataset & \url{https://huggingface.co/datasets/heliosbrahma/mental\_health\_chatbot\_dataset} \\
                     mini-peS2o & \url{https://huggingface.co/datasets/nampdn-ai/mini-peS2o} \\
                      textbooks & \url{https://huggingface.co/datasets/open-phi/textbooks} \\
                    pile-of-law~\citep{hendersonkrass2022pileoflaw} & \url{https://huggingface.co/datasets/pile-of-law/pile-of-law} \\
          prepared-automathtext & \url{https://huggingface.co/datasets/Locutusque/prepared-automathtext} \\
                   scimag             &               \url{https://scimag.github.io/sciMAG2015/} \\
   textbook\_quality\_programming & \url{https://huggingface.co/datasets/vikp/textbook\_quality\_programming} \\
                      textbooks & \url{https://huggingface.co/datasets/open-phi/textbooks} \\
         tiny-strange-textbooks~\citep{nam_pham_2024} & \url{https://huggingface.co/datasets/nampdn-ai/tiny-strange-textbooks} \\
                        COIG-PC~\citep{zhang2023chinese} &       \url{https://huggingface.co/datasets/BAAI/COIG-PC} \\
                      FinCorpus & \url{https://huggingface.co/datasets/Duxiaoman-DI/FinCorpus} \\
                        archive &   \url{https://huggingface.co/datasets/linux-cn/archive} \\
                        medical & \url{https://huggingface.co/datasets/shibing624/medical} \\
                   AutoMathText~\citep{zhang2024automathtext} & \url{https://huggingface.co/datasets/math-ai/AutoMathText} \\
                  BioInstructQA & \url{https://huggingface.co/datasets/BioMistral/BioInstructQA} \\
                   SMolInstruct~\citep{yu2024llasmol} & \url{https://huggingface.co/datasets/osunlp/SMolInstruct} \\
                     cosmopedia~\citep{benallal2024cosmopedia} & \url{https://huggingface.co/datasets/HuggingFaceTB/cosmopedia} \\
                     starcoder~\citep{Kocetkov2022TheStack} & \url{https://huggingface.co/datasets/bigcode/starcoderdata}  \\
                    the-stack-v2-train-full-ids~\citep{lozhkov2024starcoder} & \url{https://huggingface.co/datasets/bigcode/the-stack-v2-train-full-ids} \\
                    flan\_v2~\citep{longpre2023flan} &
                    \url{https://github.com/google-research/FLAN/tree/main/flan/v2} \\
                    open-web-math~\citep{paster2023openwebmath} &
                    \url{https://huggingface.co/datasets/open-web-math/open-web-math} \\
\bottomrule
\end{tabular}}
\end{table}
\begin{table}[h]
\centering
\caption{The term ``n ROUND'' indicates the number of rounds for each dataset.}
\label{tab:dataset_round_info}
\resizebox{\textwidth}{!}{
\begin{tabular}{lcccc}
\toprule
\textbf{Dataset} & \textbf{Language} & \textbf{Used during the Fundamental Phase} & \textbf{Used during the Decay Phase} \\ \midrule
MNBVC(gov report) & Chinese & 2 ROUND & - \\ \midrule
US-PD-Books & English & 1 ROUND & 1 ROUND \\ \midrule
MNBVC(law judgement) & Chinese & 2 ROUND & - \\ \midrule
cosmopedia & English & - & 2 ROUND \\ \midrule
AutoMathText & English & 1 ROUND & 2 ROUND \\ \midrule
BioInstructQA & English & 1 ROUND & 2 ROUND \\ \midrule
SMolInstruct & English & 1 ROUND & 2 ROUND \\ \midrule
Agent-FLAN & English & - & 2 ROUND \\ \midrule
MNBVC(gov xuexiqiangguo) & Chinese & 2 ROUND & - \\ \midrule
open-web-math & English & 1 ROUND & 2 ROUND \\ \midrule
The Stack & Code & 2 ROUND & - \\ \bottomrule
\end{tabular}}
\end{table}
\subsection{Detailed Compression Rate}

\subsection{Additional Experimental Results in Scaling Law}\label{scaling-appendix}
\begin{figure}[htbp]
    \centering
    \includegraphics[width=0.9\textwidth]{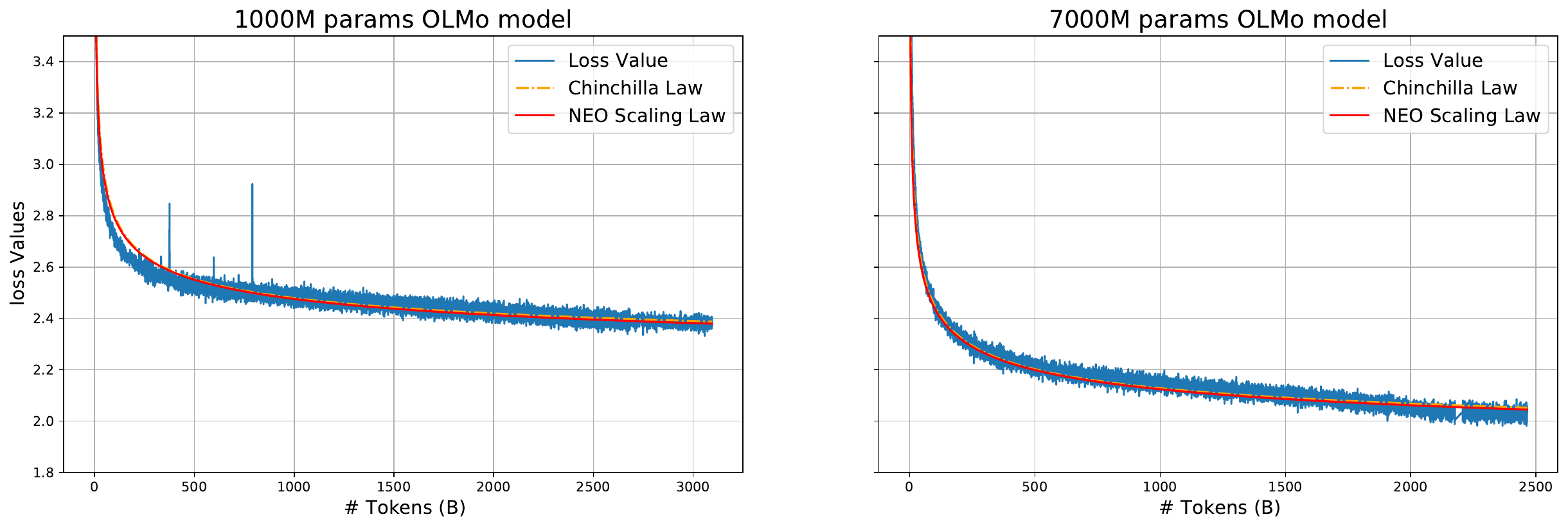}
    \caption{The loss curve, Chinchilla Law prediction and the NEO Scaling law prediction of the OLMo LLM. We use loss values from both 1B and 7B for fitting and prediction.}
    \label{fig:OLMo}
\end{figure}
From figure \ref{fig:OLMo}, We can observe that the Chinchilla law already provides a good fit for OLMo and does not underestimate the loss when the model parameters are large but the data volume is small like MAP-Neo 7B and DeepSeek 67B. Such a phenomenon might be due to the distribution of the pre-training dataset. Deepseek's pre-training data distribution closely resembles that of NEO, with a higher proportion of Chinese data, code data, and high-quality filtered data compared to OLMo whose pre-training data in English 

The Chinchilla scaling law is originally formulated for scenarios where the training data is relatively homogeneous and primarily English-centric. It tends to perform well under these conditions. However, when the training dataset size is smaller (e.g., significantly less than 500 billion tokens) and the model parameter count is high (e.g., 7 billion or more), the diversity of the data leads to a slower reduction in loss than predicted by Chinchilla. Conversely, with larger datasets (e.g., greater than 1.5 trillion tokens), the diversity contributes to a continued decrease in loss, diverging from the flattening and lower-bounded trajectory suggested by the $\frac{B}{D^\beta}$ term in the Chinchilla law. Current evidence is limited as few models are pre-trained across multiple large high-quality corpora. Yi and Qwen~\citep{bai2023qwen} undergo multi-stage pre-training with a significantly smaller initial training corpus compared to MAP-Neo and DeepSeek, while OpenLLaMA~\citep{openlm2023openllama} lacks smaller-scale data to validate these observations.


\clearpage
\subsection{Compression Rate}
\begin{table}[ht]
\caption{Detailed Compression Rates by Category and Dataset.}
\label{tab:detailed_compression}
\renewcommand{\arraystretch}{1.5}
\centering
\begin{tabular}{c|l|c}
\hline
\textbf{Category} & \textbf{Dataset}                               & \textbf{Compression Rate} \\ \hline
\multirow{8}{*}{\centering \textbf{Code}} & Sampled Code(cpp)                            & 2.988 \\ \cline{2-3} 
                                 & Sampled Code(Java)                           & 3.301 \\ \cline{2-3}
                                 & Sampled Code(All)                                & 3.355 \\ \cline{2-3}
                                 & Sampled Github               & 2.988 \\ \cline{2-3}
                                 & Sampled Code(Other)                            & 3.426 \\ \cline{2-3}
                                 & CodeGPT-CN        & 2.458 \\ \cline{2-3}
                                 & Sampled LeetCode                     & 2.050 \\ \cline{2-3}
                                 & \href{https://huggingface.co/datasets/bigcode/starcoderdata}{The Stack V1}                                 & 3.041 \\ \hline
\multirow{5}{*}{\centering \textbf{HQ\_cn}} & \href{https://huggingface.co/datasets/BAAI/COIG-PC}{COIG-PC}                       & 1.835 \\ \cline{2-3}
                                 & Sampled Novel                                 & 1.284 \\ \cline{2-3}
                                 & Sampled Reference Book                    & 1.240 \\ \cline{2-3}
                                 & Exams High Quality    
                                 & 2.290 \\ \cline{2-3}
                                 & Zhihu High Quality                    & 1.377 \\ \cline{2-3}
                                 & Zhihu Instruction                               & 1.434 \\ \hline
\multirow{6}{*}{\centering \textbf{HQ\_en}} & Arxiv High Quality                & 2.976 \\ \cline{2-3}
                                 & Sampled News Paper                           & 3.613 \\ \cline{2-3}
                                 & Sampled English Books                 & 2.079 \\ \cline{2-3}
                                 & \href{https://github.com/google-research/FLAN/tree/main/flan/v2}{flan\_v2}                                   & 3.645 \\ \cline{2-3}
                                 & Huggingface Wiki                                   & 3.520 \\ \cline{2-3}
                                 & \href{https://huggingface.co/datasets/Locutusque/UltraTextbooks}{UltraTextbooks}                           & 4.030 \\ \hline
\multirow{5}{*}{\centering \textbf{Others}} & \href{https://huggingface.co/datasets/math-ai/AutoMathText}{AutoMathText}                            & 2.756 \\ \cline{2-3}
                                 & \href{https://huggingface.co/datasets/BioMistral/BioInstructQA}{BioInstructQA}                            & 3.284 \\ \cline{2-3}
                                 & Synthetic science exam instruction                                      & 1.508 \\ \cline{2-3}
                                 & \href{https://huggingface.co/datasets/open-web-math/open-web-math}{open-web-math}                               & 3.263 \\ \cline{2-3}
                                 & \href{https://huggingface.co/datasets/osunlp/SMolInstruct}{SMolInstruct}                              & 1.978 \\ \hline
\multirow{1}{*}{\centering \textbf{Web\_cn}} & Common Crawl                        & 1.418 \\ \hline
\multirow{1}{*}{\centering \textbf{Web\_en}} & Common Crawl                     & 3.699 \\ \hline
\end{tabular}

\end{table}

\subsection{OCR Post Processing}
\label{ocrpost}
\begin{table}[htbp]
    \centering
    \caption{The OCR prompt templates with Demonstrations in Chinese and English}
    \label{tab: ocr_prompt}
    \begin{CJK*}{UTF8}{gbsn}
    \begin{tabular}{p{7cm} p{7cm}}
    \toprule
            \multicolumn{2}{l}{ {\color{olive}\textit{Prompt Template for OCR Post-processing}} } \\
        \midrule
    \multicolumn{2}{l}{\ding{228} Prompt Templates } \\
    \midrule
    \textbf{Prompt for English Contents} & \textbf{Prompt for Chinese Contents} \\
    \midrule
    From an original document using OCR technology, there may be errors in character recognition, potentially including spelling mistakes, grammatical errors, incorrect punctuation, or formatting issues. Pay special attention to misplaced spaces and line breaks that often occur in OCR-generated content. I need you to reorganize the paragraph into a properly formatted and semantically coherent form. Here's the text I've provided. Kindly check and correct it meticulously. Please output only the revised text without including any additional content i.e. any comments from you. The output format should be a well-organized markdown content. Do not change the language, i.e. Do not change Chinese content to English. Some contents are mixed language i.e. Chinese main content with English symbols. Also, do not change the original language. Please do not generate any unrelated additional comments! Here's one of the texts that needs to be processed: \{content\} You should output: & 请扮演一个AI校对员，我需要你的专业技能来帮助我校对一段文本。这段文本是我通过OCR技术从一份原始文档中提取出来的，我怀疑在字符识别的过程中可能发生了一些错误。具体来说，可能存在拼写错误、语法错误、标点用错或者格式排列问题。请特别注意生成的内容中有很多识别错误的空格与换行符。请将段落整理成正确的语义通顺的格式。输出格式应为组织完善的 Markdown 内容。 不能改变语言，即不能将中文内容改为英文。一些内容是混合语言的，即中文主要内容夹杂英文符号， 请按照原段落位置的语言输出。下面是我提供的文本内容，请你帮我仔细检查并校对，请直接输出修订后文本，并不要包含其他内容。 \{内容\} 你应该输出：\\
    \midrule
    \multicolumn{2}{l}{\ding{228} Demonstrations} \\
    \midrule
    \textbf{English Content Before Post-processing} & \textbf{English Content After Post-processing} \\
    \midrule
    T h e D ev elo p i n g P a th o f C i v il S er v a n t S y stem i n C h i n a ：B ased on C o m p reh en siv e I n ter p reta ti on of C i vi l S er va n t L a w A bst r a c t：C iv i l S e r va nt L a W i S the f h-st com pre he nsi v e l a w of hum an m an a g em e nt of c iv i l se rv an t i n our country．T he civil serv an t system has undergone a great leap forward from Temporary Regulation of C i v i l S erv a nts to C \}vi l S e rv a n t L a w ．C om par i ng to the T emp or a ry R eg ula tio n C \}vi l S e rv a n ts ，C i l S e r va n t W ha s ma ny ne w c ontents i nc lud ing new co nnota tio n of the co nc epts an d som e new rul e s that are written into the law fo r the fi rst time．T here are alSO some adiustments to the former articles． K ey W or ds ：C i v i l serv ant；D e v el opi ng P a th ；C i vi l S e rv a nt L a w 1 1 2 
    &
    The Developing Path of the Civil Servant System in China: Based on Comprehensive Interpretation of the Civil Servant Law
Abstract: The Civil Servant Law is the first comprehensive law on human resource management for civil servants in our country. The civil servant system has made a great leap forward from the Temporary Regulation of Civil Servants to the Civil Servant Law. Compared to the Temporary Regulation of Civil Servants, the Civil Servant Law contains many new contents, including new connotations of concepts and some new rules that are written into the law for the first time. There are also some adjustments to the former articles.
Keywords: Civil servant; Developing path; Civil Servant Law
112 \\
\midrule
    \textbf{Chinese Content Before Post-processing} & \textbf{Chinese Content After Post-processing} \\
    \midrule
    路上 \; \; 只我\; 一个人，\; \; 背\;着手 \;踱着。这一\;片天地 \;好像 \; 是 \; 我的\; ；我 \; \;也像 \; 超出 \;了 \;平常 的自己， \;到\; \; \; 了 \; 另一世 \;界里。我\;爱 \;热闹，\; 也爱 \; 冷静；爱群居， \; 也爱\; \;独处。像 \;今 \;晚上， \; 一 \;个 \;人 \; 在这苍茫的 \;月下 ， 什\;么都\;可以\; \;想，\;什 么 \;都\;可以 \;不\; \;想 \; ， 便\;觉 是 \;个 自 \;由 的 \;人。白 \;天里 \;一定 \;要 \; 做的 \; 事，一 \;定 要 \;说 的 \;话，现 \;在 \; \;都 \; 可 \;不 理 。\;这 是\; 独 处 的\;妙 \;处 ， \;我 且 \; 受 用 这 \;无边 \;的荷 香\;月 \;色 \; 好 了 。

    &
    路上只我一个人，背着手踱着。这一片天地好像是我的；我也像超出了平常的自己，到了另一世界里。我爱热闹，也爱冷静；爱群居，也爱独处。像今晚上，一个人在这苍茫的月下，什么都可以想，什么都可以不想，便觉是个自由的人。白天里一定要做的事，一定要说的话，现在都可不理。这是独处的妙处，我且受用这无边的荷香月色好了。\\
    \bottomrule
    \end{tabular}
    \end{CJK*}

\end{table}

\end{document}